\documentclass[dvipsnames]{article}
\usepackage{arxiv}
\usepackage[utf8]{inputenc} % allow utf-8 input
\usepackage[T1]{fontenc}    % use 8-bit T1 fonts
\usepackage{hyperref}       % hyperlinks
\usepackage{url}            % simple URL typesetting
\usepackage{natbib}

\usepackage{microtype}
\usepackage{graphicx}
\usepackage{subcaption}
\usepackage{booktabs} % for professional tables
\usepackage{wrapfig}
\usepackage{multirow}
\usepackage{float}

\usepackage{titlesec}
\titlespacing*{\paragraph}{0pt}{\baselineskip}{0.5em}

\newcommand*\pct{\mkern2mu\scalebox{.9}{\%}}

\usepackage{math_commands}

\newcommand{\algoname}{Mantis}

\title{MantisV2: Closing the Zero-Shot Gap in Time Series Classification with Synthetic Data and Test-Time Strategies}

\usepackage{authblk}


\author{Vasilii Feofanov}
\author{%
    {Songkang Wen}%
}
\author{%
    {Jianfeng Zhang}%
}
\author{%
    {Lujia Pan}%
}
\author{%
    {Ievgen Redko}%
}
\affil{}

\date{\vspace{-0.7cm}}

\renewcommand{\shorttitle}{\algoname V2: Foundation Model for Time Series Classification}

\definecolor{mantis_lightgreen}{HTML}{4a9a45}
\definecolor{mantis_green}{HTML}{71b431}
\definecolor{mantis_magenta}{HTML}{733fd8}
\definecolor{mantis_blue}{HTML}{3fb7d8}
\definecolor{mantis_pink}{HTML}{d83fa4}
\definecolor{mantis_seagreen}{HTML}{3fd8bf}
\definecolor{mantis_orange}{HTML}{f7aa1a}
\definecolor{mantis_yellow}{HTML}{d8d23f}
\definecolor{tabblue}{HTML}{16537e}
\definecolor{mypurple}{HTML}{95459a}

\hypersetup{
    pdftitle={MantisV2: Closing the Zero-Shot Gap in Time Series Classification with Synthetic Data and Test-Time Strategies},
    pdfauthor={V.~Feofanov, S.~Wen, L.~Pan, J.~Zhang, I.~Redko},
    colorlinks=true,
    linkcolor=black,
    citecolor=mantis_lightgreen,
    citebordercolor=mantis_lightgreen!60,
    filecolor=magenta,
    urlcolor=mantis_blue,
}

\begin{document}

\maketitle

\begin{abstract}
Developing foundation models for time series classification is of high practical relevance, as such models can serve as universal feature extractors for diverse downstream tasks. Although early models such as Mantis have shown the promise of this approach, a substantial performance gap remained between frozen and fine-tuned encoders. In this work, we introduce methods that significantly strengthen zero-shot feature extraction for time series. First, we introduce Mantis+, a variant of Mantis pre-trained entirely on synthetic time series. Second, through controlled ablation studies, we refine the architecture and obtain MantisV2, an improved and more lightweight encoder. Third, we propose an enhanced test-time methodology that leverages intermediate-layer representations and refines output-token aggregation. In addition, we show that performance can be further improved via self-ensembling and cross-model embedding fusion. Extensive experiments on UCR, UEA, Human Activity Recognition (HAR) benchmarks, and EEG datasets show that MantisV2 and Mantis+ consistently outperform prior time series foundation models, achieving state-of-the-art zero-shot performance.
% The source code of our Python package is available on \href{https://github.com/vfeofanov/mantis}{GitHub}.
\end{abstract}

\begin{table}[H]
    \centering
    \begin{tabular}{r l}
    \textbf{Mantis+}: & \url{https://huggingface.co/paris-noah/MantisPlus}\\
    \textbf{MantisV2}: & \url{https://huggingface.co/paris-noah/MantisV2}\\
    \textbf{CauKer 2M}: & \url{https://huggingface.co/datasets/paris-noah/CauKer2M}
    \end{tabular}
    % \caption{Caption}
    % \label{tab:placeholder}
\end{table}

\section{Introduction}

Classification of time series data is a fundamental problem across science and industry, arising in domains such as activity / action recognition~\citep{chen2025comodo,li2025zara,jie2026novel}, power electronics~\citep{liao2025pe,li2025data}, observability~\citep{feng2025telecomts}, healthcare~\citep{alchieri2025exploring,wong2025large}, finance~\citep{lee2023stockemotions}, and neuroscience~\citep{wang2024cbramod,gnassounou2025leveraging}. Following the success of foundation models in vision~\citep{radford2021CLIP} and language~\citep{achiam2023gpt}, the development of time series foundation models (TSFMs) has become an active research direction. These models aim to serve as universal feature extractors for diverse downstream tasks, reducing the need for extensive labeled data and simplifying the process of model selection and tuning.

Over the past three years, a wide variety of TSFMs have been introduced. Several approaches adopt decoder-only architectures for forecasting~\citep{cohen2025toto,auer2025tirex,ansari2025chronos2}, while others rely on masked autoencoders~\citep{goswami2024moment}, adapt large language~\citep{zhou2023onefitsall,ashok2025beyond} or vision models~\citep{chen2024visionts,roschmann2025tivit}. For TSFMs designed specifically for classification, the main goal is to learn discriminative embeddings, thereby making self-distillation~\citep{lin2024nutime} and contrastive objectives prevail~\citep{albelwi2022survey}.
% the dominant strategy is to rely on self-supervised pre-training contrastive self-supervised pre-training~\citep{albelwi2022survey} to obtain discriminative embeddings. 
Among these, Mantis~\citep{feofanov2025mantis} stands out for achieving strong feature-extraction performance while remaining lightweight and operating in a purely frozen-encoder (zero-shot) regime.

Despite this progress, two important limitations remain.
First, previous Mantis pre-training pipelines suffered from data leakage, as the pre-training corpus overlapped with evaluation datasets.
Second, the performance gap between frozen and fine-tuned Mantis is still substantial, raising doubts about whether current foundation models can surpass modern self-supervised methods in time series classification.
To address these issues, we present \textbf{MantisV2} and \textbf{Mantis+}, a new generation of Mantis-based time series foundation models built around three key ideas:
\begin{enumerate}
\item \textit{Synthetic-data pre-training.} We show that pre-training on large-scale synthetic time series~\citep{xie2025cauker}, generated to cover a broad range of temporal patterns, yields substantially more generalizable representations. Re-training the original Mantis architecture on synthetic data only produces Mantis+.
\item \textit{Architecture refinement.} Through controlled ablation studies, we streamline and improve the architectural design of Mantis, resulting in MantisV2, a more lightweight yet more performant encoder.
\item \textit{Test-time optimization.} We introduce a comprehensive inference-time pipeline that enhances representation quality without any additional pre-training or fine-tuning. Our method incorporates the use of intermediate-layer representations, improved output-token aggregation, input-perturbation self-ensembling, and cross-model embedding fusion. Together, these components substantially improve the robustness and expressiveness of the frozen encoder.
\end{enumerate}

Our extensive experiments on UCR~\citep{dau2019ucr}, UEA~\citep{bagnall2018uea}, Human Activity Recognition (HAR), and EEG datasets demonstrate the effectiveness of our approach. MantisV2 and Mantis+ consistently outperform prior TSFMs in the zero-shot setting. Moreover, we show that MantisV2 outperforms competitive self-supervised models such as TS2Vec~\citep{yue2022ts2vec} and T-Loss~\citep{franceschi2019tloss}, while its fusion with a vision backbone~\citep{roschmann2025tivit} can match the performance of fine-tuned Mantis, closing the zero-shot gap. We release all models by open-sourcing the pre-trained checkpoints on HuggingFace, updating the Mantis package on \href{https://github.com/vfeofanov/mantis}{GitHub}, and publishing the synthetic pre-training dataset on HuggingFace (\textbf{CauKer 2M}).
% at the \href{https://github.com/vfeofanov/mantis}{Mantis GitHub package} with  and open-source t, and evaluation code to facilitate further research in universal time series representation learning.
% The pre-trained checkpoint is open-sourced on \href{https://huggingface.co/paris-noah/Mantis-8M}{HuggingFace}, and  includes an API for inference and fine-tuning on new downstream tasks. 
% \begin{table}[H]
%     \centering
%     \begin{tabular}{r l}
%     \textbf{Mantis+}: & \url{https://huggingface.co/paris-noah/MantisPlus}\\
%     \textbf{MantisV2}: & \url{https://huggingface.co/paris-noah/MantisV2}\\
%     \textbf{CauKer 2M}: & \url{https://huggingface.co/datasets/paris-noah/CauKer2M}
%     \end{tabular}
%     % \caption{Caption}
%     % \label{tab:placeholder}
% \end{table}

\begin{figure}[t!]
    \centering
    \includegraphics[width=\linewidth, clip=True, trim=0cm 0.5cm 0.6cm 0cm]{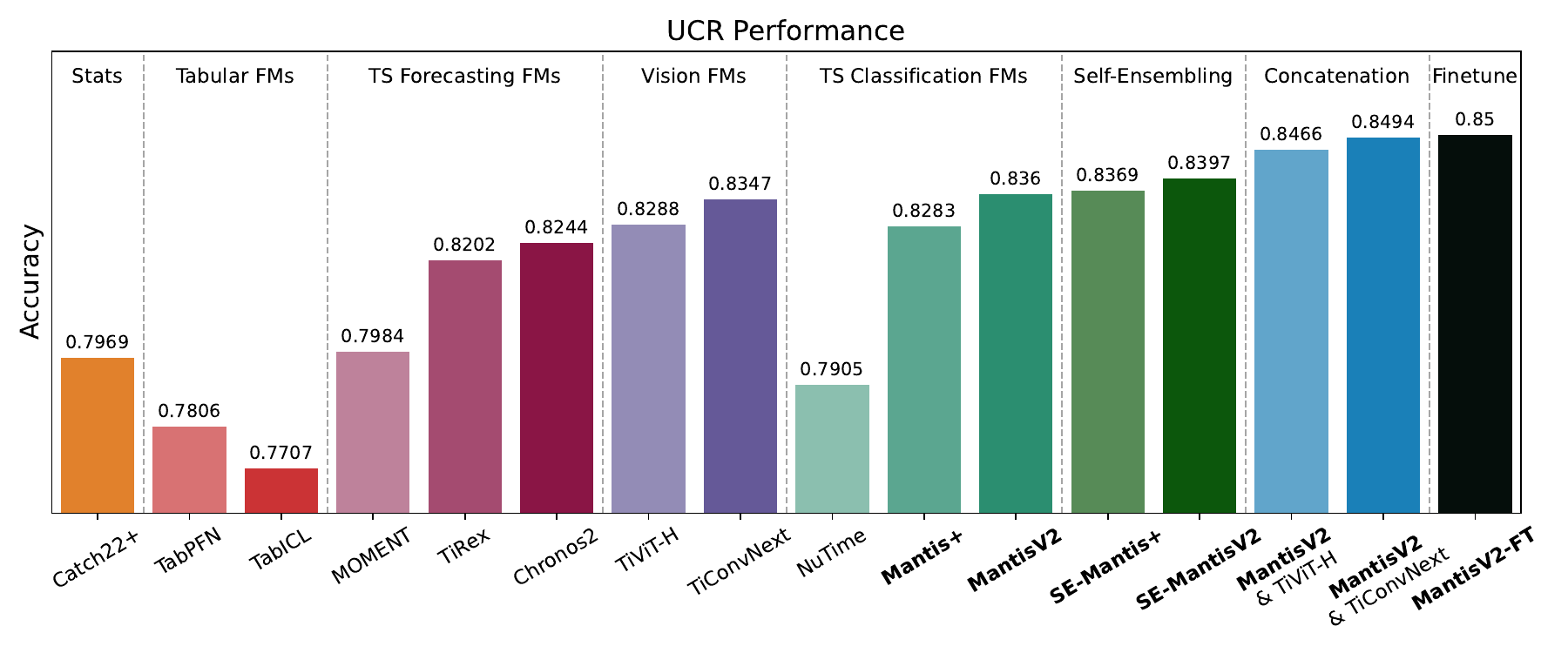}
    \caption{Final Performance on the UCR Benchmark.}
    \label{fig:teaser-plot}
\end{figure}

% \begin{figure}
%     \centering
%     \includegraphics[width=\textwidth, clip=True, trim=1cm 0 1cm 0]{pics/mantis_architecture.pdf}
%     \caption{Architecture. By symbol $+$ we denote the sum operator, while $||$ designates the vector concatenation operator.}
%     \label{fig:mantis-architecture}
% \end{figure}

% \vspace{-0.3cm}
\section{Methodology}
\label{sec:method}
In this section, we present the main technical details behind Mantis: we mathematically introduce the problem setup, present the architecture and the pre-training process.

\subsection{Problem Setup}
\label{sec:problem-setup}
Mathematically speaking, our time series classification foundation model is an encoder $F: \R^t \to \R^{q}$ that projects any time series $\mbf{x}\in\R^t$ with a fixed sequence length $t$ to a discriminative hidden space $\R^{q}$. During the pre-training phase, we observe an unlabeled pre-training set $\mathrm{X}_{\text{0}}$ that is sufficiently large in order to learn rich embeddings 
that generalize well across different tasks. During the fine-tuning phase, we observe a supervised downstream task with observations $\mathrm{X}$ and labels $\mathrm{Y}$. The two options are possible: 1) we use $F$ to extract deep embeddings $\mathrm{Z}=\{F(\mbf{x}),\ \mbf{x}\in\mathrm{X}\}$ and then learn any classifier $h:\R^{q}\to \{1,\dots,K\}$ using $\mathrm{Z}$ as features and $\mathrm{Y}$ as corresponding labels, 2) append a classification head $h:\R^{q}\to \R^K$ and fine-tune $h\circo F$ by minimizing a loss function evaluated on the downstream dataset.

When time series with multiple channels $\mbf{x}=[\mbf{x}_1,\dots,\mbf{x}_d]\in\R^{d\times t},\ d>1$, are considered, we send each channel $\mbf{x}_i,\ i\in[1,d]$, to the TSFM independently, i.e.,
% $\mbf{z}=\circleWithBars\left[F(\mbf{x}_i)\right]_{i=1}^d$, 
the embedding of $\mbf{x}$ is defined as
$\mbf{z}=\textrm{concat}\left[(F(\mbf{x}_i))_{1\leq i\leq d}\right]$, where $\textrm{concat}$ denotes the vector concatenation operator, and the input dimension of the classifier (head) is $\R^{d\times q}$. Alternatively, we can use adapters to mix channels and feed them to the encoder \citep{ilbert2025user,benechehab2025adapts}. As this approach is complementary and orthogonal to the focus of our work, we leave its integration with the proposed model variants for future research.

% When it comes to quantifying the uncertainty of a fine-tuned foundation model, we consider its predicted probabilities $\sigma(h\circo F(\mbf{x}))$, with $\sigma: \R^K\to\Delta_K$ being the softmax transformation that projects logits onto the probability simplex $\Delta_K = \{\mbf{p} \in[0, 1]^K\,|\sss\norm{\mbf{p}}_1 = 1\}$. We particularly focus on the top probability $\textrm{conf}(\mbf{x})=\max\left[\sigma(h\circo F(\mbf{x}))\right]$, which we consider the confidence of the model in classifying $\mbf{x}$. For the sake of simplicity, the predicted label we denoted by $\hat{y} = \argmax\left[\sigma(h\circo F(\mbf{x}))\right]$.

% Since time series data may vary in sequence lengths and units of measurement, pre-processing is required for both pre-training and inference. Similar to practices in computer vision, we pre-train the model with fixed input dimensions and resize inputs during inference and fine-tuning. Specifically, we set the input sequence length of Mantis to 512 and resize inputs using PyTorch's interpolation function~\citep{paszke2019pytorch}. To address the issue of varying units of measurement, we apply instance-level standard scaling. For each time series observation (and, in the case of multiple channels, for each individual channel), we subtract the mean and divide by the standard deviation calculated across the time steps. This scaling process is implemented directly within the model architecture, ensuring that inputs are transformed during the forward pass.

\subsection{Architecture}
\label{sec:architecture}

In this section, we describe the architecture of Mantis, which adapts the Transformer architecture \citep{vaswani2017transformer} to time series data through a dedicated tokenization strategy. 
% The full architecture is illustrated in Figure~\ref{fig:mantis-architecture}. 
We fix the number of tokens to 32, which implies that the input sequence length $t$ must be proportional to 32. This design choice differs from approaches such as \citet{lin2024nutime} and \citet{goswami2024moment}, which instead fix the patch (token) length. In the earlier stages of development, we empirically find that both strategies lead to comparable performance. However, fixing the number of tokens is preferable under computational constraints, as the self-attention mechanism scales quadratically with respect to the number of tokens. As a requirement, an input time series should be resized/padded to a certain length. While \citet{goswami2024moment} and \citet{lin2024nutime} require the sequence length proportional to the patch size, in our case, the length should be proportional to the number of tokens, i.e., 32. By default, we resize all inputs to length 512. Further analysis of the impact of the interpolation length is provided in Section~\ref{sec:self-ensembling}.

\vspace{0.2cm}
\textbf{Token Generator Unit. }
The first stage of the model encodes a raw time series into a set of meaningful tokens, which are subsequently processed by the Transformer. The token generation procedure consists of the following steps:

\begin{itemize}
\item[a.] \textit{Instance normalization.} For each time series instance, we subtract the mean and divide by the standard deviation computed across time steps. This normalization is implemented as part of the model architecture and applied during the forward pass.

\item[b.] \textit{Patch extraction from the signal.} We extract patch-level representations by applying a single convolutional layer with a fixed kernel size, followed by mean pooling configured to produce 32 patches. The convolution outputs 256 channels, resulting in patch embeddings of dimension 256.

\item[c.] \textit{Patch extraction from the first-order differential.} We apply the same patching procedure to the first-order temporal difference of the time series, computed as the difference between adjacent time steps. This differential representation encourages stationarity and reduces the influence of long-term trends.

\item[d.] \textit{Patch-wise statistics encoding.} To preserve information about the original measurement scale, we split the raw (unnormalized) time series into 32 non-overlapping patches. For each patch, we compute the mean and standard deviation and encode these statistics using the Multi-Scaled Scalar Encoder \citep{lin2024nutime}.

\item[e.] \textit{Token projection.} All patch-level features (signal, differential signal, and statistical descriptors) are concatenated and passed through a linear projection layer followed by layer normalization \citep{ba2016layernormalization}, producing the final set of 32 tokens with dimensionality 256.

\end{itemize}

\textbf{Transformer. }
The resulting tokens are processed by a Transformer encoder, as summarized below:

\begin{itemize}
    \item[a.] \textit{Class token.} A learnable class token is prepended to the 32 generated tokens. This token attends to all other tokens and aggregates global information, with its final representation serving as a summary of the entire input sequence.
    
    \item[b.] \textit{Positional encoding.} Positional information is incorporated using positional encodings. In the original Mantis architecture, we employ sinusoidal positional embeddings \citep{vaswani2017transformer}, which are added to the input tokens. Alternatively, Rotary Positional Encoding (RoPE; \citealp{su2024roformer}) can be used which rotates the query and key representations.
    
    \item[c.] \textit{Transformer layers.} We apply six Transformer layers, each consisting of multi-head self-attention with eight heads followed by a feedforward network. All layers use a pre-normalization design.

    \item[d.] \textit{Output representation.} The final hidden state of the class token is taken as the output embedding of the foundation model.
\end{itemize}

\textbf{Projector and Prediction Head. }
Depending on the training or evaluation regime, different heads are appended to the Transformer output:

\begin{itemize}
    \item \textit{Pre-training:} A layer normalization followed by a linear projection is applied to produce embeddings used for similarity-based objectives.
    
    \item \textit{Fine-tuning:} A task-specific classification head maps the embeddings to class logits.
    
    \item \textit{Inference:} No additional layers are applied, and the Transformer output embedding is returned directly.
\end{itemize}

\subsection{Pre-training}
\label{sec:pre-training}

We pre-train Mantis in a self-supervised manner using a contrastive learning objective. The goal is to learn an encoder that produces similar representations for two random augmentations of the same time series (a positive pair), while producing dissimilar representations for augmentations of different time series (negative pairs). Formally, let $\mathcal{T}$ be a space of transformations (augmentations) such that for all $\phi\in\mathcal{T}$ and $\mbf{x}\in\mathcal{X}$ we have $\phi(\mbf{x})\in\mathcal{X}$. To measure the similarity between two embeddings, we first project the output of the foundation model $F(\mbf{x})$ to a new dimension using a projector $g: \R^{q} \to \R^{q'}$ and then compute the cosine similarity between the two vectors defined as follows:
\begin{align*}
    s_{\cos}(\mathbf{a}, \mathbf{b}) := \frac{\mbf{a}^\top\mbf{b}}{\norm{\mbf{a}}\cdot\norm{\mbf{b}}},\qquad \forall(\mbf{a}, \mbf{b})\in\R^{2q'}.
\end{align*}
Given a batch $B=\{\mbf{x}_i\}_{i=1}^b$, for each example $\mbf{x}_i$, we independently sample two augmentation functions $\phi$ and $\psi$ uniformly from $\mathcal{T}$, i.e., $\phi,\psi\sim\mathcal{U}(\mathcal{T})$. We then compute the pairwise similarities between all the examples in the following way:
\begin{align*}
    \mbf{s}_i(\phi, \psi) = \left[s_{\cos}\left(g\circo F\circo\phi(\mbf{x}_i),  \, g\circo F\circo\psi(\mbf{x}_j)\right)\right]_{j=1}^b \in \R^b.
\end{align*}

Following \citet{oord2018representation} and \citet{he2020momentum}, and denoting the cross-entropy loss by $l_{\text{ce}}: \R^b\times\{1,\dots,b\}\to\R$, we update the parameters of $F$ and $g$ by minimizing the contrastive objective defined by
% defined by $\sum_{i=1}^b l_{\text{ce}}\left(\frac{\mbf{s}_i(\phi, \psi)}{T},\  i\right),$
\begin{align*}
    \sum_{i=1}^b l_{\text{ce}}\left(\frac{\mbf{s}_i(\phi, \psi)}{T},\  i\right),
\end{align*}
where $T\in(0,+\infty)$ is a temperature, which we fix to $0.1$ in all experiments.

\setlength{\intextsep}{-5pt}%
\setlength{\columnsep}{10pt}%
\begin{wrapfigure}[12]{r}{0.27\textwidth}
\includegraphics[width=0.26\textwidth, clip=True, trim=0 0 0 0]{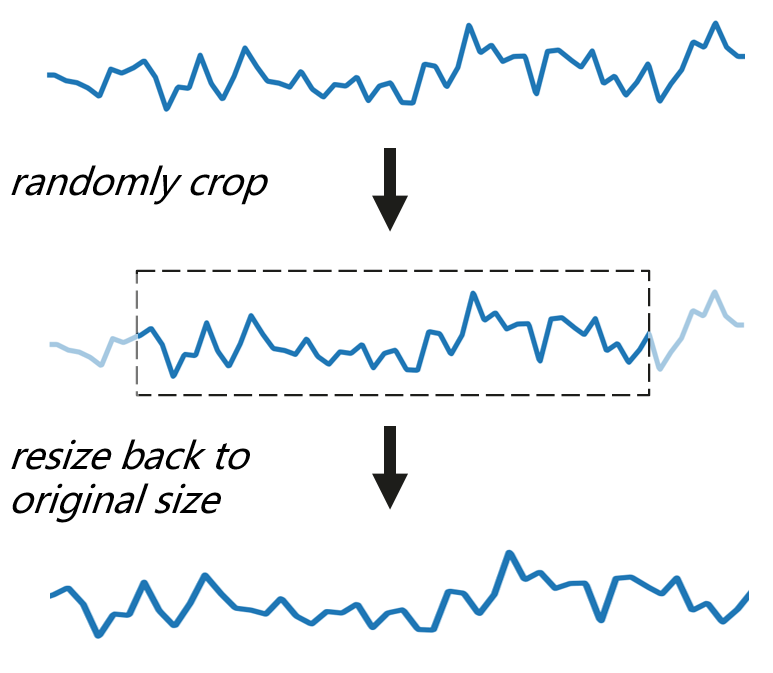}
\caption{Random Crop Resize.}
\label{fig:randomcropresize}
\end{wrapfigure}
\textbf{Augmentation. } We empirically evaluated several time-series augmentation strategies and observed that their effectiveness is highly dataset-dependent, as they may aggressively distort the signal and remove discriminative information.
For pre-training, we have chosen the Random Crop Resize (RCR) augmentation (Figure \ref{fig:randomcropresize}). 
% which involves randomly cropping $c\pct$ of a time series (with the condition that the remaining $(1\!-\!c)\pct$ forms a contiguous window) and then resizing the cropped portion to the original sequence length. 
This transformation randomly crops a contiguous segment covering $(1\!-\!c)\pct$ of the original time series and then resizes it back to the original sequence length. We apply moderate distortions by sampling the crop rate $c$ uniformly between $0\pct$ and $20\pct$, thereby preserving the overall temporal structure of the signal. A key advantage of contrastive learning with RCR is that the encoder is encouraged to be invariant to small temporal stretches and compressions, which in turn enables flexible resizing of the input without degrading performance.
It is important to note that RCR perturbs the original unit measurements, making it less suitable for forecasting tasks. However, since our focus is on time series classification, the primary goal is to capture discriminative temporal patterns rather than preserve absolute scales. Empirically, we find RCR to be the most effective augmentation for this purpose.

% As a pre-training dataset, we use a union of various public datasets including UCR~\citep{dau2019ucr}, UEA~\citep[all except EigenWorms and InsectWingbeat]{bagnall2018uea}, ECG~\citep{clifford2017ecgdataset}, EMG~\citep{goldberger2000emgdataset}, Epilepsy~\citep{andrzejak2001epilepsydataset}, FD-A and FD-B \citep{lessmeier2016fdafdbdataset}, Gesture~\citep{liu2009gesturedataset}, HAR~\citep{anguita2013hardataset}, SleepEEG~\citep{kemp2000sleepeegdataset}. We ensure that test sets used for evaluation in Section \ref{sec:experiments} are not part of the pre-training dataset. The total pre-training consists of 1.89 million time series examples. We pre-trained the model for 100 epochs with a batch size equal to 2048 on 4 NVIDIA Tesla V100-32GB GPUs.

\section{Key Improvements}
\label{sec:key-improvements}

In this section, we describe the proposed methodology used to improve the zero-shot performance of Mantis. Since optimizing the architecture of a foundation model is a complex and high-dimensional problem, we conduct a series of controlled ablation studies in which each modification is evaluated relative to the original Mantis architecture. The final MantisV2 architecture is obtained by combining all components that individually lead to performance improvements.
In parallel, we show that the proposed methodology can also substantially improve the original Mantis model without any architectural changes; we refer to this enhanced variant as Mantis+. All architectural and methodological choices are evaluated on the UCR benchmark \citep{dau2019ucr}. Experiments are conducted on NVIDIA Tesla V100 GPUs with 32GB of memory: pre-training is performed using four GPUs, while feature extraction and evaluation are carried out on a single GPU.

\subsection{Pre-training with Synthetic Data}

Recently, \citet{xie2025cauker} proposed CauKer, a synthetic time-series generation framework based on Gaussian process kernel composition and structural causal models. A key and somewhat surprising finding of their work is that Mantis can be pre-trained entirely on synthetic data while retaining strong downstream performance. Unlike the original Mantis pre-training corpus (1.89 million samples), which partially overlapped with UCR and UEA training sets and therefore did not constitute a strictly out-of-distribution (OOD) setting, CauKer-generated data are OOD by construction. This property substantially increases the reliability of zero-shot evaluation results.

To further validate the effectiveness of synthetic pre-training data, we compare 100{,}000 time series generated by CauKer against two alternative pre-training datasets: (i) Anomaly UCR \citep{wu2021current}, a collection of real-world datasets for anomaly detection, and (ii) a 100{,}000-sample subset of the original Mantis pre-training corpus that explicitly excludes all UCR and UEA samples. Table~\ref{tab:pre-train-data-comparison} reports zero-shot feature extraction results following the evaluation protocol used by \citet{feofanov2025mantis}. Specifically, a Random Forest classifier \citep{breiman2001random} is trained on encoded training time series and evaluated on test sets, with accuracy averaged over 128 UCR datasets. The results show that synthetic data not only match but even outperform real data of the same size, demonstrating their high effectiveness for pre-training.

\vspace{0.3cm}
\begin{table}[ht!]
\centering
\caption{Performance of Mantis on UCR collection with different pre-training datasets.}
\label{tab:pre-train-data-comparison}
\begin{tabular}{lcccc}
\toprule
Pre-training Data   & Nature       & Size & UCR Included?                                                      & UCR acc. (\%) \\ \midrule
Anomaly           & Real             & 38K  &   \textcolor{ForestGreen}{No}                    & $0.7473_{\pm 0.0014}$          \\

Subset of V1 Dataset & Real & 100K                       & \textcolor{ForestGreen}{No} &   $0.7829_{\pm 0.0008}$       \\
CauKer  & Synth & 100K                     & \textcolor{ForestGreen}{No}                     & $78.81_{\pm 0.001}$         \\
V1 Dataset          & Real            & 1.89M                    & \textcolor{BrickRed}{Yes}                       & $79.21_{\pm 0.0012}$      \\
\bottomrule
\end{tabular}
\end{table}
\vspace{0.55cm}

This observation can be intuitively explained by the nature of the pre-training objective. Self-supervised contrastive learning explicitly promotes \emph{uniformity} in the embedding space, encouraging representations to be evenly distributed \citep{wang2020understanding}. Achieving such uniformity requires high data diversity, which synthetic generation methods such as CauKer can provide in a scalable and controllable manner. As a result, synthetic data offer both strong sample efficiency and improved generalization performance.

% \href{https://huggingface.co/datasets/paris-noah/CauKer2M}{https://huggingface.co/datasets/paris-noah/CauKer2M}.
In the remainder of Section~\ref{sec:key-improvements}, we use 100{,}000 synthetic samples for pre-training unless stated otherwise. This choice allows us to significantly reduce pre-training time while maintaining high accuracy, as evidenced in Table~\ref{tab:pre-train-data-comparison}. Each model is pre-trained using three different random seeds, and we report the average accuracy across seeds and the 128 UCR datasets. For downstream classification, we employ a Random Forest classifier with 200 trees and unlimited maximum depth. Once architectural choices are finalized, we pre-train the selected models on the full set of 2 million synthetic time series, which is publicly available at \href{https://huggingface.co/datasets/paris-noah/CauKer2M}{HuggingFace}.

\subsection{Refining Architecture through Ablation}
\label{sec:refined-architecture}

Before introducing improvements to Mantis, we first conduct additional ablation studies to further validate several design choices made in the original architecture. In particular, we analyze the proposed Token Generator Unit, which combines three parallel branches: tokens extracted from the raw time series, tokens derived from its first-order differential, and the encoded patch-wise statistics (mean and standard deviation).

\vspace{0.35cm}
\begin{figure}[ht]
    \centering
    \includegraphics[width=0.7\linewidth, clip=true, trim=0 1cm 0 0]{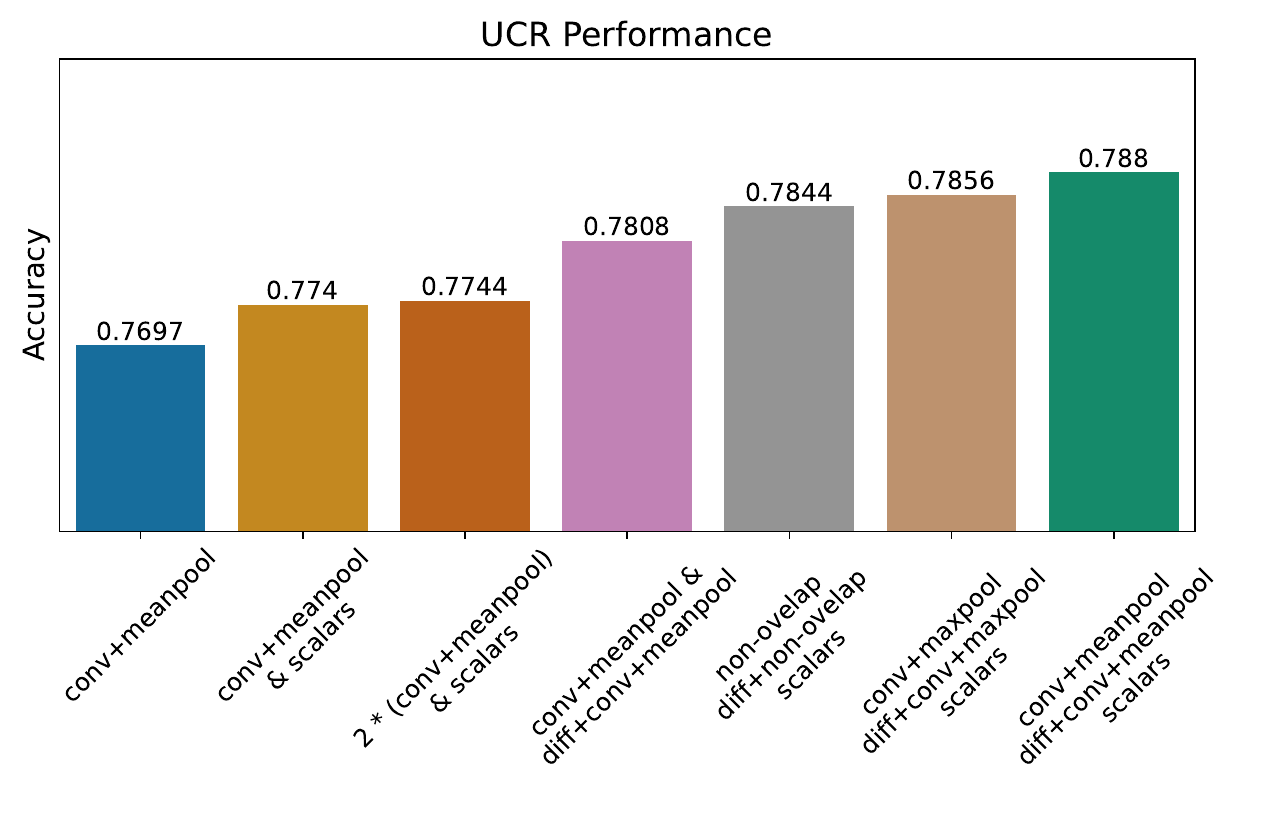}
    \caption{Ablation study confirming the proposed Token Generator Unit.}
    \label{fig:ablation-tokenizer}
\end{figure}
\vspace{0.35cm}

We compare the proposed tokenizer against the following baselines:
\begin{itemize}
\item Only the first branch is used, consisting of a convolutional layer followed by mean pooling.
\item The first and third branches are combined, i.e., patch-wise scalar encoding is added to the raw signal branch.
\item The first branch is duplicated and combined with the scalar statistics branch. This baseline is introduced to determine whether the benefit of the second branch arises from incorporating the first-order differential or simply from increasing the number of parameters.
\item The first and second branches are combined, removing the scalar encoder.
\item All three branches are retained, but the convolution with mean pooling is replaced by embeddings of non-overlapping patches. This baseline is introduced to compare the proposed convolution-based patching against the non-overlapping patching used in NuTime~\citep{lin2024nutime}.
\item All three branches are retained, but mean pooling is replaced with max pooling to assess the impact of the pooling strategy.
\end{itemize}

Figure~\ref{fig:ablation-tokenizer} presents the corresponding results. Each branch of the proposed Token Generator Unit contributes positively to performance (when comparing the 1st, 2nd, 4th, and 7th bars). Moreover, comparing the 3rd and 4th bars shows that incorporating features derived from the first-order differential yields a clear performance gain, confirming that the improvement is not merely due to increased model capacity. Regarding patching strategies, convolution with mean pooling consistently outperforms both max pooling and the non-overlapping patch embedder.

\vspace{0.3cm}
\begin{figure}[ht!]
    \centering
    \begin{subfigure}{0.48\linewidth}
        \includegraphics[width=\textwidth]{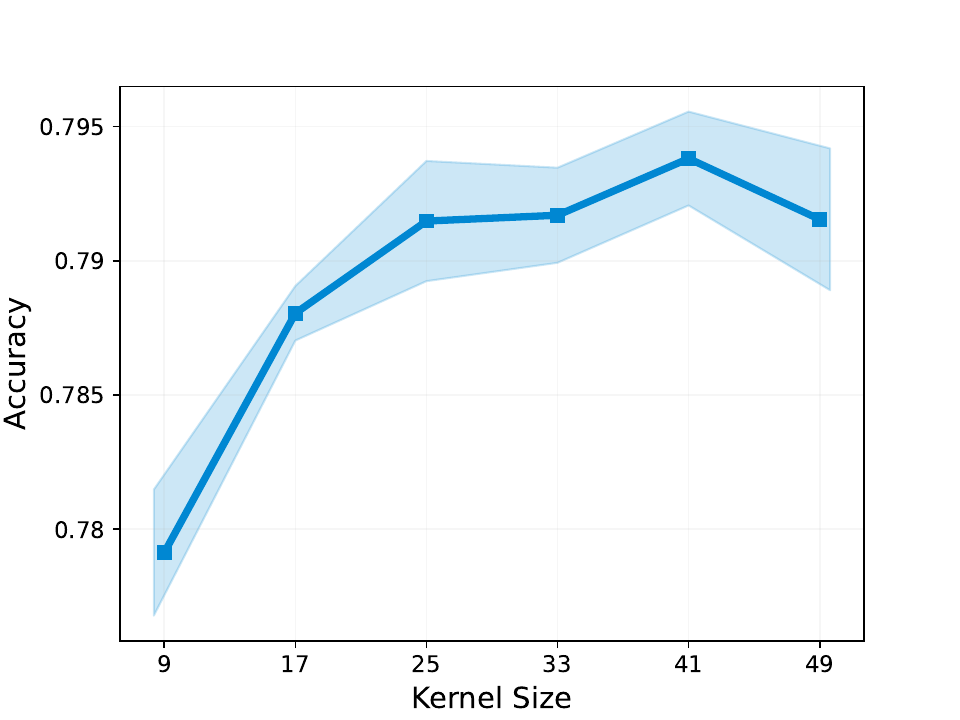}
        \caption{Convolution kernel size.}
        \label{fig:ablation-kernel}
    \end{subfigure}
    \begin{subfigure}{0.48\linewidth}
        \includegraphics[width=\textwidth]{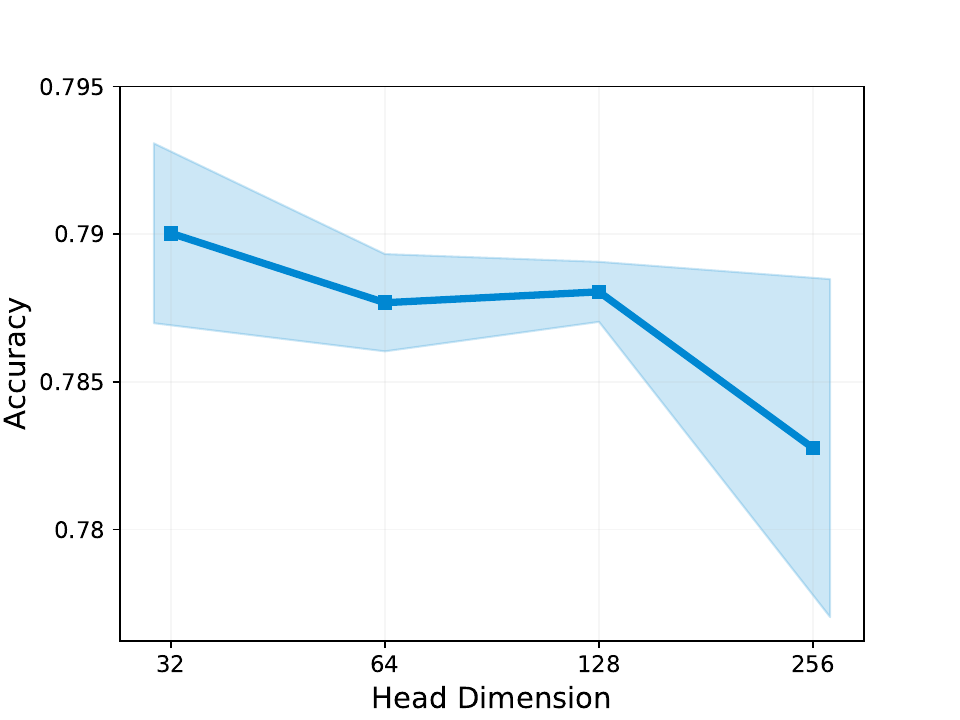}
        \caption{Transformer head dimension.}
        \label{fig:ablation-dimhead}
    \end{subfigure}
    \caption{Ablation study on different hyperparameters of the architecture.}
    \label{fig:ablation-architecture}
\end{figure}
\vspace{0.4cm}

Continuing our ablation study, we investigate the impact of several architectural hyperparameters. First, we vary the kernel size of the convolutional layers, which was originally set to 17. We evaluate kernel sizes in ${9, 17, 25, 33, 41, 49}$ and observe that performance improves monotonically up to a kernel size of 41, which provides the best trade-off between temporal resolution and receptive field (Figure~\ref{fig:ablation-kernel}).

Next, we study the effect of the per-head projection dimension in the Transformer, varying it over ${32, 64, 128, 256}$ (Figure~\ref{fig:ablation-dimhead}). In standard Transformer implementations, this dimension is typically set to the hidden dimension divided by the number of attention heads (32 for 8 heads). In the original Mantis architecture, it was increased to 128. However, when averaging performance over three random seeds, this increase does not yield consistent gains, indicating a seed-dependent bias of the previous conclusion. Consequently, we revert to the default value of 32, which slightly improves average performance while reducing the number of parameters.

\begin{figure}[ht!]
    \centering
    \includegraphics[width=0.5\linewidth]{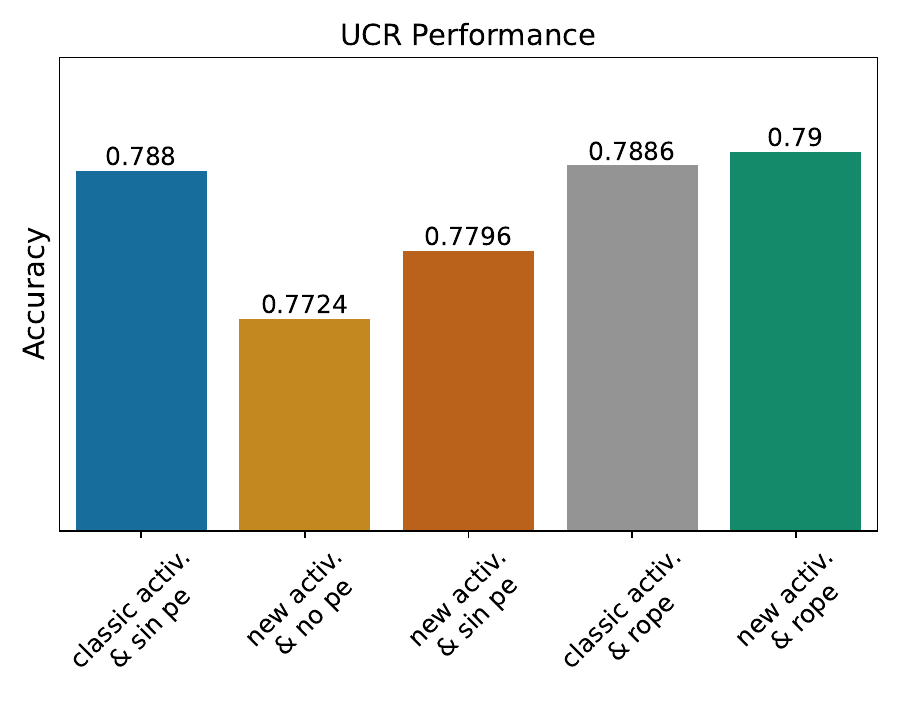}
    \caption{Comparison of different transformer configurations.}
    \label{fig:ablation-transformer}
\end{figure}

% Our another improvement concerns the architecture of the transformer. We have decided to try another architecture following \citep{cohen2025toto}. In principal, there are 3 main differences: Rotary Positional Encoding (RoPE, \citeauthor{su2024roformer}, \citeyear{su2024roformer}),  RMS Layer Normalization \citep{zhang2019root} instead of the usual one, and SwiGLU~\citep{shazeer2020glu} instead of GELU in the feedforward part of a transformer layer. We have performed an ablation study by comparing the following version of Mantis: (a) the original variant that use classical activations/normalizations (referred further just by activations) and sinusoidal position encoding, (b) new activations and no positional encoding, (c) new activations and sinusoidal positional encoding, (d) classical activations and RoPE, (e) new activations and RoPE. Figure \ref{fig:ablation-transformer} depicts the performance on the UCR benchmark in average over 3 seeds. One can see that combination of SwiGLU, RMS Layer Normalization and RoPE give the best performance. Although the improvement seems very little, it appears consistent, we demonstrate this also for 1 million pre-training data in Appendix \ref{sec:appendix-arch-refine}.
Finally, we explore refinements to the Transformer architecture itself, inspired by the design choices of \citet{cohen2025toto}. Specifically, we consider three modifications: Rotary Positional Encoding (RoPE; \citealp{su2024roformer}), RMS Layer Normalization \citep{zhang2019root} in place of standard Layer Normalization, and the use of SwiGLU activations \citep{shazeer2020glu} instead of GELU in the feedforward layers. We empirically evaluate the following variants: (a) Classical activations and normalization with sinusoidal positional encoding (original Mantis), (b) SwiGLU and RMS normalization without positional encoding, (c) SwiGLU and RMS normalization with sinusoidal positional encoding, (d) Classical activations and normalization with RoPE, (e) SwiGLU and RMS normalization with RoPE.

% \begin{itemize}
% \item Classical activations and normalization with sinusoidal positional encoding (original Mantis).
% \item SwiGLU and RMS normalization without positional encoding.
% \item SwiGLU and RMS normalization with sinusoidal positional encoding.
% \item Classical activations and normalization with RoPE.
% \item SwiGLU and RMS normalization with RoPE.
% \end{itemize}
% \setlength{\intextsep}{-5pt}%
% \setlength{\columnsep}{10pt}%
% \begin{wrapfigure}[15]{r}{0.45\textwidth}
% \centering    
% \vspace{-0.2cm}
%     \includegraphics[width=0.9\linewidth]{pics/version2/ablation-transformer.pdf}
%     \caption{Different transformer configurations.}
%     \label{fig:ablation-transformer}
% \end{wrapfigure}
Figure~\ref{fig:ablation-transformer} reports the corresponding results on the UCR benchmark, averaged over three seeds. The combination of SwiGLU, RMS Layer Normalization, and RoPE yields the best performance. Although the absolute improvement is modest, it is stable across runs. We further confirm this trend for larger-scale pre-training with one million synthetic samples in Appendix~\ref{sec:appendix-arch-refine}.

\subsection{Layer by Layer, Epoch by Epoch}
\label{sec:layer-by-layer}

\cite{skean2025layerbylayer} showed that intermediate layers of large language models encode rich representations and can even outperform final layers on downstream tasks. This observation has since motivated a broader line of work investigating layer-wise representations, including their use for cross-domain transfer. In particular, \citet{roschmann2025tivit} demonstrated that intermediate layers of vision transformers can be highly effective for time series classification when signals are converted to images. More recently, \citet{auer2025tirexclassification} showed that forecasting-oriented foundation models can also be leveraged for classification by aggregating representations across layers.

\setlength{\intextsep}{5pt}%
\setlength{\columnsep}{10pt}%
\begin{wrapfigure}[17]{r}{0.45\textwidth}
\includegraphics[width=\linewidth, clip=true, trim=0 0 1.3cm 1.4cm]{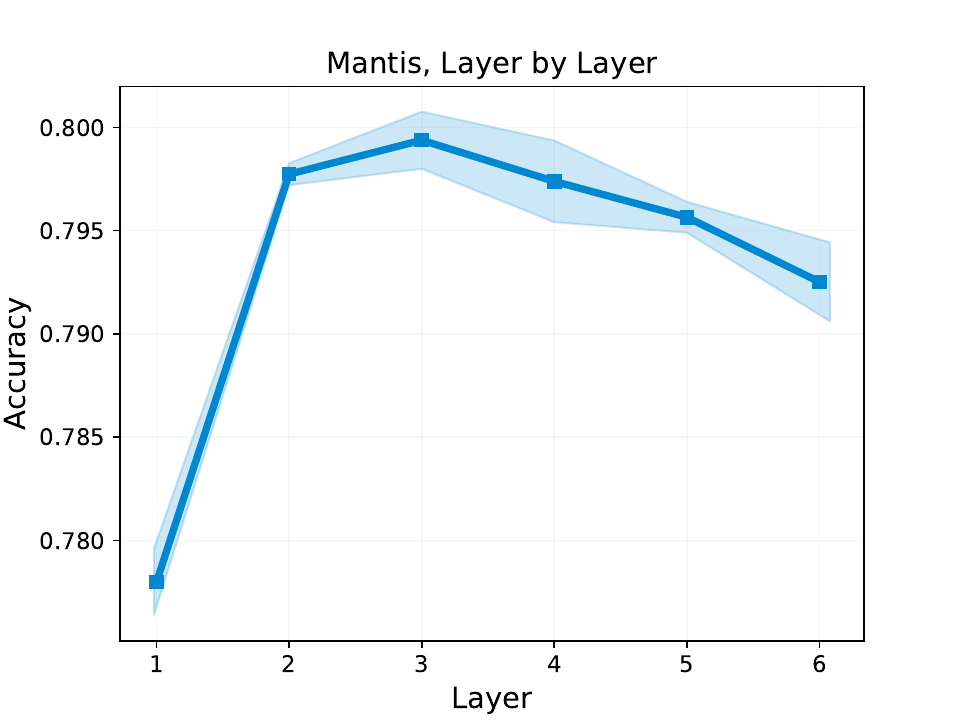}
\caption{Mantis, layer by layer performance.}
\label{fig:mantis-layer-by-layer}
\end{wrapfigure}

In this section, we extend the layer-by-layer analysis to Mantis. Figure~\ref{fig:mantis-layer-by-layer} reports the zero-shot feature extraction performance of each of Transformer layers. Despite Mantis being a relatively compact model, we observe the same phenomenon: intermediate layers yield stronger representations than the final layer, with the third layer achieving the highest accuracy. This observation motivated us a deeper investigation into the evolution of layer-wise representations during pre-training.

\begin{figure}[t]
    \centering
    \includegraphics[width=\textwidth]{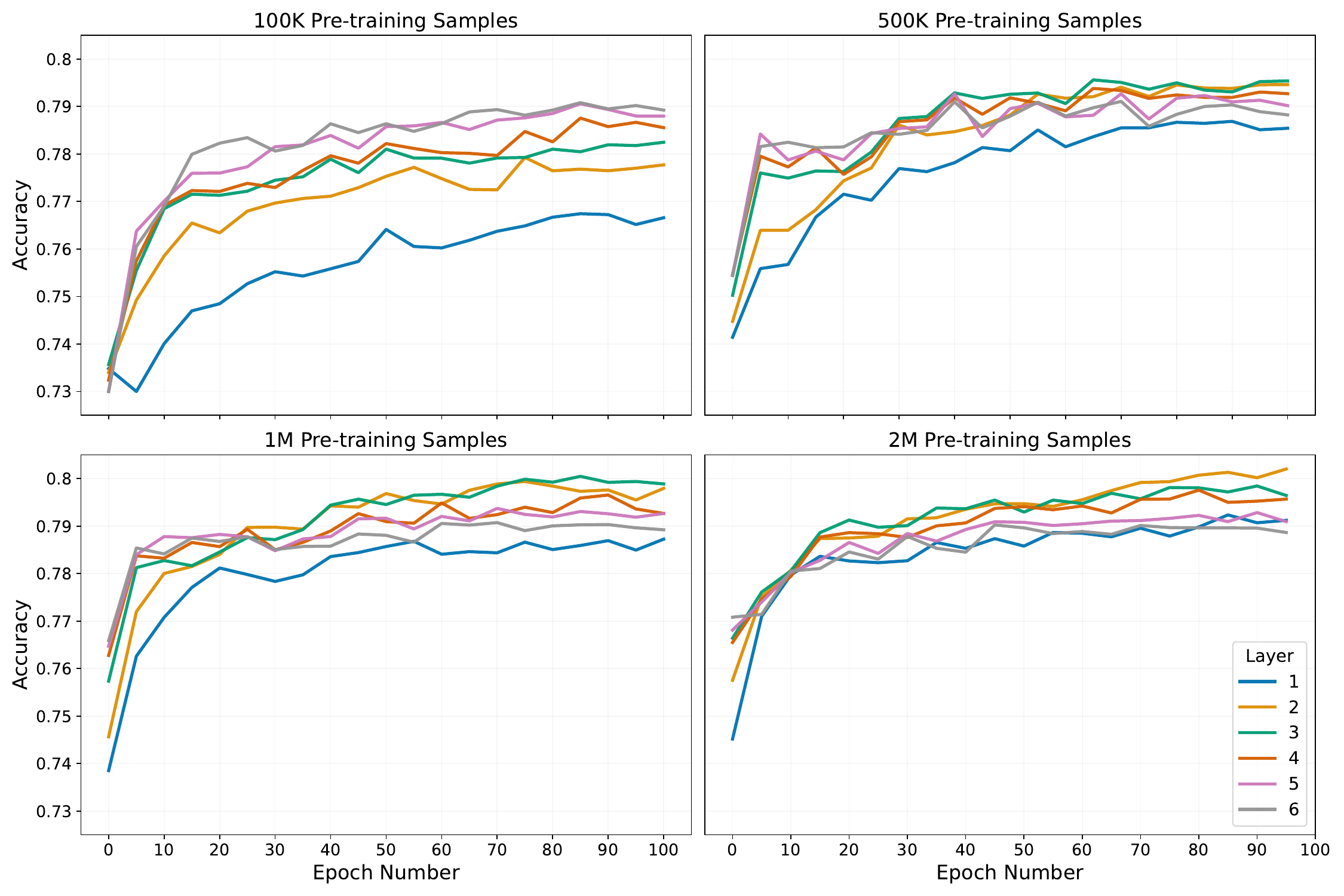}
    \caption{Downstream performance evolution: layer by layer, epoch by epoch.}
    \label{fig:layer-by-layer-epoch-by-epoch}
\end{figure}

In Figure~\ref{fig:layer-by-layer-epoch-by-epoch}, we track the UCR classification performance of all six Transformer layers across pre-training epochs. We additionally vary the size of the synthetic pre-training dataset from 100{,}000 to 2{,}000{,}000 samples. Several notable patterns emerge. First, the relative improvement of intermediate layers appears to be closely tied to the total number of parameter updates. In our training setup, each epoch processes the full dataset, so increasing the dataset size directly increases the number of updates. With 100K samples, the final layer remains the strongest throughout training. In contrast, for larger datasets, intermediate layers steadily improve over time and eventually surpass the final layer. To test whether this behavior is indeed driven by the number of updates rather than dataset size per se, we pre-train Mantis on 100K samples for 1{,}000 epochs, matching the number of updates used for 1M samples over 100 epochs. The corresponding results, reported in Appendix~\ref{sec:layer-by-layer-appendix} (Figure \ref{fig:layer-by-layer-100k-1000epochs}), confirm that one of the intermediate layers eventually becomes the most performant even in this setting.

However, increasing the number of epochs for the 100K dataset does not lead to further gains in overall performance. Instead, performance follows a double-descent–like behavior and converges to the same level achieved during the first 100 epochs. This leads to our second key observation. While the final-layer performance remains largely unchanged (or may even degrade) as the pre-training dataset grows, the \emph{best-performing intermediate layer} improves consistently with increased data scale. In other words, \textbf{intermediate representations unlock the scaling benefits of pre-training}. By selecting the most informative layer, Mantis exhibits a clear and monotonic improvement in performance as the number of pre-training samples increases.

\begin{figure}[t]
    \centering
    \includegraphics[width=\textwidth]{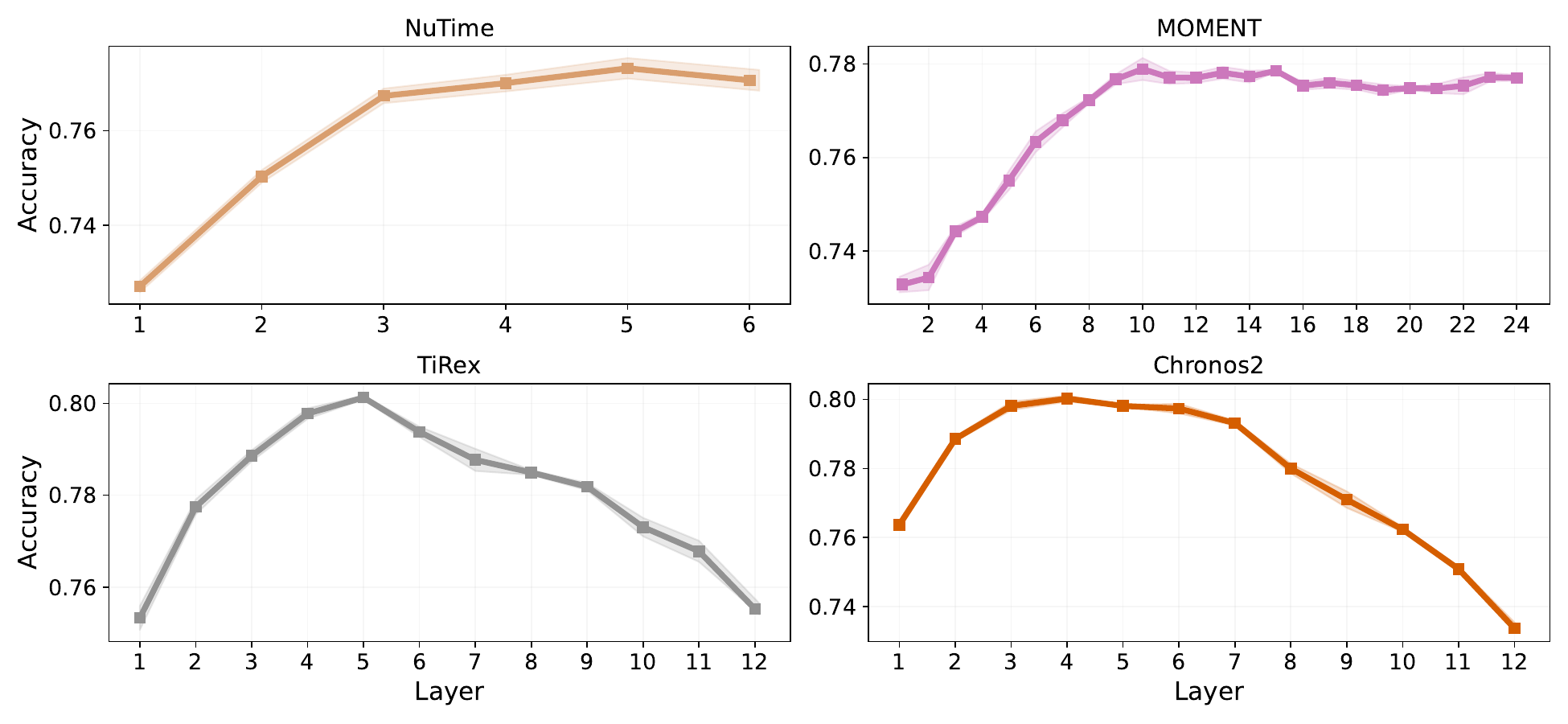}
    \caption{Layer-by-Layer for other models.}
    \label{fig:layer-by-layer-sota}
\end{figure}

For a more comprehensive comparison, we conduct a layer-by-layer analysis of several other foundation models, including NuTime \citep{lin2024nutime}, MOMENT \citep{goswami2024moment}, TiRex \citep{auer2025tirexclassification}, and Chronos2 \citep{ansari2025chronos2}, which are formally introduced in Section~\ref{sec:exp-setup}. Figure~\ref{fig:layer-by-layer-sota} summarizes the results. TiRex and Chronos2 exhibit strong discriminative power in early layers, while their final layers appear to be more specialized for forecasting. MOMENT’s performance plateaus after approximately the 10th layer, suggesting that the model could be significantly compressed by truncating its final layers. NuTime benefits the least from intermediate representations, with its best-performing layer located immediately before the final one.

\textbf{Final pre-training. }
Based on these findings, we perform the final pre-training using 2 million synthetic time series generated by CauKer for 200 epochs. In the remainder of the paper, we refer to the original Mantis architecture trained under this protocol as \textbf{Mantis+}. The variant incorporating all architectural refinements—convolution kernel size set to 41, Transformer head dimension set to 32, and the upgraded Transformer design—is referred to as \textbf{MantisV2}. Layer by layer, epoch by epoch performance evolution curves for both models are provided in Appendix~\ref{sec:layer-by-layer-appendix} (Figure \ref{fig:final-pre-training}).

Finally, we emphasize that the advantage of intermediate-layer representations is specific to the zero-shot feature extraction setting, where the encoder is kept frozen. In the fine-tuning regime, retaining and updating all layers remains preferable; additional details are provided in Appendix~\ref{sec:appendix-fine-tuning}.

\subsection{Aggregation of Output Tokens}
\label{sec:output-token}
When extracting representations from Transformer models, the choice of how to aggregate output tokens remains an open question. In discriminative Transformers such as BERT \citep{devlin2019bert} and ViT \citep{dosovitskiy2021vit}, it is standard practice to discard all outputs except the classification (CLS) token. After multiple self-attention layers, this token is expected to aggregate information from all other tokens, whose representations are typically considered redundant and are often ignored to avoid increasing the dimensionality of the final embedding.

\begin{figure}[h]
    \centering
    \begin{subfigure}{0.54\linewidth}
        \centering
        \includegraphics[height=5.5cm, clip=true, trim=0.4cm 0 0 0]{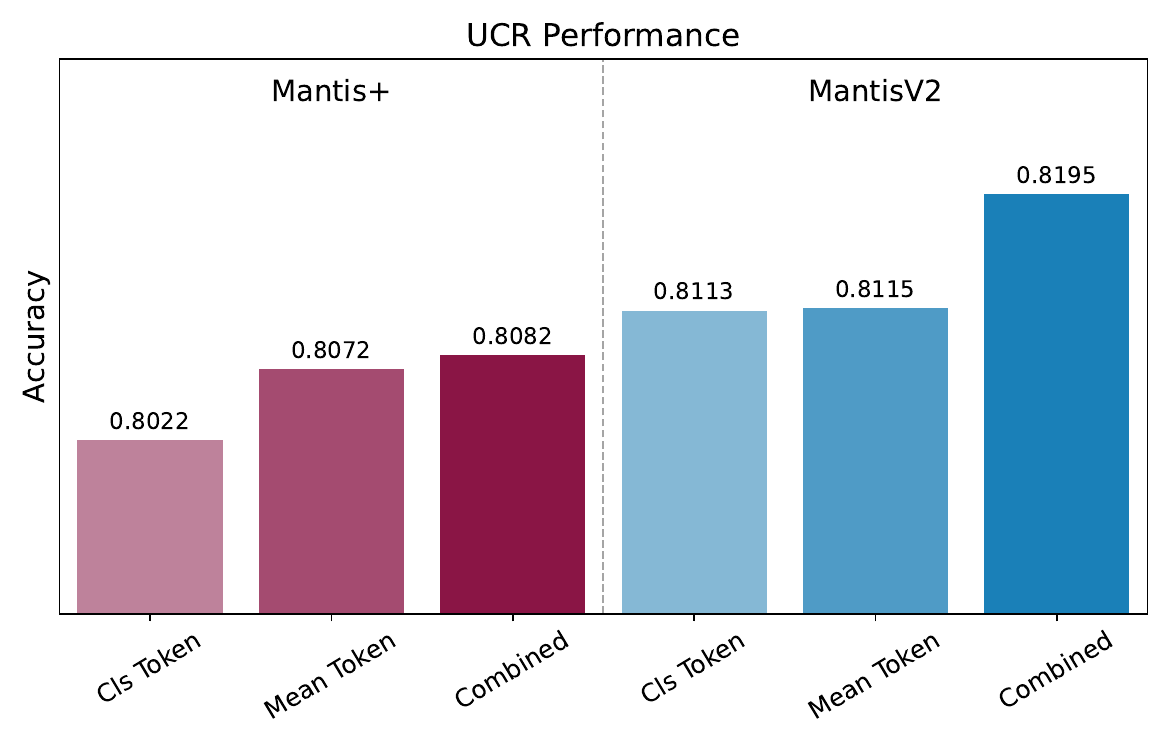}
        \caption{Output-token aggregation.}
        \label{fig:ablation-token}
    \end{subfigure}
    \begin{subfigure}{0.44\linewidth}
        \includegraphics[height=5.5cm, clip=true, trim=0cm 0.4cm 0 0cm]{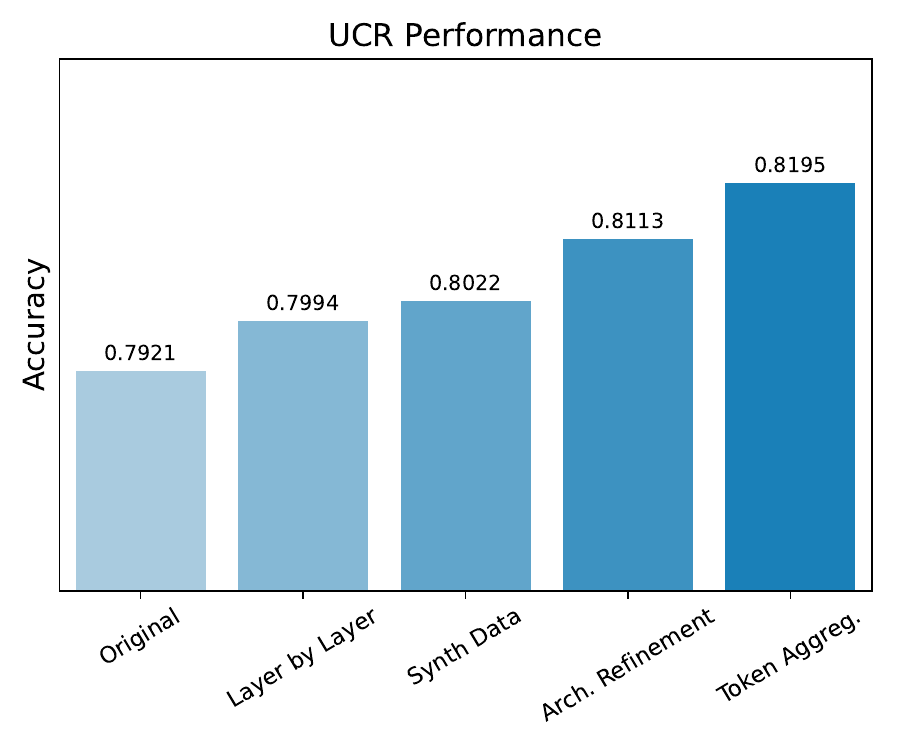}
        \caption{All improvements.}
        \label{fig:improvement}
    \end{subfigure}
    \caption{Ablation study on output-token aggregation and the summary of all improvements made in this section.}
    % \label{fig:ablation-architecture}
\end{figure}

However, this assumption may no longer hold when considering intermediate layer representations. Recently, \citet{roschmann2025tivit} showed that, for vision transformers applied to time series classification, averaging representations across tokens can yield more discriminative embeddings than relying solely on the classification token. Inspired by this observation, we investigate alternative token aggregation strategies for Mantis. Specifically, we evaluate three approaches to generate the final embedding: (a) using only the classification token (as in the original Mantis),
(b) computing the mean of all tokens except the classification token, (c) concatenating the classification token with the mean of the remaining tokens.
% \begin{itemize}
% \item using only the classification token (as in the original Mantis),
% \item computing the mean of all tokens except the classification token,
% \item concatenating the classification token with the mean of the remaining tokens.
% \end{itemize}
% We consider three ways to generate embeddings: (a) keeping only classification token as we did for the first version of Mantis, (b) computing the mean over all tokens except the classification one, (c) combining these two by concatenation. The experimental results on UCR are displayed in Figure \ref{fig:ablation-token}. Interestingly, we find that conclusion is similar as for TiViT: other tokens contain important information and should not be simply discarded. What we have additionally found is that concatenating the classification and mean tokens further boost the performance (by $0.1\pct$ for Mantis+ and by $0.8\pct$ for MantisV2). From now on, we use this concatenation strategy as the default one, which increases the final embedding size of our models from 256 to 512.
Figure~\ref{fig:ablation-token} reports the corresponding results on the UCR benchmark. Consistent with the findings of \citet{roschmann2025tivit}, we observe that non-classification tokens encode complementary and discriminative information and should not be discarded. Moreover, concatenating the classification token with the mean token consistently yields the best performance, improving accuracy by $0.1\pct$ for Mantis+ and by $0.8\pct$ for MantisV2. Based on these results, we adopt the concatenation strategy as the default aggregation method, increasing the embedding dimensionality from 256 to 512.

We note that the effectiveness of this combined aggregation strategy may depend on the depth of the Transformer, as the number of layers determines how effectively the classification token aggregates information from the remaining tokens. In Appendix~\ref{sec:appendix-output-token}, we show that the combined strategy does not help to strengthen representations from the last transformer layer. Nevertheless, in Appendix~\ref{sec:appendix-fine-tuning}, we show that concatenation of the classification and mean tokens improves the performance also for fine-tuned truncated models.

\section{Experimental Results, Part I: Towards Strongest Feature Extractor.}
In this section, we compare Mantis with the State-of-the-Art (SOTA) time series classification foundation modeling. As in the previous section, we follow the zero-shot feature extraction setup, i.e., extract features keeping the encode frozen and use them together with the training labels to learn a classifier.

\subsection{Setup}
\label{sec:exp-setup}

We compare Mantis with the following baselines, whose implementation details can be found in Appendix~\ref{sec:appendix-exp-setup}:
\begin{itemize}
    \item Catch22~\citep{lubba2019catch22} is a set of statistical features powerful for time series classification. In this paper, we have significantly improved this baseline by incorporating also patched statistics. We name this modification as \textbf{Catch22+} and refer to Appendix \ref{sec:catch22plus} for more details and the corresponding ablation study.
    \item \textbf{TabPFN}~\citep{hollmann2022tabpfn} and \textbf{TabICL}~\citep{qu2025tabicl} are tabular foundation models. In this case, we treat each timestamp as a feature and ignore the sequential nature of data.
    % \item \textbf{TabICL}~\citep{qu2025tabicl} is another tabular foundation model.
    \item \textbf{MOMENT}~\citep{goswami2024moment} is a T5-based auto-encoder~\citep{raffel2020exploring} pre-trained in a self-supervised way on 1.13 billion samples. Following Section \ref{sec:layer-by-layer}, we use the output of its 10th transformer layer, which shrinks the model size from 341.2 to 161.4 million parameters.
    \item \textbf{TiRex}~\citep{auer2025tirex} a time series forecasting foundation model based on the xLSTM architecture~\citep{beck2024xlstm}. We use its 5th layer for the output, reducing the model size from 35.3 to 16.5 million parameters.
    \item \textbf{Chronos2}~\citep{ansari2025chronos2} is a transformer-based time series forecasting foundation model. We use the output of its 4th layer, shrinking the model size from 119.5 to 44 million parameters.
    \item \textbf{TiViT-H} is the approach proposed by~\citep{roschmann2025tivit} to leverage a pre-trained vision model for time series classification. We follow their recommendations and use the 14th layer of CLIP ViT-H leading to 276.6 million parameters. \textbf{TiConvNext} leverages the pre-trained CLIP ConvNext in the same spirit. We use its 15th layer which results in 180.3 million parameters.
    \item \textbf{NuTime}~\citep{lin2024nutime} is a classification TSFM based on the BYOL self-distillation pre-training \citep{grill2020byol}. Their pre-training dataset is similar to the one used for pre-training of the first version of Mantis (1.89 million time series examples). We use its 5th layer to get embeddings, which reduces the model size from 2.4 to 2 million parameters.

\end{itemize}
We have decided to remove UniTS~\citep{gao2024units} and GPT4TS~\citep{zhou2023onefitsall} from consideration due to their low performance (please see \citeauthor{feofanov2025mantis}, \citeyear{feofanov2025mantis} for more details).

\textbf{Datasets. } We use the following datasets to evaluate the performance of Mantis and compare it to other methods: 
\begin{itemize}
    \item \textbf{UCR}~\citep{dau2019ucr} is our default benchmark that consists of 128 univariate datasets.
    \item \textbf{UEA}~\citep{bagnall2018uea} benchmark that has 30 multivariate datasets. We exclude 3 datasets due to small test size or small sequence length, and subsample one dataset to ease computations (see more details in Appendix \ref{sec:appendix-exp-setup}). We further tag the remaining 27 datasets by UEA-27.
    \item We additionally gathered a collection of datasets for Human Activity Recognition (\textbf{HAR}) that is one of the most spread applications of time series classification models \citep{chen2025comodo,li2025zara}. We consider 7 datasets, where data come from either an inertial measurement unit (Ego4D, HHAR, UCI-HAR) or motion capture (EMOPain, MP8, MP50). For HHAR dataset, we follow \citep{gagnon2022woods} and consider two splits: in-distribution (ID) and out-of-distribution (OOD). Please refer Appendix \ref{sec:har-data-appendix} for more details.

    \item In a similar fashion, we test Mantis also on electroencephalogram (\textbf{EEG}) data that represent recordings of the brain's electrical activity. We consider different tasks including sleep stage prediction (CAP, SEDFx), brain–computer interface (BCI) control tasks (FingerMovements, PCL, SelfRegulationSCP), seizure detection (Epilepsy-EEG), blink-type classification (Blink). We take FingerMovements and SelfRegulationSCP from UEA, the other datasets we describe in Appendix \ref{sec:eeg-data-appendix}.

\end{itemize}

\subsection{Experimental Results}
\label{sec:sota-exp-res-rf}

\begin{figure}[ht!]
    \centering
    \includegraphics[width=0.75\linewidth]{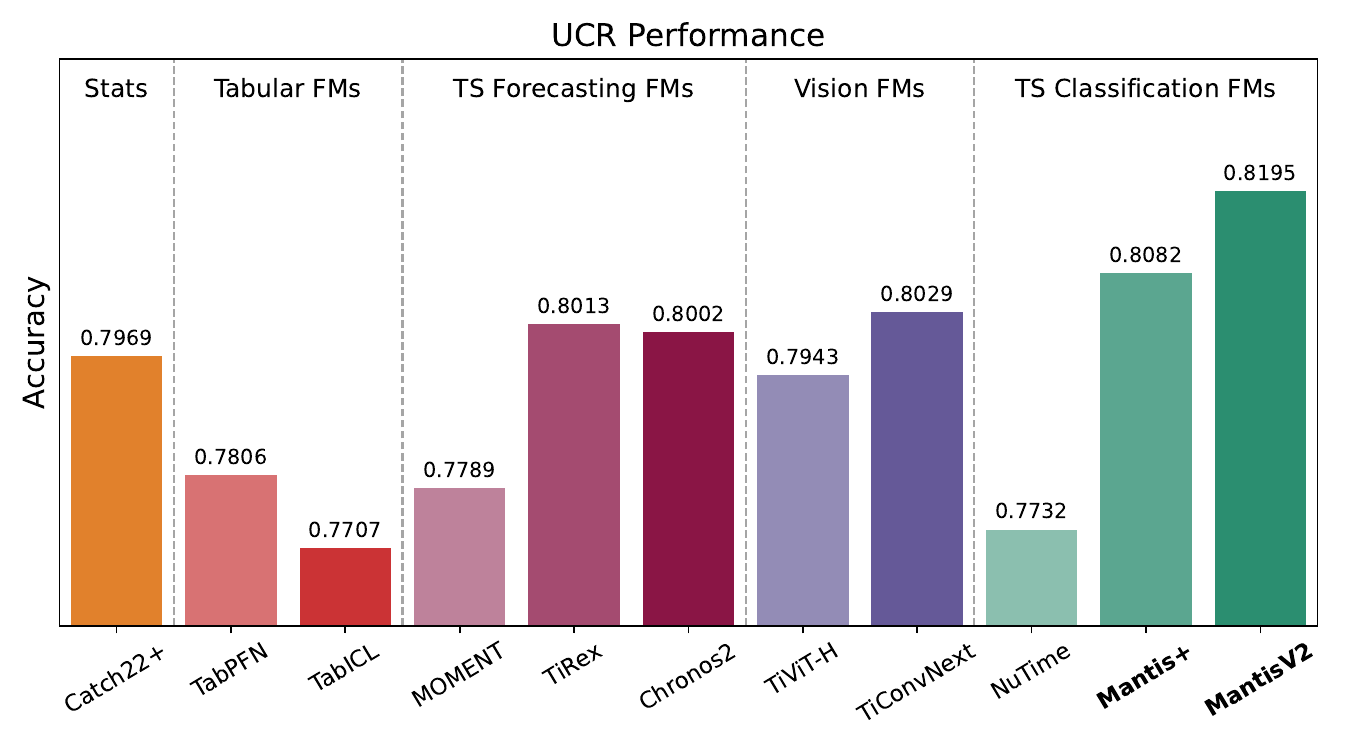}
    \caption{Classification accuracy of different methods in average of 128 univariate UCR datasets.}
    \label{fig:ucr-sota-rf}
\end{figure}

\textbf{UCR. } Figure \ref{fig:ucr-sota-rf} displays the performance of the considered models in average over 128 univariate datasets, while the complete table with results is deferred to Appendix \ref{sec:complete-results} (Table \ref{tab:sota-ucr-res-rf-a} and \ref{tab:sota-ucr-res-rf-b}). We can see that MantisV2 significantly outperforms the other models with Mantis+ being the second best model. Modern forecasting models (TiRex, Chronos2) and adapted vision models (TiViT-H, TiConvNext) are very competitive establishing strong baselines. Interestingly, Catch22+, which is a set of statistical features, also establishes a good baseline, even outperforming foundation models such as NuTime and MOMENT. Tabular foundation models are competitive as well: although their average performance is not very high, their win rate is similar to MantisV2 (see Appendix \ref{sec:complete-results}). This also indicates that a considerable portion of datasets in UCR are nearly tabular, so their sequential nature can simply be ignored.

% Human Activity Recognition (HAR) is one of the most spread applications of time series classification models \citep{chen2025comodo,li2025zara}. In this section, we study Mantis performance specifically for this domain. We consider 7 datasets, where data come from either an inertial measurement unit (Ego4D, HHAR, UCI-HAR) or motion capture (EMOPain, MP8, MP50). For HHAR dataset, we follow \citep{gagnon2022woods} and consider two splits: in-distribution (ID) and out-of-distribution (OOD). Please refer Appendix \cite{TODO} for more details.

\begin{figure}[ht!]
    \centering
    \includegraphics[width=0.75\linewidth]{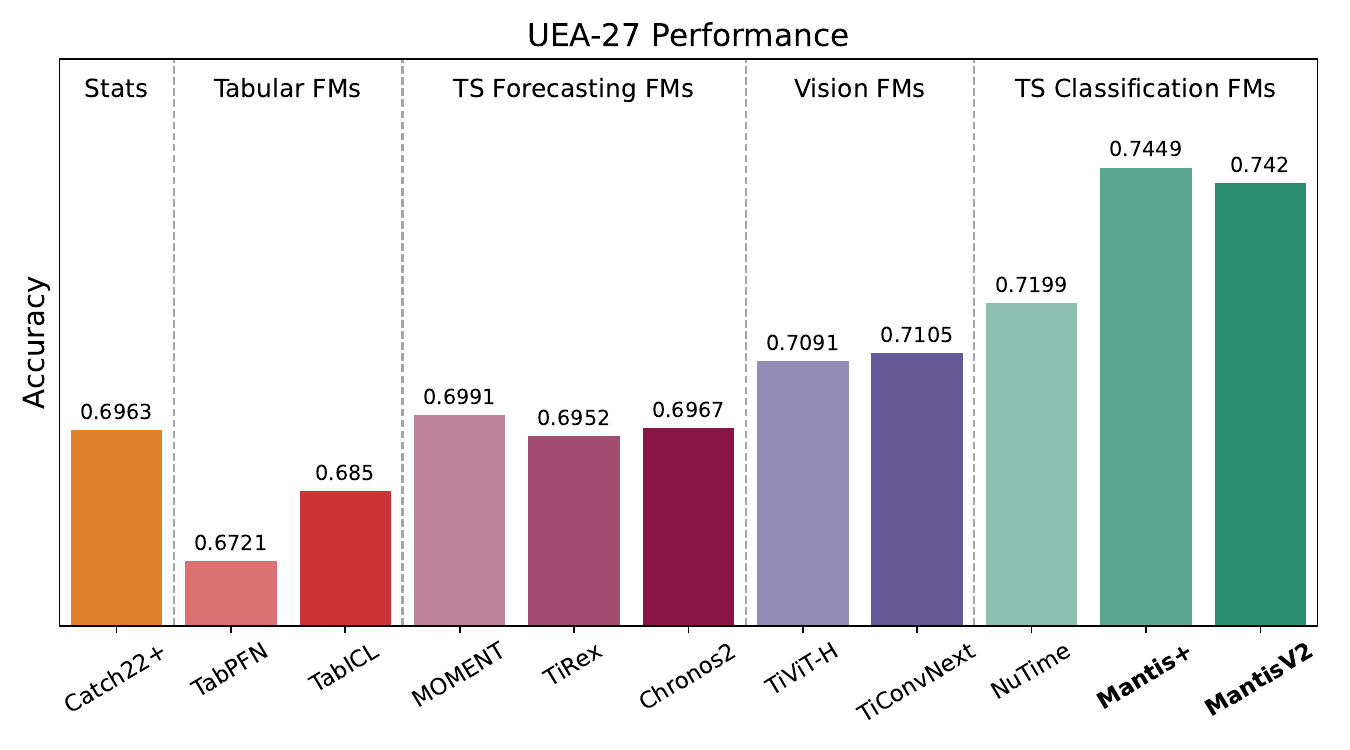}
    \caption{Classification accuracy of different methods in average of 27 multivariate UEA datasets.}
    \label{fig:uea-sota-rf}
\end{figure}

\textbf{UEA-27. } The average performance of the considered models in average over 27 multivariate datasets is illustrated in Figure \ref{fig:uea-sota-rf} while the complete results are deferred to Appendix \ref{sec:complete-results} (Table \ref{tab:uea-sota-rf}). One can see that both Mantis+ and MantisV2 significantly outperform the other models for more than 2\%. Interestingly the model ranking is quite different in this benchmark. For example, NuTime, being one of the worst on UCR, is the third best model this time. Since its hidden dimension is the smallest among the foundation models (128, to be exact), we hypothesize that NuTime may be less sensitive to the curse of dimensionality as the total number of deep features grows linearly with the number of channels.
% MO Modern forecasting models (TiRex, Chronos2) and adapted vision models (TiViT-H, TiConvNext) are very competitive establishing strong baselines. Interestingly, Catch22+, which is a set of statistical features, also establishes a good baseline, even outperforming foundation models such as NuTime and MOMENT. Tabular foundation models are competitive as well: although their average performance is not very high, their win rate is similar to MantisV2 (see Appendix \ref{sec:complete-results}). This also indicates that a considerable portion of datasets in UCR are nearly tabular, so their sequential nature can simply be ignored.

\begin{table}[ht]
    \centering
    \caption{Classification accuracy of different methods on a collection of HAR datasets. Boldface highlights the best method per row while NaN indicates that the model have not passed the scale.}
    \label{tab:exp-res-har-rf}
    {\scalebox{0.6}{
    \begin{tabular}{l|lllllllllll}
    \toprule
                                   & Catch22+ & TabPFN & TabICL & MOMENT & TiRex & Chronos2 & TiViT-H & TiConvNext & NuTime & Mantis+ & MantisV2\\
    \midrule
    Ego4D & $0.4397_{\pm 0.002}$ & \texttt{NaN} & \texttt{NaN} & $0.4068_{\pm 0.001}$ & $0.5037_{\pm 0.001}$ & $0.4742_{\pm 0.0}$ & 0.1912 & 0.1907 & $0.5108_{\pm 0.0}$ & \textbf{0.5273}$_{\pm 0.0}$ & $0.5258_{\pm 0.001}$\\
    EMOPain & $0.8826_{\pm 0.006}$ & 0.7831 & 0.7831 & $0.8469_{\pm 0.002}$ & $0.7915_{\pm 0.003}$ & $0.8075_{\pm 0.004}$ & $0.83_{\pm 0.002}$ & $0.8225_{\pm 0.003}$ & $0.8901_{\pm 0.005}$ & \textbf{0.8939}$_{\pm 0.002}$ & $0.8798_{\pm 0.009}$\\
    HHAR-ID & $0.9738_{\pm 0.001}$ & 0.8938 & 0.9073 & $0.9299_{\pm 0.0}$ & $0.9481_{\pm 0.0}$ & $0.9475_{\pm 0.002}$ & $0.9338_{\pm 0.001}$ & $0.9461_{\pm 0.001}$ & $0.9808_{\pm 0.001}$ & $0.9835_{\pm 0.001}$ & \textbf{0.9845}$_{\pm 0.001}$\\
    HHAR-OOD & $0.4808_{\pm 0.008}$ & 0.5311 & 0.5441 & $0.3081_{\pm 0.001}$ & $0.5135_{\pm 0.014}$ & $0.4174_{\pm 0.008}$ & $0.3748_{\pm 0.007}$ & $0.4031_{\pm 0.01}$ & $0.56_{\pm 0.005}$ & $0.5624_{\pm 0.016}$ & \textbf{0.5822}$_{\pm 0.005}$\\
    MP8 & $0.572_{\pm 0.008}$ & 0.6235 & 0.6185 & $0.6392_{\pm 0.011}$ & $0.5994_{\pm 0.016}$ & $0.5966_{\pm 0.006}$ & $0.6022_{\pm 0.009}$ & $0.5966_{\pm 0.003}$ & $0.6504_{\pm 0.008}$ & $0.6403_{\pm 0.006}$ & \textbf{0.6857}$_{\pm 0.012}$\\
    MP50 & $0.3602_{\pm 0.016}$ & 0.5664 & 0.5042 & \textbf{0.7434}$_{\pm 0.017}$ & $0.6644_{\pm 0.01}$ & $0.6639_{\pm 0.007}$ & $0.6908_{\pm 0.008}$ & $0.6801_{\pm 0.008}$ & $0.6913_{\pm 0.012}$ & $0.7227_{\pm 0.008}$ & $0.7345_{\pm 0.007}$\\
    UCI-HAR & $0.8273_{\pm 0.003}$ & 0.809 & 0.8157 & $0.8782_{\pm 0.001}$ & $0.8744_{\pm 0.002}$ & $0.8873_{\pm 0.002}$ & $0.8936_{\pm 0.001}$ & $0.8918_{\pm 0.001}$ & $0.8842_{\pm 0.001}$ & $0.8911_{\pm 0.0}$ & \textbf{0.9013}$_{\pm 0.002}$\\
    \midrule
    Avg & 0.6481 & \texttt{NaN} & \texttt{NaN} & 0.6789 & 0.6993 & 0.6849 & 0.6452 & 0.6473 & 0.7382 & 0.7459 & \textbf{0.7562}\\
    Avg UCR HAR & 0.7564 & 0.7479 & 0.7285 & 0.7572 & 0.7687 & 0.7702 & 0.7726 & 0.772 & 0.756 & 0.7867 & \textbf{0.8007}\\
    Avg UEA HAR & 0.8138 & 0.7591 & 0.764 & 0.8431 & 0.846 & 0.8492 & 0.8535 & 0.8488 & 0.8539 & \textbf{0.8796} & 0.8739\\
    \bottomrule
    \end{tabular}
    }}
\end{table}

\textbf{HAR. } Table \ref{tab:exp-res-har-rf} displays the experimental results for human activity recognition tasks. We also include the average performance over 19 UCR and 9 UEA datasets that are related to HAR. We can see that the gap between MantisV2/Mantis+ and the other models is quite noticeable, indicating high relevance of the proposed models to this application domain. Similarly to UEA, NuTime performs quite well, while other models show less robust results.

% \subsection{Application to Electroencephalogram (EEG) data}

% In a similar fashion, we also test Mantis for classification of EEG signals that are recordings of the brain's electrical activity. We consider different tasks including sleep stage prediction (CAP, SEDFx), 
% % motor imagery (PCL), movement intention prediction (FingerMovements), slow cortical potential self-regulation (SelfRegulationSCP),
% brain–computer interface (BCI) control tasks (FingerMovements, PCL, SelfRegulationSCP), seizure detection (Epilepsy-EEG), blink-type classification (Blink). We take FingerMovements and SelfRegulationSCP from UEA, the other datasets we describe in Appendix \cite{TODO}.

\vspace{0.2cm}
\begin{table}[ht]
    \centering
    \caption{Classification accuracy of different methods on a collection of EEG datasets. Boldface highlights the best method per row while \texttt{NaN} indicates that the model have not passed the scale.}
    \label{tab:exp-res-eeg-rf}
    {\scalebox{0.6}{
    \begin{tabular}{l|lllllllllll}
    \toprule
                                   & Catch22+ & TabPFN & TabICL & MOMENT & TiRex & Chronos2 & TiViT-H & TiConvNext & NuTime & Mantis+ & MantisV2\\
    \midrule
    Blink & $0.9963_{\pm 0.001}$ & 0.9178 & 0.8978 & $0.9674_{\pm 0.003}$ & \textbf{0.997}$_{\pm 0.001}$ & $0.9733_{\pm 0.01}$ & $0.9852_{\pm 0.006}$ & $0.98_{\pm 0.002}$ & $0.66_{\pm 0.006}$ & $0.9778_{\pm 0.002}$ & $0.9956_{\pm 0.002}$\\
    CAP-ID & $0.751_{\pm 0.001}$ & \texttt{NaN} & \texttt{NaN} & $0.7374_{\pm 0.002}$ & $0.8104_{\pm 0.002}$ & $0.8179_{\pm 0.001}$ & $0.7983_{\pm 0.001}$ & $0.8126_{\pm 0.001}$ & $0.8044_{\pm 0.0}$ & \textbf{0.8189}$_{\pm 0.001}$ & $0.8152_{\pm 0.001}$\\
    CAP-OOD & $0.71_{\pm 0.003}$ & \texttt{NaN} & \texttt{NaN} & $0.7408_{\pm 0.001}$ & $0.7821_{\pm 0.001}$ & $0.7857_{\pm 0.001}$ & $0.7781_{\pm 0.001}$ & $0.784_{\pm 0.001}$ & $0.767_{\pm 0.002}$ & \textbf{0.791}$_{\pm 0.001}$ & $0.7859_{\pm 0.0}$\\
    Epilepsy-EEG & $0.9507_{\pm 0.002}$ & 0.9496 & 0.9447 & $0.952_{\pm 0.001}$ & \textbf{0.9614}$_{\pm 0.001}$ & $0.95_{\pm 0.001}$ & $0.9548_{\pm 0.0}$ & $0.9495_{\pm 0.001}$ & $0.9234_{\pm 0.002}$ & $0.9518_{\pm 0.001}$ & $0.9558_{\pm 0.001}$\\
    FingerMovements & $0.4967_{\pm 0.064}$ & 0.5 & 0.49 & $0.53_{\pm 0.02}$ & $0.52_{\pm 0.04}$ & $0.5233_{\pm 0.021}$ & $0.5367_{\pm 0.045}$ & $0.5333_{\pm 0.059}$ & $0.5267_{\pm 0.015}$ & $0.51_{\pm 0.026}$ & \textbf{0.55}$_{\pm 0.01}$\\
    PCL-ID & $0.5241_{\pm 0.003}$ & \texttt{NaN} & \texttt{NaN} & $0.5431_{\pm 0.012}$ & $0.5272_{\pm 0.011}$ & $0.5308_{\pm 0.002}$ & $0.5347_{\pm 0.007}$ & $0.5385_{\pm 0.005}$ & $0.5615_{\pm 0.003}$ & \textbf{0.5764}$_{\pm 0.006}$ & $0.5716_{\pm 0.003}$\\
    PCL-OOD & $0.5025_{\pm 0.004}$ & \texttt{NaN} & \texttt{NaN} & $0.5129_{\pm 0.001}$ & $0.4988_{\pm 0.001}$ & $0.5072_{\pm 0.004}$ & $0.5003_{\pm 0.005}$ & $0.5132_{\pm 0.001}$ & $0.5415_{\pm 0.004}$ & \textbf{0.5427}$_{\pm 0.002}$ & $0.5311_{\pm 0.009}$\\
    SEDFx-ID & $0.7574_{\pm 0.0}$ & \texttt{NaN} & \texttt{NaN} & $0.7516_{\pm 0.001}$ & $0.7904_{\pm 0.0}$ & $0.8_{\pm 0.0}$ & $0.7884_{\pm 0.001}$ & $0.8008_{\pm 0.001}$ & $0.7822_{\pm 0.001}$ & \textbf{0.8066}$_{\pm 0.0}$ & $0.8_{\pm 0.0}$\\
    SEDFx-OOD & $0.7142_{\pm 0.001}$ & \texttt{NaN} & \texttt{NaN} & $0.7244_{\pm 0.001}$ & $0.7714_{\pm 0.0}$ & \textbf{0.7758}$_{\pm 0.0}$ & $0.7709_{\pm 0.0}$ & $0.7755_{\pm 0.001}$ & $0.741_{\pm 0.0}$ & $0.7731_{\pm 0.0}$ & $0.7636_{\pm 0.0}$\\
    SelfRegulationSCP1 & $0.7702_{\pm 0.007}$ & \textbf{0.8942} & 0.8874 & $0.7747_{\pm 0.006}$ & $0.7884_{\pm 0.012}$ & $0.785_{\pm 0.003}$ & $0.7986_{\pm 0.007}$ & $0.7929_{\pm 0.01}$ & $0.7952_{\pm 0.003}$ & $0.7736_{\pm 0.005}$ & $0.8134_{\pm 0.009}$\\
    SelfRegulationSCP2 & $0.4926_{\pm 0.031}$ & 0.4778 & 0.5056 & $0.4907_{\pm 0.023}$ & $0.4963_{\pm 0.049}$ & $0.5167_{\pm 0.006}$ & $0.4907_{\pm 0.033}$ & $0.5056_{\pm 0.011}$ & $0.5074_{\pm 0.018}$ & \textbf{0.5611}$_{\pm 0.02}$ & $0.5167_{\pm 0.006}$\\
    \midrule
    Avg & 0.6969 & \texttt{NaN} & \texttt{NaN} & 0.7023 & 0.7221 & 0.7242 & 0.7215 & 0.726 & 0.6919 & 0.7348 & \textbf{0.7363}\\
    \bottomrule
    \end{tabular}
    }}
\end{table}
\vspace{0.4cm}

\textbf{EEG. } The experimental results for EEG data can be found in Table \ref{tab:exp-res-eeg-rf}. What it concerns big datasets (CAP, PCL, SEDFx), tabular foundation models do not pass the memory and/or time constraints (we set 10 hours as the time deadline for runtime of one method over one dataset). In contrast to the HAR results, the performance gap between Mantis and other models is smaller, and model ranking reminds more the one was on UCR. As EEG classification is notoriously known to be a difficult task, it will be interesting to see if time series foundation models can have an impact in the progress in this field. Recently, \citet{gnassounou2025leveraging} have demonstrated that Mantis is able to outperform CBraMod (a recent EEG-focused foundation model, \citeauthor{wang2024cbramod}, \citeyear{wang2024cbramod}) on sleep stage and motor imagery tasks, giving positive expectations from TSFMs.

\subsection{Complexity Analysis}

In this section, we compare all the models in terms of memory consumption and running time. As a rule of thumb, the memory consumption of a deep learning model correlates with the total number of its parameters. Table \ref{tab:model-sizes} shows the number of parameters before and after layer pruning for each model (see Section \ref{sec:layer-by-layer} for more details). For TabPFN and TabICL it is not possible to perform layer pruning due to the fundamental difference of their approach. As we can see, the four biggest models are TiConvNext, TiViT-H, MOMENT and Chronos2 with more than 100 million parameters. After layer pruning, their sizes are significantly reduced though still being above 100 million parameters fro TiViT-H, TiConvNext and MOMENT. Layer pruning also helps to reduce the size of Mantis+ and MantisV2 to just less than 3 millions, making them even more lightweight and comparable to the smallest model, namely, NuTime.

\begin{table}[ht!]
\caption{Comparison of Model Sizes.}
\label{tab:model-sizes}
{\setlength{\tabcolsep}{2.3pt}
\begin{tabular}{l|cccccccccc}
\toprule
\# of Params.                       & TabPFN & TabICL & MOMENT & TiRex & Chronos2 & TiViT-H & TiConvNext & NuTime & Mantis+ & MantisV2 \\
\midrule
Original      & 7.2M   & 27.1M  & 341.2M & 35.3M & 119.5M   & 630.8M  & 843.4M     & 2.4M   & 8.1M    & 4.2M     \\
After Pruning & -      & -      & 161.4M & 16.5M & 44M      & 276.6M  & 180.3M     & 2M     & 2.9M    & 2.2M   \\
\bottomrule
\end{tabular}
}
\end{table}

\begin{figure}[ht!]
    \centering
    \includegraphics[width=0.7\textwidth]{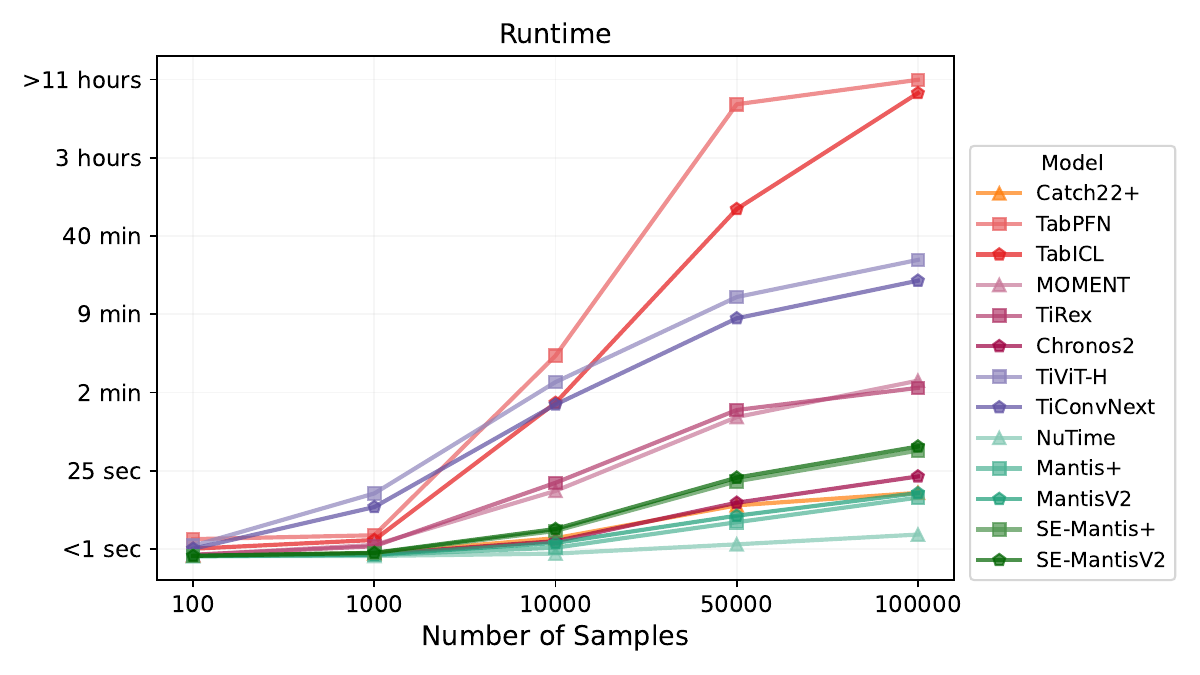}
    \caption{The inference time of the considered models as a function of the number of samples.}
    \label{fig:runtime}
\end{figure}

To measure running time, we compare how much it takes for a forward pass over an entire dataset with batch size equal to 256. We generate univariate synthetic data with length 100, varying the total number of samples within $n\in\{10^2, 10^3, 10^4, 5\cdot10^4, 10^5\}$. As TabPFN and TabICL require both train and test as an input, for them, we use $n/2$ samples as train data and $n/2$ for test, while using batch strategy (256) for test data. The experimental results are displayed in Figure \ref{fig:runtime} and reveal that tabular foundation models are the slowest for large sample sizes taking more than 10 hours for 100,000 examples. They are followed by vision-based models, TiViT-H and TiConvNext, that are slow but, nevertheless, feasible. Next cluster is represented by MOMENT and TiRex. Note that we did not manage to compile TiRex as they suggest, otherwise their model should be faster. The other models, including Chronos2, Mantis+, MantisV2 and NuTime are significantly faster. It is interesting that Mantis performs as fast as Catch22+, which consists in simply calculating pre-determined statistics. Together with Table \ref{tab:model-sizes}, this result demonstrates the wide-applicability of our models including scenarios when computational resources are limited.

\section{Experimental Results, Part II: Closing the Zero-Shot Gap.}

In this section, we extend our experiments by showing how we can push the zero-shot feature extraction performance further. We introduce two simple strategies, namely, self-ensembling and cross-model fusion, and show the importance of the choice of the classification method.
% \subsection{Other Powerful Test-time Strategies}
\subsection{Self-Ensembling (SE)}
\label{sec:self-ensembling}
% (41+(128/32-1))/128=34\%
% (41+(256/32-1))/256=19\%
% (41+(512/32-1))/512=11\%
% (41+(1024/32-1))/1024=7\%
% First, we propose a simple way to improve the performance in test-time by perturbing the input. Our motivation stems from observation that by interpolating it to different sequence lengths we can generate different kind of features.

First, we propose a simple test-time strategy to improve performance by perturbing the input time series. Our motivation stems from the observation that interpolating the same signal to different sequence lengths leads to the extraction of complementary features.

\begin{figure}[ht!]
    \centering
    \includegraphics[width=0.9\linewidth]{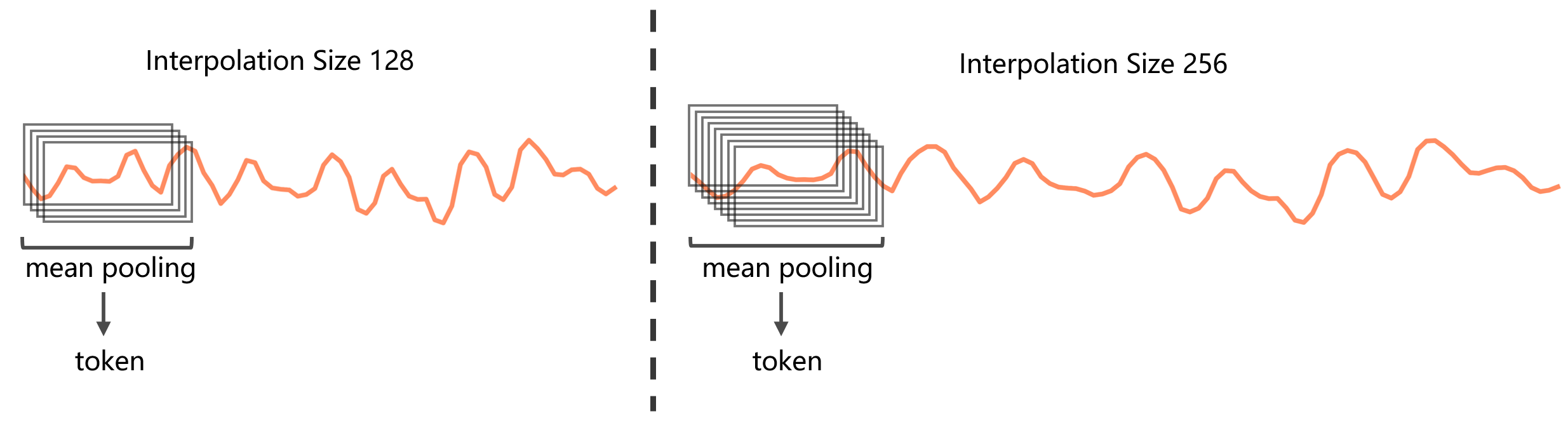}
    \caption{Motivation to look at different interpolation sizes that, in the context of our architecture, imply different degree of overlap between tokens.}
    \label{fig:se-motivation}
\end{figure}

In Figure~\ref{fig:se-motivation}, we illustrate how the convolution–mean-pooling pipeline generates a single token for the same input time series when interpolated to two different lengths. Since the number of tokens in our architecture is fixed, the pooling window size increases with the interpolation length, though not linearly. In the example shown, with a convolution kernel size of 41, the receptive field of a single token for interpolation lengths of 128 and 256 corresponds to approximately $(41 + (128/32 - 1)) / 128 \approx 34\pct$ and $(41 + (256/32 - 1)) / 256 \approx 19\pct$ of the input sequence, respectively. This implies that different interpolation lengths result in different degrees of overlap between tokens. As the sequence length tends to infinity, the receptive field converges to $1/32$, corresponding to the case of non-overlapping tokens. Conversely, decreasing the sequence length increases the degree of overlap between neighboring tokens.

\begin{figure}[t]
    \centering
    \includegraphics[width=0.8\linewidth]{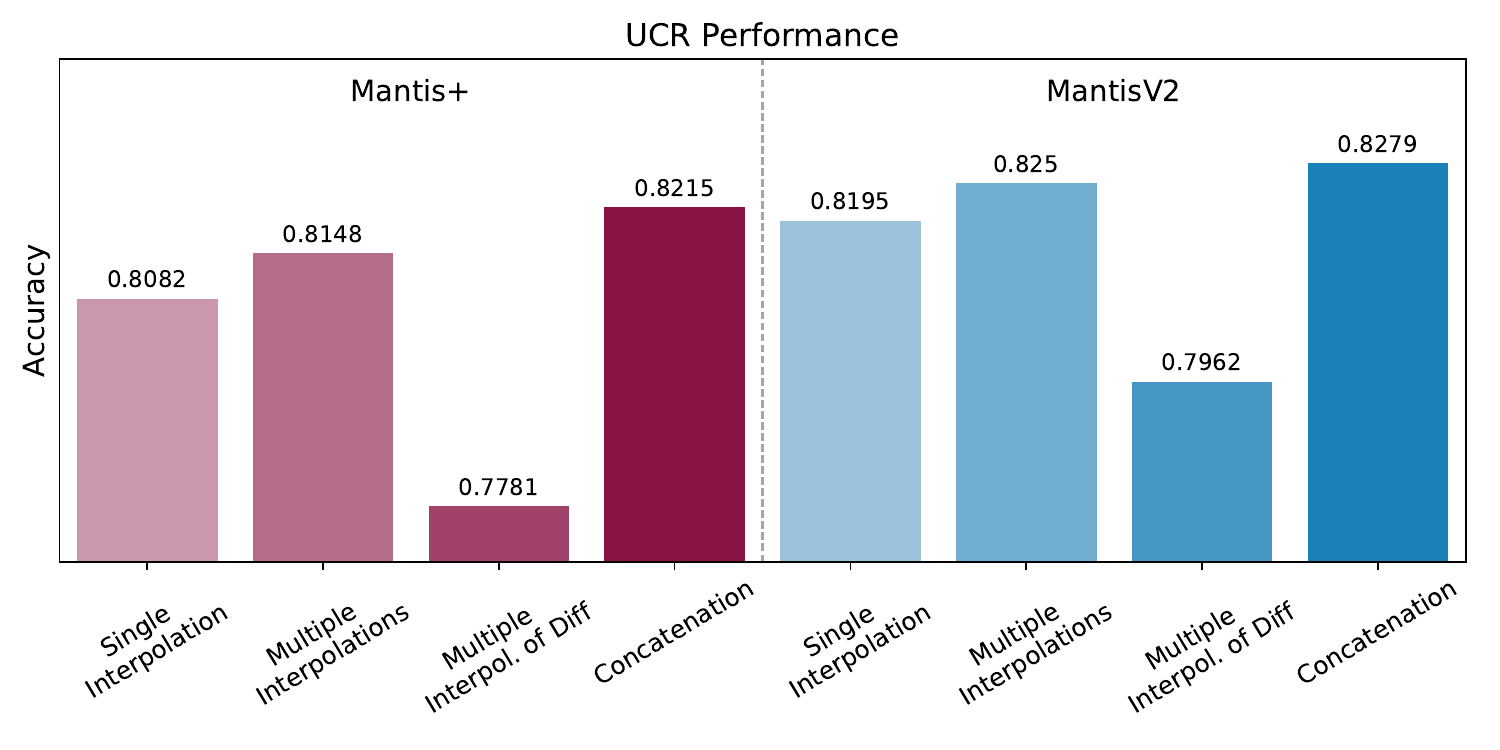}
    \caption{Self-Ensembling Ablation Study.}
    \label{fig:ablation-self-ensembling}
\end{figure}

We therefore can generate multiple interpolated versions of the same input, encode each independently, and concatenate the resulting embeddings. In addition, we explore augmenting the representation with embeddings computed from the first-order difference of the input time series. \citet{auer2025tirex} showed that this strategy improves classification performance for TiRex and other forecasting-oriented foundation models, motivating us to investigate its effectiveness for Mantis.

We compare the following four approaches: 
\begin{itemize}
\item[(a)] interpolating the input time series to a fixed length of 512, as in the original setup.
\item[(b)] generating four interpolated versions of the input at lengths 128, 256, 512, and 1024, encoding each version independently, and concatenating the resulting embeddings.
\item[(c)] applying the same multi-scale interpolation strategy to the first-order difference of the input time series.
\item[(d)] concatenating the embeddings obtained from (b) and (c).
\end{itemize}

Figure~\ref{fig:ablation-self-ensembling} reports the corresponding results for both Mantis+ and MantisV2. Multi-scale interpolation alone improves performance by $0.66\pct$ for Mantis+ and by $0.55\pct$ for MantisV2 on the UCR benchmark. While embeddings derived solely from the first-order difference are individually weak, incorporating them further improves performance by $0.67\pct$ and $0.29\pct$, respectively. This result is particularly noteworthy given that first-order differential features are already included in the Mantis architecture.

In the remainder of the paper, we refer to approach (d) as \textit{Self-Ensembling (SE)}, and denote the resulting models as \textbf{SE-Mantis+} and \textbf{SE-MantisV2}.

\subsection{Cross-model Embedding Fusion}

\citet{roschmann2025tivit} have showed that the embedding of TiViT and Mantis are complimentary to each other, so their combination can improve the performance. We perform a similar experiment combining the embedding of MantisV2 with the embedding of other models. The experimental results are shown in Figure \ref{fig:combination-ucr}. One can observe that model combination is indeed beneficial, even when MantisV2 is combined with statistical features (Catch22+). The smallest improvement comes from MOMENT, NuTime and Mantis+, which may be connected to the lack of complementarity (especially for Mantis+) or low individual performance (particularly for NuTime and MOMENT). The highest performance is achieved when MantisV2 is combined with TiConvNext.

\subsection{The Magic of Logistic Regression}
\label{sec:log-reg}

\begin{figure}[ht!]
\begin{subfigure}{0.48\textwidth}
    \centering
    \includegraphics[width=\linewidth]{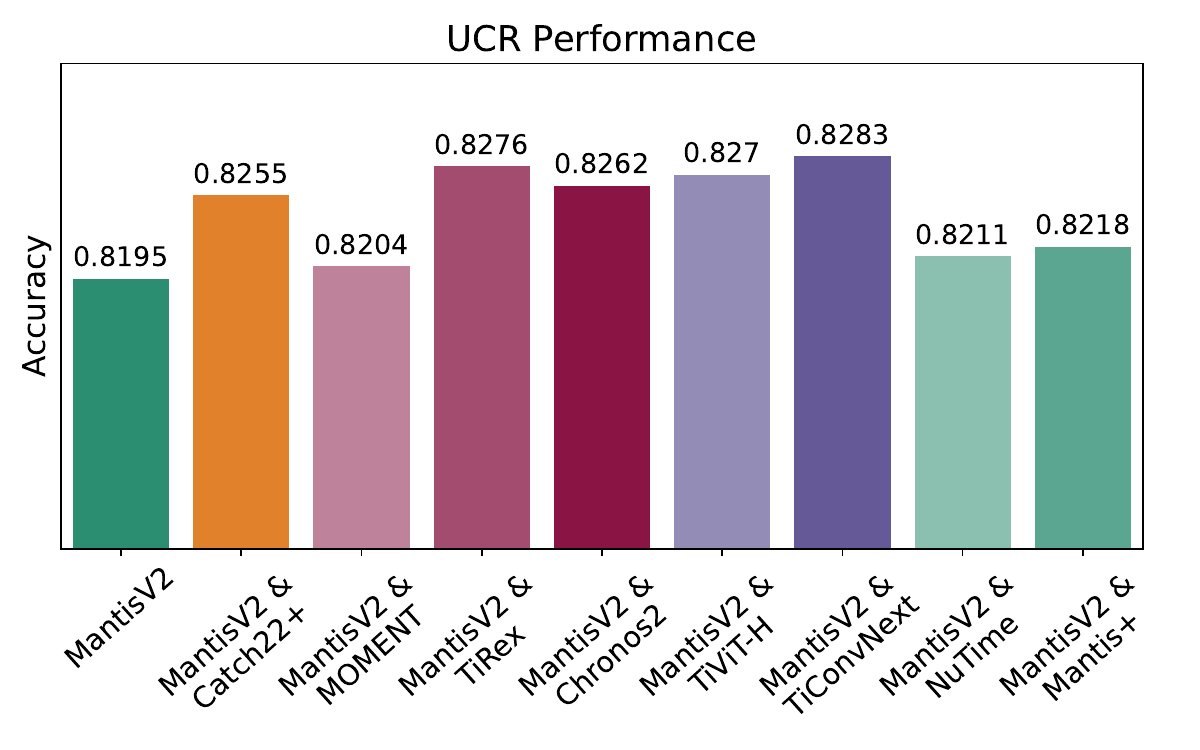}
    \caption{Combination of MantisV2 with other models.}
    \label{fig:combination-ucr}
\end{subfigure}
\begin{subfigure}{0.48\textwidth}
    \centering
    \includegraphics[width=\linewidth]{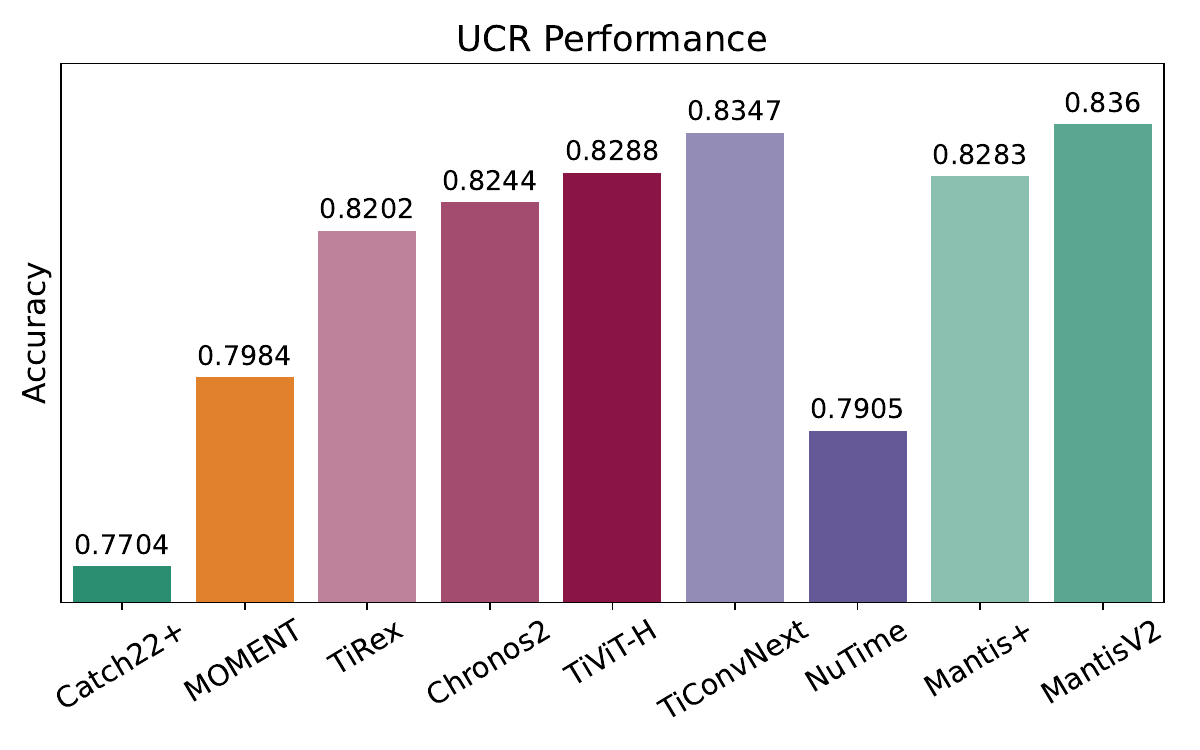}
    \caption{Using logistic regression for classification.}
    \label{fig:logreg-ucr}
\end{subfigure}
\caption{Performance comparison on UCR.}
\end{figure}

In our earlier work \citep{feofanov2025mantis}, we have found that using Random Forest on deep features outperform linear probing (more specifically, layer norm followed by a linear classification layer). However, we have surprisingly found that it is heavily depends on the implementation details. Following \citet{roschmann2025tivit}, we couple the Standard Scaler and Logistic Regression from the scikit-learn package \citep{pedregosa2011scikit}. We set the maximum number of iterations to 500 and leave the other hyperparameters to the default ones, including the solver, which is L-BFGS-B~\citep{zhu1997algorithm,morales2011remark}. The results on UCR are displayed in Figure \ref{fig:logreg-ucr}, while the other results are deferred to Appendix \ref{sec:log-reg-exp-appendix}. From the obtained results we find that using Logistic Regression significantly improves the performance of all methods except Catch22+ on UCR and UEA benchmarks. It is not the case, however, for EEG benchmark, which indicate that there is no free lunch for practitioners. It is interesting to note that Logistic Regression is in practice slower than Random Forest when a sufficient number of CPUs is available. Thus, we have decided to keep the results for both classifiers.

\vspace{0.3cm}

\subsection{Final Comparison}
\label{sec:final-comparison}

Finally, we are now able to explain Figure \ref{fig:teaser-plot} we presented at the beginning of the paper. For Catch22+, we take Random Forest as a classifier, and Logistic Regression for all the other methods. We add self-ensembling results for Mantis+ as well as MantisV2 and select two best cross-model fusion results, namely, combination of MantisV2 with TiViT-H and TiConvNext. In addition, we display the performance of MantisV2 when it is fine-tuned to each dataset (more details can be found in Appendix \ref{sec:appendix-fine-tuning}). The obtained results are very encouraging: using frozen foundation models, we can reach the performance of fine-tuned Mantis, thereby closing the zero-shot gap been present before.

\begin{figure}[ht!]
    \centering
    \includegraphics[width=\linewidth,trim=0.4cm 0 0.2cm 0]{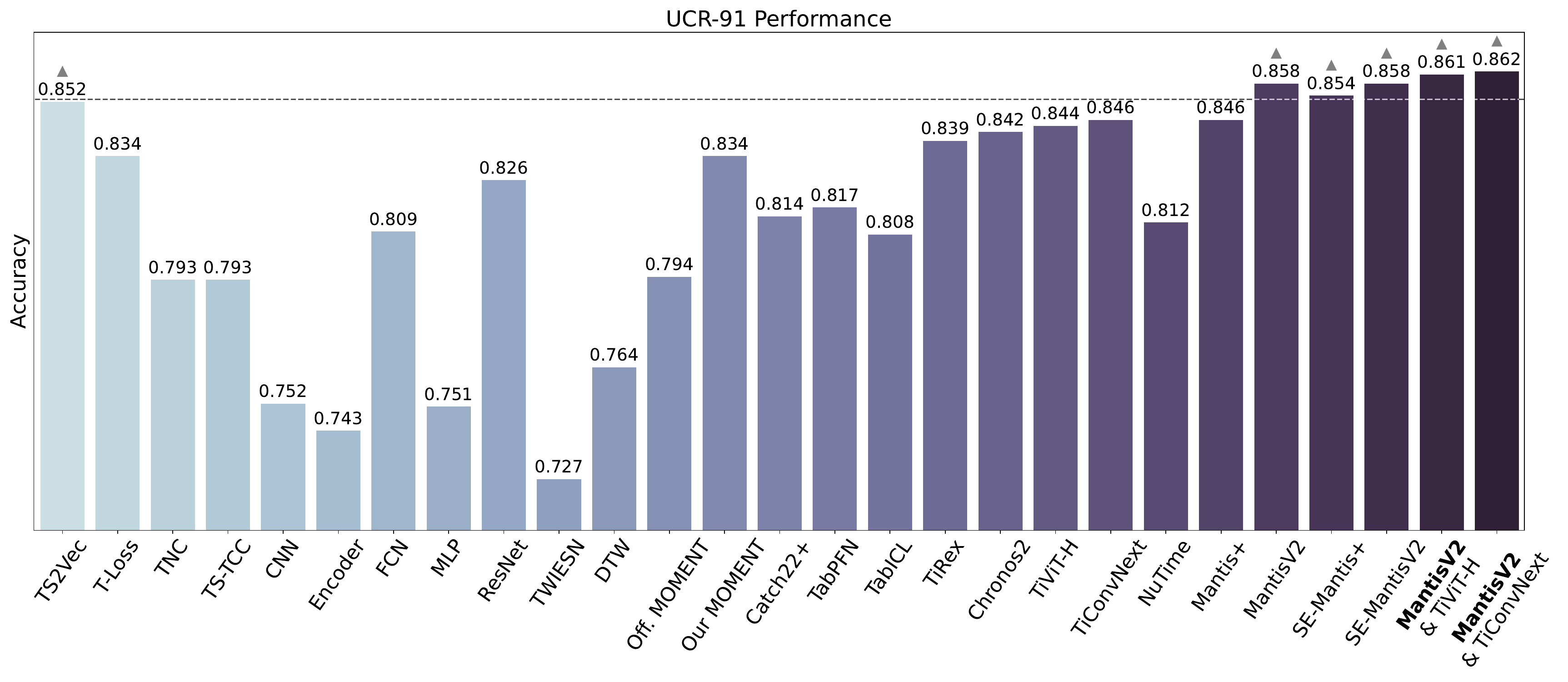}
    \caption{Comparison with more baselines (taken from \citealp{goswami2024moment}) on 91 UCR datasets.}
    \label{fig:more-baselines-comparison}
\end{figure}

In addition, by extracting experimental results from \cite{ismail2019deep} and \cite{goswami2024moment}, we compare Mantis with 11 more baselines including self-supervised methods such as TS2Vec~\citep{yue2022ts2vec}, T-Loss~\citep{franceschi2019tloss}, TNC~\citep{tonekaboni2021unsupervised}, TS-TCC~\citep{eldele2021time}, supervised deep learning methods such as CNN~\citep{zhao2017convolutional}, Encoder~\citep{serra2018towards}, TWIESN~\citep{tanisaro2016time}, FCN, MLP and ResNet~\citep{wang2017time}, and statistical methods such as DTW~\citep{dau2019ucr}. In Figure \ref{fig:more-baselines-comparison}, we illustrate the average performance over 91 UCR datasets for these 11 models, the official MOMENT's pipeline~\citep{goswami2024moment} and all previously considered methods. This result also gives us encouragement as foundation models with frozen encoders (more specifically, MantisV2, SE-Mantis+, SE-MantisV2, MantisV2 \& TiViT-H, MantisV2 \& TiConvNext) finally beat TS2Vec, which performs self-supervised learning for each dataset independently. This indicates that in time series classification it is possible to learn a universal encoder for all problems without sacrificing performance.

\section{Conclusion and Future Work}

In this paper, we presented MantisV2 and Mantis+, a new generation of time series classification foundation models pre-trained exclusively on synthetic data. We proposed an enhanced test-time methodology, achieving a positive outcome: we showed that zero-shot capabilities of a pre-trained model can be significantly improved by leveraging intermediate-layer representations, refined output-token aggregation, and self-ensembling. These findings suggest that scaling laws in time series classification are attainable, and that larger models can be trained whose full potential is unlocked at test time. Beyond scaling, several promising research directions remain open, including multi-modal architectures, zero-shot classification via in-context learning, and joint foundation models for classification and forecasting.

\section*{Acknowledgements}

We would like to thank all the members of Paris Noah's Ark Lab for their constructive comments that helped to improve the manuscript.

\bibliography{references}
\bibliographystyle{apalike}

\appendix
\newpage

\section{Experimental Setup}
\label{sec:appendix-exp-setup}
In this section, we provide more details on the chosen datasets and models. We also experimentally justify Catch22+, a strong baseline that we have proposed.

\subsection{Datasets}
\label{sec:datasets}
We consider four benchmarks. For UCR~\citep{dau2019ucr}, we follow the standard protocol and detail the three other benchmarks below.

\subsubsection{UEA Benchmark}
From the original set~\citep{bagnall2018uea}, we have excluded AtrialFibrillation and StandWalkJump datasets due to their very small test size and PenDigits due to its very small sequence length. For InsectWingbeat dataset, we subsampled 1000 examples from the original training set (which contains 30,000 examples) and 1000 from the original test set (of 20,000 examples) to reduce computational overhead while maintaining sufficient variety in the data for robust model evaluation.
    
\subsubsection{HAR Datasets}
\label{sec:har-data-appendix}

Below we give more details on the human activity recognition datasets used in our experiments, which characteristics can be found in Table \ref{tab:har-datasets}.
\begin{table}[ht!]
    \caption{Characteristics of HAR datasets.}
    \label{tab:har-datasets}
    \centering
    \begin{tabular}{l|cccccc}
        \toprule
        Dataset & \# of Channels & Seq. Length & Train Size & Test Size & Num. classes \\
        \midrule
        Ego4D & 6 & 1000 & 247095 & 66994 & 31 \\
        EMOPain & 30 & 200 & 968 & 355 & 3 \\
        HHAR-ID & 6 & 500 & 8716 & 3419 & 6 \\
        HHAR-OOD & 6 & 500 & 11150 & 2154 & 6 \\
        MP8 & 8 & 161 & 1426 & 595 & 4 \\
        MP50 & 50 & 161 & 1426 & 595 & 4 \\
        UCI-HAR & 3 & 206 & 5881 & 2947 & 6 \\
        \bottomrule
    \end{tabular}
\end{table}

\begin{itemize}
    \item \textit{Ego4D}~\citep{grauman2022ego4d} is a multi-modal dataset for ego-centric human activity recognition. The use of the dataset requires signing the license agreement. We take time series data coming from Inertial Measurement Units (IMU) and pre-process them using the script provided by \citet{chen2025comodo}.
    \item \textit{EMOPain}~\citep{egede2020emopain} is a movement-based chronic pain detection dataset. We downloaded it
    using the script provided by \citet{gao2024units}.
    \item \textit{HHAR}~\citep {stisen2015smart}: we take from the WOODS benchmark~\citep{gagnon2022woods}. For \textit{ID} setting, we merge all domains and make a train-validation-test split in proportion 63.75\%-11.25\%-25\%. For \textit{OOD} setting, we use three domains for training, namely, "nexus4", "s3" and "s3mini", and "lgwatch" domain for test. 
    % gear for validation, 
    \item Military Press (MP, \citeauthor{singh2023fast}, \citeyear{singh2023fast}) is a dataset for human exercise performance classification. We use two versions of the dataset provided by \citet{sungu2025empirical}: \textit{MP8} (8 key body point coordinates) and \textit{MP50} (all 50 coordinates from 25 body parts).
    \item \textit{UCI-HAR}~\citep{anguita2013hardataset} is preprocessed following \citet{NEURIPS2022_194b8dac}.
    \item In the UCR collection, the following datasets are related to HAR: AllGestureWiimoteX,AllGestureWiimoteY, AllGestureWiimoteZ, CricketX, CricketY, CricketZ, GestureMidAirD1, GestureMidAirD2, GestureMidAirD3, GesturePebbleZ1, GesturePebbleZ2, GunPoint, GunPointAgeSpan, GunPointMaleVersusFemale, GunPointOldVersusYoung, PickupGestureWiimoteZ, ShakeGestureWiimoteZ, UWaveGestureLibraryAll, UWaveGestureLibraryX, UWaveGestureLibraryY, UWaveGestureLibraryZ.
    \item In the UEA collection, these datasets are related to HAR: BasicMotions, Cricket, ERing, Epilepsy, Handwriting, Libras, NATOPS, RacketSports, UWaveGestureLibrary.
\end{itemize}

\subsubsection{EEG Datasets}
\label{sec:eeg-data-appendix}

Below we provide more details on the EEG datasets used in our experiments, which main characteristics are given in Table \ref{tab:eeg-datasets}.

\begin{table}[ht!]
    \caption{Characteristics of EEG datasets.}
    \label{tab:eeg-datasets}
    \centering
    \begin{tabular}{l|cccccc}
        \toprule
        Dataset & \# of Channels & Seq. Length & Train Size & Test Size & Num. Classes \\
        \midrule
        Blink & 4 & 510 & 500 & 450 & 2 \\
        CAP-ID & 19 & 3000 & 25748 & 10098 & 6 \\
        CAP-OOD & 19 & 3000 & 27393 & 8265 & 6 \\
        Epilepsy-EEG & 1 & 178 & 60 & 11420  & 2 \\
        FingerMovements & 28 & 50 & 316 & 100 & 2 \\
        PCL-ID & 48 & 750 & 14405 & 5650 & 2\\
        PCL-OOD & 48 & 750 & 9880 & 7800 & 2 \\
        SEDFx-ID & 4 & 3000 & 152178 & 59678 & 6 \\
        SEDFx-OOD & 4 & 3000 & 133746 & 52838 & 6 \\
        SelfRegulationSCP1 & 6 & 896 & 268 & 293 & 2 \\
        SelfRegulationSCP2 & 7 & 1152 & 200 & 180 & 2 \\
        \bottomrule
    \end{tabular}
\end{table}

\begin{itemize}
    \item \textit{Blink}~\citep{chicaiza2021blink} is a dataset for classification of eye blink types. We downloaded it using the script provided by \citet{gao2024units}.
    \item \textit{Epilepsy-EEG}~\citep{andrzejak2001epilepsydataset} is preprocessed following \citet{NEURIPS2022_194b8dac}.
    \item \textit{FingerMovements}, \textit{SelfRegulationSCP1} and \textit{SelfRegulationSCP2} are taken from the UEA archive.
    \item 3 datasets are taken from the WOODS benchmark~\citep{gagnon2022woods}: motor imagery classification with \textit{PCL} \citep{schalk2004bci2000,cho2017eeg,lee2019eeg}, and sleep stage classification with \textit{CAP}~\citep{terzano2002atlas} and \textit{SEDFx}~\citep{kemp2000analysis}. For \textit{ID} setting, we merge all domains and make a train-validation-test split in proportion 63.75\%-11.25\%-25\%. For \textit{OOD} setting, we split by domains. For PCL, we use "Cho2017" for training, "PhysionetMI" for validation, "Lee2019MI" for test. For CAP, we use Machine 0,1,3 for training, Machine 2 for validation, Machine 4 for test. For SEDFx, we use Age 20-60 for training, Age 60-80 for validation, Age 80-100 for test.
\end{itemize}

\subsection{Models}
Below, we give more implementation details on the baselines.
\begin{itemize}
        \item \textit{Catch22.} We use the official python implementation \texttt{pycatch22==0.4.5}.
        \item \textit{NuTime.} We use the pre-trained weights provided by the authors in their \href{https://github.com/chenguolin/NuTime/blob/main/ckpt/checkpoint_bias9.pth}{GitHub} repository, while fixing the hyperparameters of the architecture according to \href{https://github.com/chenguolin/NuTime/blob/main/configs/demo_ft_epilepsy.json}{this} configuration file. In contrast to the original implementation, we do not use their adapter (described in Section 3.4 of their paper) but process all channels independently as for Mantis and MOMENT. This allows us to use NuTime in the zero-shot feature extraction setting as their adapter has to be fine-tuned.
        \item \textit{TabPFN.} We use the default version of \texttt{TabPFNClassifier} from \texttt{tabpfn==2.2.1}. As TabPFN does not support more than 10 classes, we use the \texttt{ManyClassClassifier} wrapper from TabPFN Extensions \citep{ye2025tabpfnextensions}.
        \item \textit{TabICL.} We use the official implementation with default parameters from \texttt{tabicl==0.1.3}.
        \item \textit{MOMENT:} 
        %momentfm-0.1
        We use the MOMENT-large model (\texttt{d\_model}=1024), which pre-trained weights can be found on the corresponding \href{https://huggingface.co/AutonLab/MOMENT-1-large}{HuggingFace} repository. To handle the multi-channel setup, we process every channel independently and concatenate all the embeddings before passing them to the classification head. In the paper, they have considered datasets with a sequence length $\leq 512$ and use zero-padding to fix the input size to 512. At the same time, we have also tried to interpolate sequences to 512 instead, and it did not affect the performance of MOMENT. Thus, we have decided to stick to the latter option as it allows us to evaluate MOMENT for any sequence length. 
        \item \textit{TiRex.} We use the official implementation with default parameters from \texttt{tirex-ts==1.1.1}. 
        \item \textit{Chronos2.} We use the official implementation with default parameters from \texttt{chronos-forecasting==2.0.0}.
        \item \textit{TiViT-H} and \textit{TiConvNext}. We use the official implementation available at \href{https://github.com/ExplainableML/TiViT/tree/main}{GitHub}. For CLIP ViT-H, we use \href{https://huggingface.co/laion/CLIP-ViT-H-14-laion2B-s32B-b79K}{this checkpoint}, while CLIP ConvNext's checkpoint can be found \href{https://huggingface.co/laion/CLIP-convnext_xxlarge-laion2B-s34B-b82K-augreg}{here} \citep{wolf2020transformers, radford2021CLIP, ilharco2021openclip, liu2022convnext, schuhmann2022laionb}. 
    \end{itemize}

\subsection{Catch22+}
\label{sec:catch22plus}
\setlength{\intextsep}{-5pt}%
\setlength{\columnsep}{10pt}%
\begin{wrapfigure}[15]{r}{0.4\textwidth}
\centering    
\vspace{0.1cm}
\includegraphics[width=0.825\linewidth]{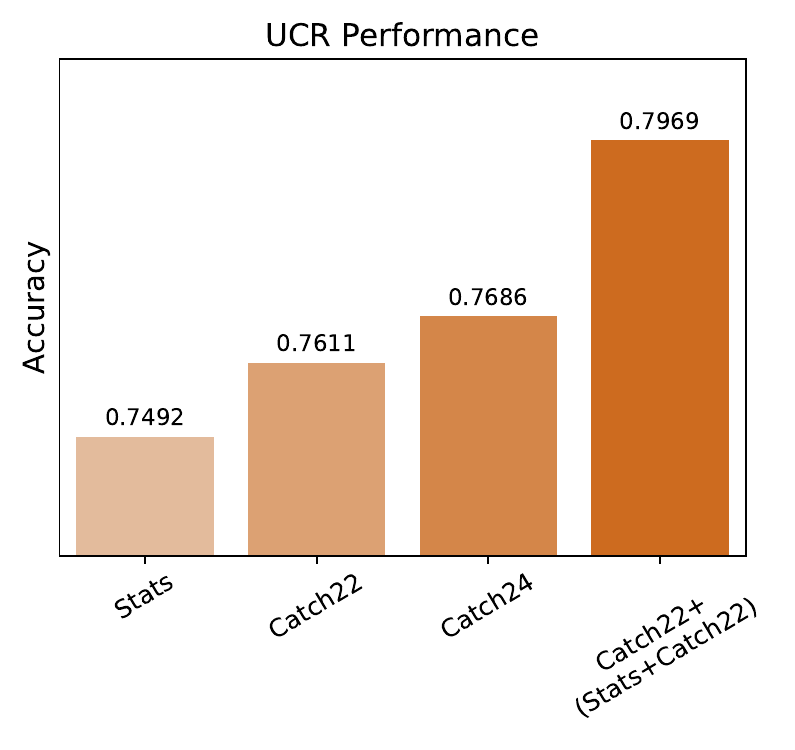}
\caption{Catch22+ ablation study.}
\label{fig:ablation-catch22}
\end{wrapfigure}
Catch22~\citep{lubba2019catch22} is a manually selected set of time series statistics with a high discriminative power. Despite the authors later proposed an improvement by adding the global mean and  standard deviation to the set (so-called Catch24), we have found that it can be further improved by splitting the input into non-overlapping patches (we follow \citet{auer2025tirexclassification} and set their number to 8) and computing the mean and standard deviation for each of them. Using these means and standard deviations as features for classification we further refer by Stats. Figure \ref{fig:ablation-catch22} illustrates the average accuracy of different configurations across UCR datasets. One can notice that the combination of Catch22 with Stats yields a big improvement by 3.58\%, while the improvement from adding global mean and standard deviation (Catch24) is more modest (0.75\%). For this reason, we have considered Catch22+ (concatenation of Catch22 and Stats) as a baseline in the main part of our paper.

\section{Additional Experiments}

In this section, we present our additional experimental results that complement Section \ref{sec:key-improvements}.

\subsection{Layer by Layer, Epoch by Epoch}
\label{sec:layer-by-layer-appendix}

\begin{figure}[ht!]
    \centering
    \includegraphics[width=0.65\linewidth, clip=true, trim=0 0 0 0.3cm]{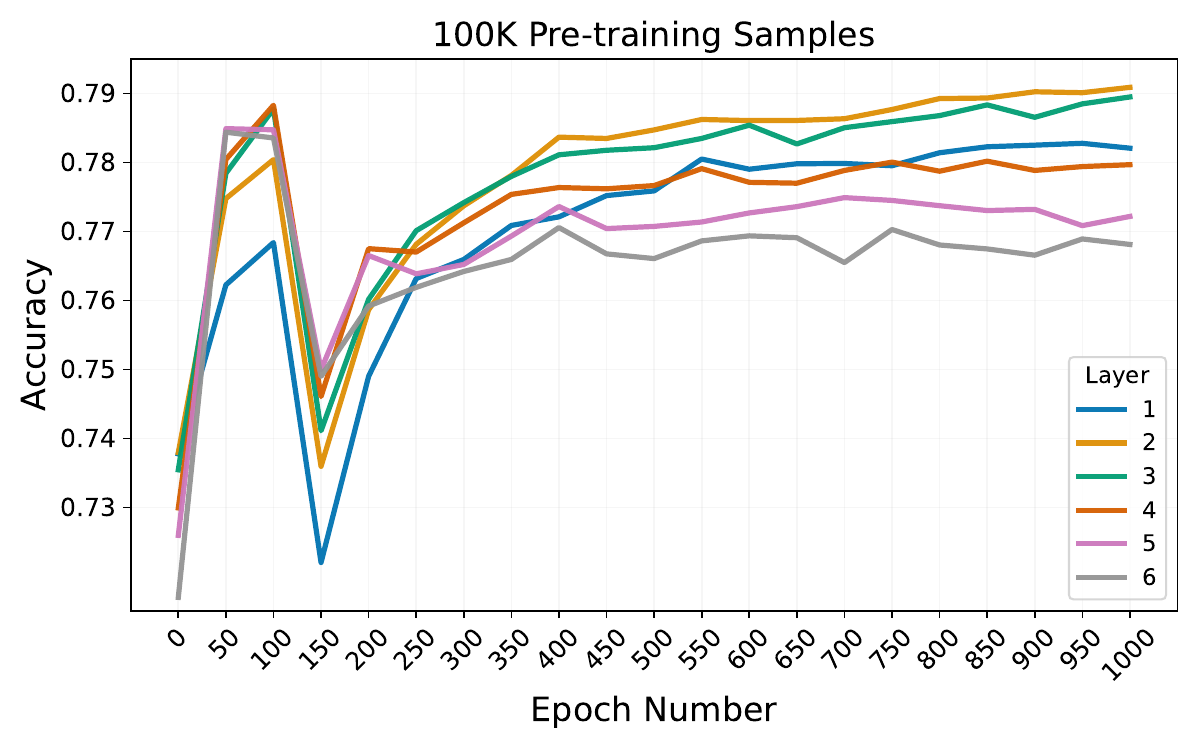}
    \caption{Layer-by-layer downstream performance (UCR) over 1000 epochs with 100K pre-training samples.}
    \label{fig:layer-by-layer-100k-1000epochs}
\end{figure}

Figure \ref{fig:layer-by-layer-100k-1000epochs} illustrates the downstream performance on UCR over 1000 pre-training epochs for all transformer layers of Mantis pre-trained on 100,000 samples. We observe an interesting pattern where the intermediate layers start to dominate in performance when increasing the number of updates. This suggests that the final layer may overfit the contrastive objective, so intermediate layers start to generalize better.

\vspace{0.5cm}
\vspace{0.35cm}

\begin{figure}[t]
    \centering
    \includegraphics[width=\linewidth]{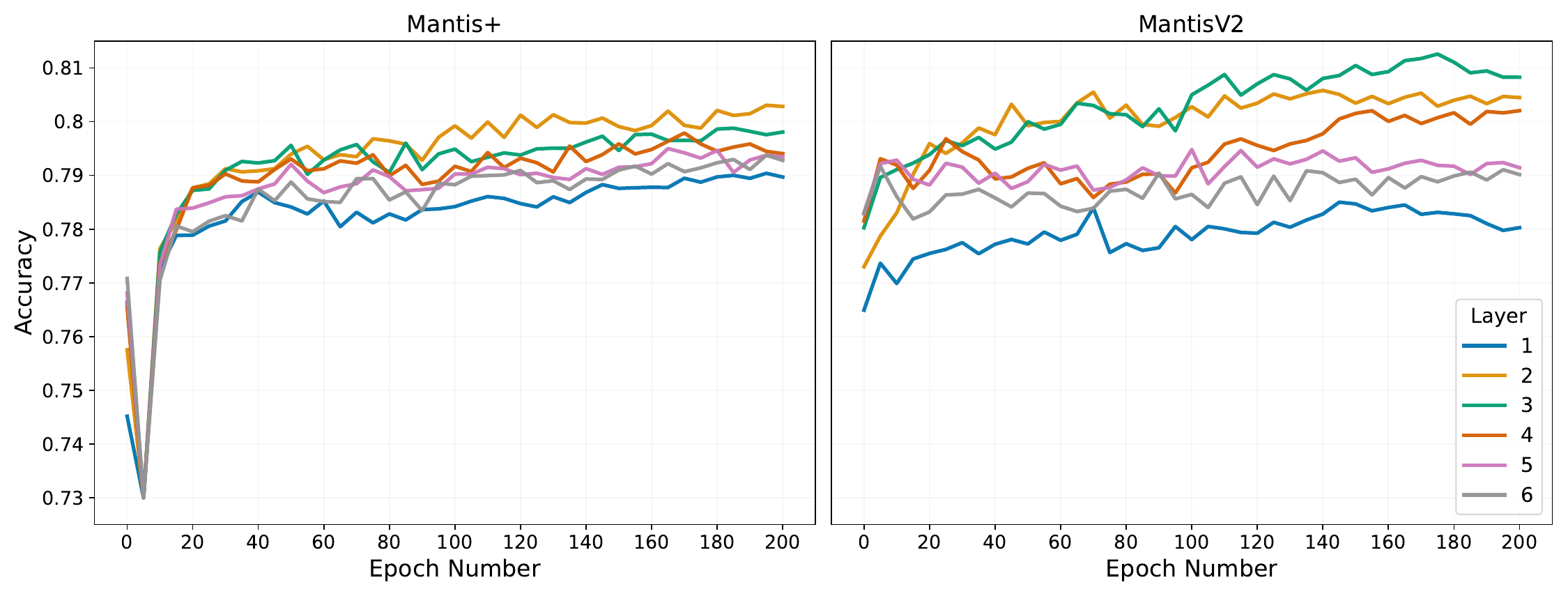}
    \caption{Final pre-training. 200 epochs with 2M pre-training samples.}
    \label{fig:final-pre-training}
\end{figure}

In Figure \ref{fig:final-pre-training}, we show the layer-by-layer downstream performance on UCR over 200 pre-training epochs with 2 million synthetic time series samples. These curves are used to derive final checkpoints for Mantis+ and MantisV2.

\subsection{Aggregation of Output Tokens}
\label{sec:appendix-output-token}

\begin{figure}
    \centering
    \includegraphics[width=0.7\linewidth]{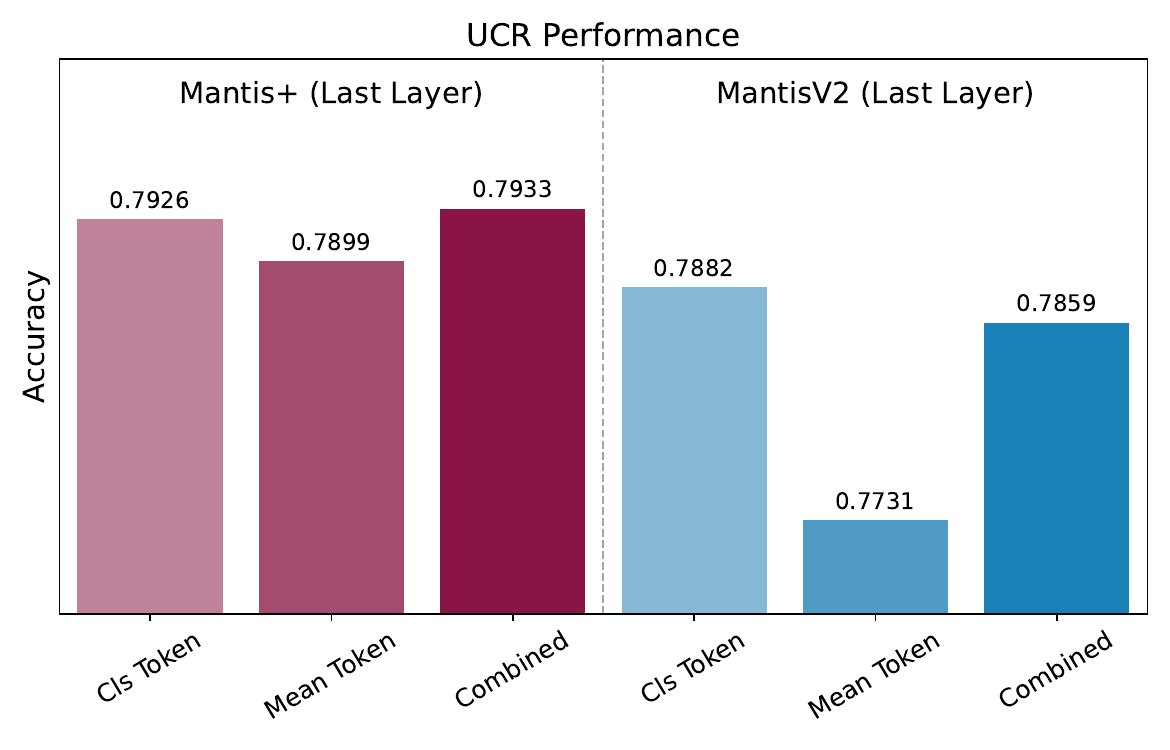}
    \caption{Ablation study on output-token aggregation when the last transformer layer is used.}
    \label{fig:output-token-last-layer}
\end{figure}

In Section \ref{sec:output-token}, we showed that the combined output-token strategy improves the performance when intermediate representations are used. This may be connected to the fact that with a small transformer depth the classification token does not aggregate well the information from the remaining tokens. To support this hypothesis, we compare the same three strategies (cls token, mean token, the combination of both) when the last transformer layer is used. From the results, which are illustrated in Figure \ref{fig:output-token-last-layer}, we can see that the combined strategy is not really beneficial anymore: although it improves slightly the performance of Mantis+, it is not the case for MantisV2.

\subsection{Architecture Refining}
\label{sec:appendix-arch-refine}

In Table \ref{tab:v1-vs-v2-transformer-1m}, we compare performance of Mantis with the first (classical activations / sinusoidal PE) and the second version (new activations / RoPE) of the transformer architecture when pre-trained on 1 million synthetic examples. Note that for 1 million samples, the final layer is not anymore most powerful, so we have to look at all intermediate layers. As we can see, the second version of the transformer yields slightly better results.

\vspace{0.35cm}
\begin{table}[ht!]
    \centering
    \caption{Performance comparison between the first and the second version of transformer architecture (keeping all the other parameters of Mantis architecture fixed as in the first version). Both architectures were pre-trained on 1 million synthetic examples generated by CauKer, while the last epoch accuracy on UCR benchmark is reported.}
    \label{tab:v1-vs-v2-transformer-1m}
    \begin{tabular}{c|cc}
        \toprule
        Layer & \multirow{2}{*}{V1 Transformer} & \multirow{2}{*}{V2 Transformer} \\
        Number & & \\
        \midrule
        1 & 0.7872 & 0.7669 \\
        2 &	0.7979 & 0.7987 \\
        3 &	0.7989 & \textbf{0.7994} \\
        4 &	0.7927 & 0.7986 \\
        5 &	0.7927 & 0.7919 \\ 
        6 &	0.7892 & 0.78  \\
        \bottomrule
    \end{tabular}
\end{table}

\subsection{Fine-tuning}
\label{sec:appendix-fine-tuning}

Following \cite{feofanov2025mantis}, we append a prediction head after an encoder and fine-tune all layers on the training data of a dataset. The prediction head is a layer normalization step followed by a linear layer. We fix a fine-tuning scheme: we minimize the cross-entropy loss for 500 epochs with a fixed batch size equal to 128, using an AdamW optimizer~\citep{loshchilov2017fixing} with a learning rate of $2\cdot10^{-4}$ and a weight decay of 0.05. We report the performance of a model at the last epoch in average over 3 experimental runs. 
% As the prediction head, we use a combination of a layer normalization and a linear layer.

\begin{figure}[ht!]
    \centering
    \begin{subfigure}{0.49\linewidth}
        \centering
        \includegraphics[width=0.9\linewidth]{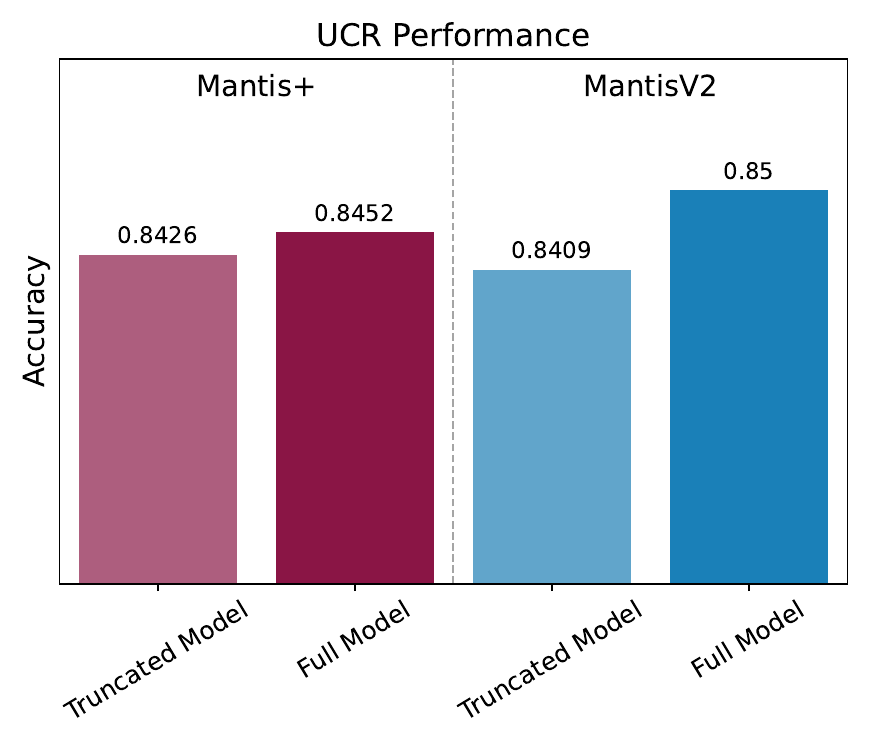}
        \caption{Truncated Model vs Full Model.}
        \label{fig:fine-tuning-layer}
    \end{subfigure}
    \begin{subfigure}{0.49\linewidth}
        \centering
        \includegraphics[width=0.9\linewidth]{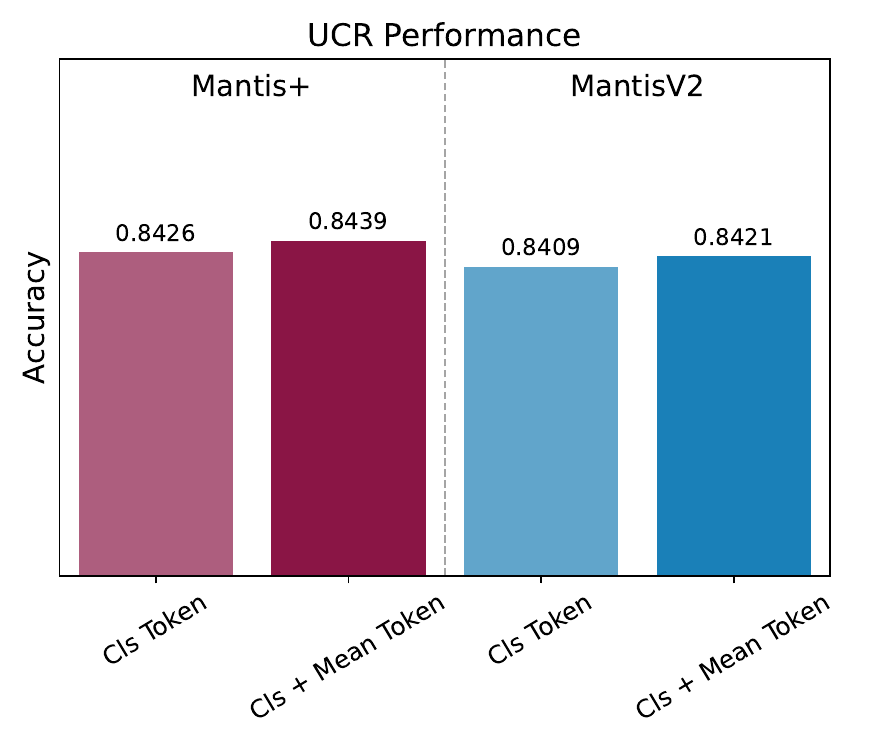}
        \caption{Output-token aggregation strategies for Truncated Model.}
        \label{fig:fine-tuning-token}
    \end{subfigure}
    \caption{Fine-tuning results for different configurations of Mantis+ and MantisV2. "Truncated Model" means that all transformer layers that succeed the intermediate layer with the highest zero-shot performance are removed. "Full Model" refers to keeping all the layers.}
\end{figure}

First, we explore whether it is reasonable to fine-tune a truncated model instead of keeping all the layers, following the success of intermediate layers for zero-shot feature extraction. As we can see in Figure \ref{fig:fine-tuning-layer}, the answer to the question is negative, and keeping the original mode size leads to the superior performance. We may conclude that, despite of their low utility at the end of pre-training, the latter transformer layers remain to be important for fine-tuning as they enrich the capacity of the model to fit the training data.

\vspace{0.35cm}
\vspace{0.4cm}

As a side experiment, we also tested if the output-token aggregation strategy proposed in Section \ref{sec:output-token} improves the fine-tuning performance for the truncated Mantis+ and MantisV2. As it can be seen in Figure \ref{fig:fine-tuning-token}, the impact of incorporating the information from other tokens is positive, though being smaller than for the zero-shot experiment.

\section{Complete Results}
\label{sec:complete-results}

Below we provide the complete tables that display the performance per dataset.

\subsection{Tables for Section \ref{sec:sota-exp-res-rf}}
First, we show results for Section \ref{sec:sota-exp-res-rf}, which performs the model comparison with the random forest as a classifier. Table \ref{tab:sota-ucr-res-rf-a} and Table \ref{tab:sota-ucr-res-rf-b} correspond to the results on UCR, while Table \ref{tab:uea-sota-rf} refers to UEA.

\newcommand{\scalefactor}{0.67}

\vspace{0.3cm}
\begin{table}[H]
    \centering
    \caption{Zero-shot feature extraction results on UCR (Random Forest). First Part.}
    \label{tab:sota-ucr-res-rf-a}
    {\scalebox{0.55}{
    \begin{tabular}{l|lllllllllll}
\toprule
                               & Catch22+ & TabPFN & TabICL & MOMENT & TiRex & Chronos2 & TiViT-H & TiConvNext & NuTime & Mantis+ & MantisV2\\
\midrule
ACSF1 & $0.8233_{\pm 0.021}$ & 0.8 & 0.81 & $0.82_{\pm 0.026}$ & $0.85_{\pm 0.01}$ & $0.8467_{\pm 0.012}$ & $0.85_{\pm 0.0}$ & \textbf{0.8667}$_{\pm 0.006}$ & $0.75_{\pm 0.02}$ & $0.8033_{\pm 0.006}$ & $0.8233_{\pm 0.012}$\\
Adiac & $0.734_{\pm 0.007}$ & 0.8031 & 0.8031 & $0.7894_{\pm 0.003}$ & $0.7852_{\pm 0.011}$ & $0.8107_{\pm 0.009}$ & $0.7059_{\pm 0.007}$ & $0.6547_{\pm 0.018}$ & $0.7357_{\pm 0.006}$ & $0.8321_{\pm 0.006}$ & \textbf{0.8483}$_{\pm 0.003}$\\
AllGestureWiimoteX & $0.5957_{\pm 0.01}$ & 0.6229 & 0.5043 & $0.6138_{\pm 0.002}$ & $0.6495_{\pm 0.012}$ & $0.6038_{\pm 0.005}$ & $0.6029_{\pm 0.011}$ & $0.6676_{\pm 0.002}$ & $0.6505_{\pm 0.005}$ & $0.6414_{\pm 0.012}$ & \textbf{0.6843}$_{\pm 0.006}$\\
AllGestureWiimoteY & $0.6319_{\pm 0.016}$ & 0.6329 & 0.5114 & $0.6624_{\pm 0.004}$ & \textbf{0.7052}$_{\pm 0.019}$ & $0.6419_{\pm 0.002}$ & $0.6514_{\pm 0.01}$ & $0.6805_{\pm 0.011}$ & $0.63_{\pm 0.01}$ & $0.691_{\pm 0.006}$ & $0.691_{\pm 0.004}$\\
AllGestureWiimoteZ & $0.5462_{\pm 0.005}$ & 0.5329 & 0.4529 & $0.5767_{\pm 0.007}$ & $0.6057_{\pm 0.001}$ & $0.6205_{\pm 0.004}$ & $0.6129_{\pm 0.014}$ & $0.6067_{\pm 0.011}$ & $0.6224_{\pm 0.009}$ & $0.6452_{\pm 0.004}$ & \textbf{0.6881}$_{\pm 0.001}$\\
ArrowHead & $0.741_{\pm 0.012}$ & 0.7543 & 0.7429 & $0.7943_{\pm 0.015}$ & $0.7886_{\pm 0.01}$ & $0.779_{\pm 0.037}$ & $0.7638_{\pm 0.012}$ & $0.7714_{\pm 0.006}$ & $0.7733_{\pm 0.023}$ & $0.7714_{\pm 0.006}$ & \textbf{0.8057}$_{\pm 0.015}$\\
BME & \textbf{1.0}$_{\pm 0.0}$ & \textbf{1.0} & 0.98 & $0.9867_{\pm 0.007}$ & $0.96_{\pm 0.007}$ & $0.9667_{\pm 0.007}$ & $0.9778_{\pm 0.017}$ & $0.9933_{\pm 0.007}$ & $0.8467_{\pm 0.035}$ & $0.9667_{\pm 0.007}$ & $0.9956_{\pm 0.004}$\\
Beef & $0.6889_{\pm 0.051}$ & 0.8 & 0.7667 & $0.7889_{\pm 0.019}$ & \textbf{0.8444}$_{\pm 0.019}$ & $0.7333_{\pm 0.033}$ & $0.7556_{\pm 0.019}$ & $0.8222_{\pm 0.019}$ & $0.6667_{\pm 0.067}$ & $0.6333_{\pm 0.033}$ & $0.6889_{\pm 0.019}$\\
BeetleFly & $0.7833_{\pm 0.029}$ & 0.9 & 0.8 & $0.95_{\pm 0.0}$ & $0.9167_{\pm 0.029}$ & $0.8667_{\pm 0.029}$ & $0.9333_{\pm 0.029}$ & $0.95_{\pm 0.0}$ & $0.8_{\pm 0.05}$ & $0.9167_{\pm 0.029}$ & \textbf{0.9833}$_{\pm 0.029}$\\
BirdChicken & $0.85_{\pm 0.0}$ & 0.85 & 0.75 & $0.9833_{\pm 0.029}$ & $0.8833_{\pm 0.029}$ & $0.9_{\pm 0.0}$ & $0.85_{\pm 0.087}$ & \textbf{1.0}$_{\pm 0.0}$ & $0.9667_{\pm 0.029}$ & $0.8833_{\pm 0.029}$ & $0.9_{\pm 0.0}$\\
CBF & $0.9763_{\pm 0.002}$ & 0.9133 & 0.9244 & $0.9678_{\pm 0.005}$ & $0.9915_{\pm 0.003}$ & $0.993_{\pm 0.002}$ & \textbf{0.9978}$_{\pm 0.001}$ & $0.9937_{\pm 0.002}$ & $0.9726_{\pm 0.002}$ & $0.9859_{\pm 0.002}$ & $0.9948_{\pm 0.002}$\\
Car & $0.7556_{\pm 0.025}$ & 0.7833 & \textbf{0.8167} & $0.7556_{\pm 0.025}$ & $0.6667_{\pm 0.017}$ & $0.7389_{\pm 0.025}$ & $0.7389_{\pm 0.019}$ & $0.7222_{\pm 0.035}$ & $0.7667_{\pm 0.029}$ & $0.7944_{\pm 0.01}$ & $0.7556_{\pm 0.025}$\\
Chinatown & $0.9796_{\pm 0.003}$ & \textbf{0.9854} & 0.9796 & $0.9776_{\pm 0.002}$ & $0.9689_{\pm 0.006}$ & $0.9718_{\pm 0.004}$ & $0.9631_{\pm 0.002}$ & $0.8717_{\pm 0.02}$ & $0.9281_{\pm 0.008}$ & $0.9495_{\pm 0.003}$ & $0.9534_{\pm 0.003}$\\
ChlorineConcentration & $0.6682_{\pm 0.001}$ & 0.95 & \textbf{0.9773} & $0.6973_{\pm 0.007}$ & $0.7018_{\pm 0.004}$ & $0.7095_{\pm 0.003}$ & $0.7076_{\pm 0.001}$ & $0.7073_{\pm 0.002}$ & $0.671_{\pm 0.002}$ & $0.6874_{\pm 0.004}$ & $0.7041_{\pm 0.001}$\\
CinCECGTorso & $0.8872_{\pm 0.013}$ & 0.8341 & 0.8225 & $0.6831_{\pm 0.005}$ & \textbf{0.8976}$_{\pm 0.002}$ & $0.8135_{\pm 0.018}$ & $0.8164_{\pm 0.004}$ & $0.8101_{\pm 0.012}$ & $0.7314_{\pm 0.016}$ & $0.7384_{\pm 0.005}$ & $0.8089_{\pm 0.01}$\\
Coffee & $0.9881_{\pm 0.021}$ & 0.9643 & \textbf{1.0} & $0.9286_{\pm 0.0}$ & $0.9881_{\pm 0.021}$ & $0.9286_{\pm 0.0}$ & \textbf{1.0}$_{\pm 0.0}$ & \textbf{1.0}$_{\pm 0.0}$ & $0.9286_{\pm 0.0}$ & $0.9762_{\pm 0.021}$ & \textbf{1.0}$_{\pm 0.0}$\\
Computers & $0.7347_{\pm 0.006}$ & 0.62 & 0.656 & $0.6267_{\pm 0.022}$ & $0.76_{\pm 0.011}$ & $0.7413_{\pm 0.008}$ & $0.7333_{\pm 0.008}$ & $0.7413_{\pm 0.014}$ & \textbf{0.7627}$_{\pm 0.01}$ & $0.74_{\pm 0.004}$ & $0.7213_{\pm 0.012}$\\
CricketX & $0.6974_{\pm 0.01}$ & 0.6667 & 0.6641 & $0.6855_{\pm 0.01}$ & $0.6923_{\pm 0.017}$ & $0.6991_{\pm 0.017}$ & $0.7137_{\pm 0.005}$ & $0.6641_{\pm 0.003}$ & $0.6667_{\pm 0.01}$ & $0.735_{\pm 0.007}$ & \textbf{0.7735}$_{\pm 0.011}$\\
CricketY & $0.6923_{\pm 0.008}$ & 0.7 & 0.6333 & $0.6735_{\pm 0.005}$ & $0.7222_{\pm 0.01}$ & $0.7137_{\pm 0.012}$ & $0.7128_{\pm 0.012}$ & $0.665_{\pm 0.004}$ & $0.6889_{\pm 0.01}$ & $0.7581_{\pm 0.009}$ & \textbf{0.788}$_{\pm 0.02}$\\
CricketZ & $0.7427_{\pm 0.008}$ & 0.6718 & 0.6692 & $0.7487_{\pm 0.017}$ & $0.7248_{\pm 0.004}$ & $0.7299_{\pm 0.004}$ & $0.7641_{\pm 0.007}$ & $0.712_{\pm 0.008}$ & $0.6778_{\pm 0.005}$ & $0.7795_{\pm 0.008}$ & \textbf{0.8171}$_{\pm 0.005}$\\
Crop & $0.7523_{\pm 0.001}$ & 0.7989 & \textbf{0.812} & $0.7141_{\pm 0.001}$ & $0.7201_{\pm 0.001}$ & $0.7208_{\pm 0.003}$ & $0.6552_{\pm 0.002}$ & $0.6409_{\pm 0.002}$ & $0.6754_{\pm 0.001}$ & $0.731_{\pm 0.001}$ & $0.7341_{\pm 0.0}$\\
DiatomSizeReduction & $0.9401_{\pm 0.008}$ & \textbf{0.9608} & 0.951 & $0.8824_{\pm 0.006}$ & $0.8617_{\pm 0.005}$ & $0.8998_{\pm 0.014}$ & $0.8649_{\pm 0.016}$ & $0.9009_{\pm 0.008}$ & $0.8617_{\pm 0.011}$ & $0.829_{\pm 0.03}$ & $0.8301_{\pm 0.02}$\\
DistalPhalanxOutlineAgeGroup & $0.717_{\pm 0.004}$ & 0.7626 & 0.7626 & $0.753_{\pm 0.004}$ & $0.741_{\pm 0.014}$ & \textbf{0.7818}$_{\pm 0.015}$ & \textbf{0.7818}$_{\pm 0.008}$ & $0.777_{\pm 0.007}$ & $0.7386_{\pm 0.004}$ & \textbf{0.7818}$_{\pm 0.015}$ & $0.7602_{\pm 0.008}$\\
DistalPhalanxOutlineCorrect & $0.7911_{\pm 0.002}$ & 0.7826 & 0.7754 & $0.7959_{\pm 0.006}$ & $0.7874_{\pm 0.008}$ & $0.8031_{\pm 0.002}$ & $0.779_{\pm 0.01}$ & $0.7548_{\pm 0.004}$ & $0.7778_{\pm 0.006}$ & \textbf{0.8188}$_{\pm 0.004}$ & $0.7886_{\pm 0.002}$\\
DistalPhalanxTW & $0.6475_{\pm 0.019}$ & 0.6978 & 0.6835 & $0.6571_{\pm 0.004}$ & $0.6667_{\pm 0.011}$ & $0.693_{\pm 0.015}$ & $0.6715_{\pm 0.004}$ & $0.6954_{\pm 0.015}$ & $0.6882_{\pm 0.018}$ & $0.6882_{\pm 0.015}$ & \textbf{0.7074}$_{\pm 0.011}$\\
DodgerLoopDay & $0.6417_{\pm 0.052}$ & 0.6125 & \textbf{0.725} & $0.4583_{\pm 0.052}$ & $0.5042_{\pm 0.019}$ & $0.4917_{\pm 0.007}$ & $0.4417_{\pm 0.014}$ & $0.5333_{\pm 0.014}$ & $0.5417_{\pm 0.029}$ & $0.5875_{\pm 0.025}$ & $0.5333_{\pm 0.019}$\\
DodgerLoopGame & \textbf{0.8333}$_{\pm 0.007}$ & 0.7899 & 0.7971 & $0.8116_{\pm 0.007}$ & $0.7633_{\pm 0.004}$ & \textbf{0.8333}$_{\pm 0.014}$ & $0.8164_{\pm 0.004}$ & $0.7826_{\pm 0.0}$ & $0.7826_{\pm 0.007}$ & $0.8164_{\pm 0.011}$ & $0.7995_{\pm 0.022}$\\
DodgerLoopWeekend & \textbf{0.9855}$_{\pm 0.0}$ & \textbf{0.9855} & 0.9783 & $0.9758_{\pm 0.004}$ & $0.9493_{\pm 0.007}$ & $0.9638_{\pm 0.013}$ & $0.9275_{\pm 0.0}$ & $0.901_{\pm 0.015}$ & $0.9589_{\pm 0.004}$ & \textbf{0.9855}$_{\pm 0.0}$ & $0.9517_{\pm 0.004}$\\
ECG200 & $0.85_{\pm 0.017}$ & \textbf{0.89} & 0.88 & $0.87_{\pm 0.01}$ & $0.86_{\pm 0.017}$ & $0.84_{\pm 0.01}$ & $0.83_{\pm 0.0}$ & $0.7833_{\pm 0.025}$ & $0.8233_{\pm 0.006}$ & $0.8167_{\pm 0.015}$ & $0.8367_{\pm 0.012}$\\
ECG5000 & $0.9398_{\pm 0.001}$ & 0.942 & \textbf{0.9447} & $0.938_{\pm 0.001}$ & $0.9331_{\pm 0.002}$ & $0.9331_{\pm 0.001}$ & $0.9403_{\pm 0.001}$ & $0.9386_{\pm 0.001}$ & $0.9332_{\pm 0.0}$ & $0.9374_{\pm 0.001}$ & $0.9381_{\pm 0.0}$\\
ECGFiveDays & $0.7975_{\pm 0.013}$ & 0.9245 & \textbf{0.9826} & $0.813_{\pm 0.002}$ & $0.8955_{\pm 0.022}$ & $0.9102_{\pm 0.02}$ & $0.9346_{\pm 0.015}$ & $0.971_{\pm 0.004}$ & $0.7871_{\pm 0.008}$ & $0.8448_{\pm 0.016}$ & $0.8707_{\pm 0.001}$\\
EOGHorizontalSignal & $0.5948_{\pm 0.004}$ & 0.5276 & 0.5 & $0.5654_{\pm 0.01}$ & $0.5313_{\pm 0.02}$ & $0.5387_{\pm 0.013}$ & $0.5387_{\pm 0.007}$ & $0.5626_{\pm 0.007}$ & $0.4521_{\pm 0.008}$ & \textbf{0.6013}$_{\pm 0.003}$ & $0.5884_{\pm 0.006}$\\
EOGVerticalSignal & \textbf{0.5092}$_{\pm 0.002}$ & 0.489 & 0.4337 & $0.453_{\pm 0.008}$ & $0.4144_{\pm 0.003}$ & $0.384_{\pm 0.003}$ & $0.442_{\pm 0.011}$ & $0.3757_{\pm 0.007}$ & $0.267_{\pm 0.002}$ & $0.4696_{\pm 0.015}$ & $0.4816_{\pm 0.011}$\\
Earthquakes & \textbf{0.7506}$_{\pm 0.008}$ & 0.7482 & 0.7482 & $0.7458_{\pm 0.011}$ & $0.7482_{\pm 0.0}$ & $0.7458_{\pm 0.004}$ & $0.741_{\pm 0.0}$ & $0.7458_{\pm 0.004}$ & $0.7434_{\pm 0.004}$ & $0.741_{\pm 0.012}$ & $0.7434_{\pm 0.004}$\\
ElectricDevices & $0.7396_{\pm 0.003}$ & 0.7025 & 0.6614 & $0.7258_{\pm 0.002}$ & $0.704_{\pm 0.003}$ & $0.7489_{\pm 0.002}$ & \textbf{0.762}$_{\pm 0.003}$ & $0.7617_{\pm 0.002}$ & $0.7066_{\pm 0.002}$ & $0.7269_{\pm 0.001}$ & $0.7424_{\pm 0.004}$\\
EthanolLevel & $0.38_{\pm 0.013}$ & \textbf{0.848} & 0.694 & $0.44_{\pm 0.007}$ & $0.306_{\pm 0.004}$ & $0.4367_{\pm 0.005}$ & $0.4047_{\pm 0.005}$ & $0.348_{\pm 0.007}$ & $0.3487_{\pm 0.005}$ & $0.332_{\pm 0.003}$ & $0.3793_{\pm 0.014}$\\
FaceAll & $0.7682_{\pm 0.028}$ & \textbf{0.8077} & 0.771 & $0.7178_{\pm 0.003}$ & $0.7651_{\pm 0.015}$ & $0.7024_{\pm 0.005}$ & $0.6911_{\pm 0.006}$ & $0.6866_{\pm 0.004}$ & $0.6487_{\pm 0.002}$ & $0.6921_{\pm 0.005}$ & $0.7205_{\pm 0.003}$\\
FaceFour & $0.8977_{\pm 0.02}$ & 0.9091 & 0.8864 & $0.7689_{\pm 0.046}$ & $0.7917_{\pm 0.046}$ & $0.6818_{\pm 0.057}$ & $0.6477_{\pm 0.05}$ & $0.8182_{\pm 0.011}$ & $0.803_{\pm 0.013}$ & $0.7841_{\pm 0.063}$ & \textbf{0.9167}$_{\pm 0.007}$\\
FacesUCR & $0.8558_{\pm 0.004}$ & 0.8766 & \textbf{0.8771} & $0.7959_{\pm 0.001}$ & $0.7995_{\pm 0.005}$ & $0.7603_{\pm 0.004}$ & $0.7815_{\pm 0.005}$ & $0.7665_{\pm 0.006}$ & $0.714_{\pm 0.009}$ & $0.8016_{\pm 0.005}$ & $0.8439_{\pm 0.006}$\\
FiftyWords & $0.726_{\pm 0.006}$ & \textbf{0.7385} & 0.7165 & $0.7099_{\pm 0.004}$ & $0.6381_{\pm 0.01}$ & $0.6689_{\pm 0.009}$ & $0.6425_{\pm 0.003}$ & $0.674_{\pm 0.006}$ & $0.6139_{\pm 0.005}$ & $0.6813_{\pm 0.011}$ & $0.7106_{\pm 0.001}$\\
Fish & $0.7619_{\pm 0.013}$ & 0.88 & 0.8857 & $0.8476_{\pm 0.014}$ & $0.8324_{\pm 0.012}$ & $0.8629_{\pm 0.006}$ & $0.9029_{\pm 0.011}$ & $0.9067_{\pm 0.012}$ & $0.9162_{\pm 0.007}$ & $0.8686_{\pm 0.01}$ & \textbf{0.9314}$_{\pm 0.0}$\\
FordA & $0.9101_{\pm 0.004}$ & 0.897 & 0.8758 & $0.8611_{\pm 0.006}$ & \textbf{0.9409}$_{\pm 0.002}$ & $0.9174_{\pm 0.002}$ & $0.8975_{\pm 0.004}$ & $0.9081_{\pm 0.001}$ & $0.8934_{\pm 0.003}$ & $0.896_{\pm 0.005}$ & $0.9303_{\pm 0.0}$\\
FordB & $0.7292_{\pm 0.007}$ & 0.7556 & 0.7136 & $0.7374_{\pm 0.002}$ & \textbf{0.814}$_{\pm 0.006}$ & $0.7918_{\pm 0.004}$ & $0.7794_{\pm 0.003}$ & $0.7626_{\pm 0.001}$ & $0.7527_{\pm 0.008}$ & $0.7584_{\pm 0.002}$ & $0.7979_{\pm 0.003}$\\
FreezerRegularTrain & \textbf{0.9996}$_{\pm 0.0}$ & 0.9986 & 0.9877 & $0.9108_{\pm 0.005}$ & $0.9261_{\pm 0.002}$ & $0.9718_{\pm 0.005}$ & $0.9827_{\pm 0.002}$ & $0.9908_{\pm 0.001}$ & $0.9766_{\pm 0.002}$ & $0.9608_{\pm 0.003}$ & $0.9857_{\pm 0.001}$\\
FreezerSmallTrain & $0.9251_{\pm 0.002}$ & 0.8933 & 0.8098 & $0.7878_{\pm 0.011}$ & $0.8208_{\pm 0.016}$ & $0.9185_{\pm 0.009}$ & $0.9353_{\pm 0.005}$ & \textbf{0.9559}$_{\pm 0.005}$ & $0.9395_{\pm 0.004}$ & $0.8713_{\pm 0.018}$ & $0.935_{\pm 0.004}$\\
Fungi & $0.9229_{\pm 0.041}$ & 0.8656 & 0.7849 & \textbf{1.0}$_{\pm 0.0}$ & $0.8405_{\pm 0.05}$ & $0.9104_{\pm 0.008}$ & $0.914_{\pm 0.014}$ & $0.8835_{\pm 0.03}$ & $0.7007_{\pm 0.031}$ & $0.7796_{\pm 0.019}$ & $0.9427_{\pm 0.028}$\\
GestureMidAirD1 & $0.6795_{\pm 0.036}$ & 0.6231 & 0.6769 & $0.6744_{\pm 0.009}$ & $0.6615_{\pm 0.013}$ & $0.6692_{\pm 0.023}$ & $0.7_{\pm 0.008}$ & \textbf{0.7359}$_{\pm 0.016}$ & $0.6564_{\pm 0.009}$ & $0.7077_{\pm 0.008}$ & $0.7128_{\pm 0.018}$\\
GestureMidAirD2 & $0.6205_{\pm 0.016}$ & 0.5615 & 0.5846 & $0.5769_{\pm 0.013}$ & $0.6513_{\pm 0.031}$ & \textbf{0.7436}$_{\pm 0.024}$ & $0.6667_{\pm 0.019}$ & $0.6795_{\pm 0.012}$ & $0.5667_{\pm 0.012}$ & $0.6077_{\pm 0.013}$ & $0.6154_{\pm 0.013}$\\
GestureMidAirD3 & $0.3923_{\pm 0.031}$ & 0.3692 & 0.3615 & $0.3846_{\pm 0.013}$ & $0.3897_{\pm 0.031}$ & $0.3538_{\pm 0.023}$ & $0.441_{\pm 0.019}$ & \textbf{0.4692}$_{\pm 0.008}$ & $0.3923_{\pm 0.008}$ & $0.4205_{\pm 0.024}$ & $0.4385_{\pm 0.013}$\\
GesturePebbleZ1 & $0.8779_{\pm 0.012}$ & 0.8488 & 0.8779 & $0.8779_{\pm 0.006}$ & $0.8702_{\pm 0.007}$ & $0.8857_{\pm 0.009}$ & $0.8663_{\pm 0.0}$ & $0.8295_{\pm 0.003}$ & $0.8915_{\pm 0.003}$ & $0.9167_{\pm 0.003}$ & \textbf{0.9205}$_{\pm 0.007}$\\
GesturePebbleZ2 & $0.7384_{\pm 0.004}$ & 0.7848 & 0.7532 & $0.865_{\pm 0.01}$ & $0.8755_{\pm 0.013}$ & $0.8186_{\pm 0.007}$ & $0.8439_{\pm 0.02}$ & $0.8439_{\pm 0.01}$ & $0.8376_{\pm 0.01}$ & \textbf{0.9388}$_{\pm 0.004}$ & $0.9114_{\pm 0.011}$\\
GunPoint & $0.9467_{\pm 0.018}$ & 0.9667 & 0.9533 & $0.9867_{\pm 0.007}$ & $0.9644_{\pm 0.01}$ & \textbf{0.9956}$_{\pm 0.004}$ & $0.9933_{\pm 0.0}$ & $0.9667_{\pm 0.018}$ & $0.9422_{\pm 0.004}$ & $0.98_{\pm 0.0}$ & $0.9778_{\pm 0.004}$\\
GunPointAgeSpan & $0.9884_{\pm 0.002}$ & \textbf{0.9905} & 0.9842 & $0.9525_{\pm 0.008}$ & $0.9821_{\pm 0.005}$ & $0.9852_{\pm 0.002}$ & $0.9873_{\pm 0.0}$ & $0.9673_{\pm 0.01}$ & $0.9715_{\pm 0.003}$ & $0.9852_{\pm 0.002}$ & $0.9842_{\pm 0.003}$\\
GunPointMaleVersusFemale & $0.9937_{\pm 0.0}$ & 0.9968 & \textbf{1.0} & $0.9916_{\pm 0.004}$ & $0.9947_{\pm 0.004}$ & $0.9884_{\pm 0.002}$ & $0.9968_{\pm 0.003}$ & $0.9979_{\pm 0.004}$ & $0.9789_{\pm 0.005}$ & $0.9905_{\pm 0.0}$ & \textbf{1.0}$_{\pm 0.0}$\\
GunPointOldVersusYoung & \textbf{1.0}$_{\pm 0.0}$ & \textbf{1.0} & \textbf{1.0} & $0.9598_{\pm 0.005}$ & $0.9672_{\pm 0.007}$ & $0.9778_{\pm 0.003}$ & $0.9852_{\pm 0.005}$ & $0.9937_{\pm 0.0}$ & \textbf{1.0}$_{\pm 0.0}$ & $0.9968_{\pm 0.0}$ & $0.9968_{\pm 0.0}$\\
Ham & $0.6063_{\pm 0.005}$ & \textbf{0.7429} & 0.7238 & $0.7333_{\pm 0.019}$ & $0.7143_{\pm 0.01}$ & $0.6413_{\pm 0.015}$ & $0.6508_{\pm 0.04}$ & $0.6349_{\pm 0.02}$ & $0.7111_{\pm 0.048}$ & $0.7016_{\pm 0.02}$ & $0.6698_{\pm 0.031}$\\
HandOutlines & $0.9036_{\pm 0.004}$ & 0.9162 & \textbf{0.927} & $0.909_{\pm 0.006}$ & $0.8757_{\pm 0.016}$ & $0.9108_{\pm 0.003}$ & $0.8874_{\pm 0.009}$ & $0.8874_{\pm 0.01}$ & $0.8964_{\pm 0.007}$ & $0.8973_{\pm 0.003}$ & $0.9036_{\pm 0.006}$\\
Haptics & $0.4838_{\pm 0.015}$ & 0.4708 & 0.461 & $0.5238_{\pm 0.008}$ & $0.5271_{\pm 0.014}$ & \textbf{0.5444}$_{\pm 0.009}$ & $0.5076_{\pm 0.012}$ & $0.5141_{\pm 0.008}$ & $0.474_{\pm 0.006}$ & $0.54_{\pm 0.005}$ & $0.5011_{\pm 0.011}$\\
Herring & $0.5208_{\pm 0.024}$ & 0.5938 & 0.6406 & $0.5885_{\pm 0.024}$ & $0.6354_{\pm 0.065}$ & \textbf{0.6667}$_{\pm 0.018}$ & $0.5938_{\pm 0.031}$ & $0.5573_{\pm 0.033}$ & $0.6458_{\pm 0.018}$ & $0.6042_{\pm 0.009}$ & $0.6302_{\pm 0.018}$\\
HouseTwenty & $0.9692_{\pm 0.005}$ & 0.8487 & 0.7563 & $0.9384_{\pm 0.005}$ & $0.9692_{\pm 0.005}$ & $0.958_{\pm 0.008}$ & \textbf{0.9776}$_{\pm 0.005}$ & $0.9496_{\pm 0.008}$ & $0.8852_{\pm 0.013}$ & $0.9496_{\pm 0.0}$ & $0.9524_{\pm 0.013}$\\
InlineSkate & $0.3891_{\pm 0.005}$ & 0.3345 & 0.3436 & $0.3224_{\pm 0.006}$ & $0.4491_{\pm 0.014}$ & $0.4194_{\pm 0.007}$ & $0.3848_{\pm 0.012}$ & \textbf{0.4576}$_{\pm 0.006}$ & $0.32_{\pm 0.007}$ & $0.3927_{\pm 0.008}$ & $0.3855_{\pm 0.005}$\\
InsectEPGRegularTrain & \textbf{1.0}$_{\pm 0.0}$ & \textbf{1.0} & \textbf{1.0} & $0.9304_{\pm 0.002}$ & $0.9692_{\pm 0.01}$ & $0.9518_{\pm 0.007}$ & \textbf{1.0}$_{\pm 0.0}$ & $0.9893_{\pm 0.006}$ & \textbf{1.0}$_{\pm 0.0}$ & \textbf{1.0}$_{\pm 0.0}$ & \textbf{1.0}$_{\pm 0.0}$\\
InsectEPGSmallTrain & \textbf{1.0}$_{\pm 0.0}$ & \textbf{1.0} & \textbf{1.0} & $0.8246_{\pm 0.006}$ & $0.8889_{\pm 0.002}$ & $0.8715_{\pm 0.018}$ & $0.9451_{\pm 0.026}$ & $0.9505_{\pm 0.012}$ & \textbf{1.0}$_{\pm 0.0}$ & \textbf{1.0}$_{\pm 0.0}$ & \textbf{1.0}$_{\pm 0.0}$\\
InsectWingbeatSound & $0.629_{\pm 0.003}$ & \textbf{0.6672} & 0.6556 & $0.6258_{\pm 0.003}$ & $0.651_{\pm 0.001}$ & $0.6271_{\pm 0.003}$ & $0.5258_{\pm 0.013}$ & $0.5365_{\pm 0.004}$ & $0.5167_{\pm 0.006}$ & $0.5359_{\pm 0.002}$ & $0.6056_{\pm 0.009}$\\
\bottomrule
\end{tabular}
    }}
\end{table}

\newpage

\begin{table}[H]
    \centering
    \caption{Zero-shot feature extraction results on UCR (Random Forest). Second Part.}
    \label{tab:sota-ucr-res-rf-b}
    {\scalebox{0.55}{
    \begin{tabular}{l|lllllllllll}
\toprule
                               & Catch22+ & TabPFN & TabICL & MOMENT & TiRex & Chronos2 & TiViT-H & TiConvNext & NuTime & Mantis+ & MantisV2\\
\midrule
ItalyPowerDemand & $0.9537_{\pm 0.002}$ & \textbf{0.9699} & 0.9631 & $0.9482_{\pm 0.004}$ & $0.9624_{\pm 0.004}$ & $0.9559_{\pm 0.003}$ & $0.8776_{\pm 0.003}$ & $0.9248_{\pm 0.002}$ & $0.8737_{\pm 0.004}$ & $0.9002_{\pm 0.005}$ & $0.9248_{\pm 0.004}$\\
LargeKitchenAppliances & $0.8169_{\pm 0.009}$ & 0.6373 & 0.696 & $0.7342_{\pm 0.007}$ & $0.8053_{\pm 0.005}$ & \textbf{0.8729}$_{\pm 0.003}$ & $0.816_{\pm 0.003}$ & $0.8702_{\pm 0.002}$ & $0.7556_{\pm 0.009}$ & $0.8133_{\pm 0.011}$ & $0.7662_{\pm 0.006}$\\
Lightning2 & $0.7322_{\pm 0.009}$ & 0.6721 & 0.7049 & $0.7869_{\pm 0.033}$ & $0.7432_{\pm 0.009}$ & $0.7268_{\pm 0.009}$ & $0.7596_{\pm 0.009}$ & $0.7432_{\pm 0.025}$ & $0.6776_{\pm 0.009}$ & \textbf{0.7923}$_{\pm 0.009}$ & $0.7268_{\pm 0.009}$\\
Lightning7 & \textbf{0.7671}$_{\pm 0.014}$ & 0.6986 & 0.7397 & $0.7352_{\pm 0.029}$ & $0.7352_{\pm 0.016}$ & $0.653_{\pm 0.021}$ & $0.7489_{\pm 0.029}$ & $0.7306_{\pm 0.016}$ & $0.6575_{\pm 0.027}$ & $0.7397_{\pm 0.027}$ & $0.7215_{\pm 0.044}$\\
Mallat & $0.9555_{\pm 0.009}$ & \textbf{0.9689} & 0.9484 & $0.9164_{\pm 0.016}$ & $0.9166_{\pm 0.027}$ & $0.9016_{\pm 0.009}$ & $0.9032_{\pm 0.007}$ & $0.8387_{\pm 0.005}$ & $0.8311_{\pm 0.005}$ & $0.8708_{\pm 0.01}$ & $0.8887_{\pm 0.002}$\\
Meat & $0.9222_{\pm 0.01}$ & \textbf{0.9833} & 0.9333 & $0.9222_{\pm 0.01}$ & $0.8611_{\pm 0.01}$ & $0.9333_{\pm 0.017}$ & $0.8389_{\pm 0.038}$ & $0.8667_{\pm 0.0}$ & $0.9278_{\pm 0.019}$ & $0.8944_{\pm 0.01}$ & $0.9111_{\pm 0.025}$\\
MedicalImages & $0.7737_{\pm 0.005}$ & 0.7947 & \textbf{0.8079} & $0.714_{\pm 0.008}$ & $0.7092_{\pm 0.007}$ & $0.7246_{\pm 0.009}$ & $0.7504_{\pm 0.005}$ & $0.7368_{\pm 0.0}$ & $0.7132_{\pm 0.002}$ & $0.7434_{\pm 0.005}$ & $0.7522_{\pm 0.006}$\\
MelbournePedestrian & $0.9597_{\pm 0.0}$ & \textbf{0.9803} & \textbf{0.9803} & $0.8643_{\pm 0.0}$ & $0.8812_{\pm 0.002}$ & $0.8927_{\pm 0.006}$ & $0.8074_{\pm 0.004}$ & $0.8195_{\pm 0.003}$ & $0.9172_{\pm 0.002}$ & $0.9464_{\pm 0.0}$ & $0.9513_{\pm 0.001}$\\
MiddlePhalanxOutlineAgeGroup & $0.5952_{\pm 0.004}$ & 0.6234 & \textbf{0.6299} & $0.5649_{\pm 0.017}$ & $0.5996_{\pm 0.016}$ & $0.5584_{\pm 0.017}$ & $0.6039_{\pm 0.006}$ & \textbf{0.6299}$_{\pm 0.017}$ & $0.6061_{\pm 0.004}$ & $0.6082_{\pm 0.025}$ & $0.5801_{\pm 0.004}$\\
MiddlePhalanxOutlineCorrect & $0.811_{\pm 0.012}$ & 0.8522 & 0.8351 & $0.858_{\pm 0.008}$ & $0.8179_{\pm 0.007}$ & \textbf{0.8717}$_{\pm 0.008}$ & $0.8247_{\pm 0.009}$ & $0.8156_{\pm 0.012}$ & $0.7892_{\pm 0.009}$ & $0.7927_{\pm 0.011}$ & $0.8339_{\pm 0.004}$\\
MiddlePhalanxTW & $0.5844_{\pm 0.006}$ & 0.6169 & \textbf{0.6234} & $0.5779_{\pm 0.017}$ & $0.5736_{\pm 0.01}$ & $0.5714_{\pm 0.013}$ & $0.5519_{\pm 0.023}$ & $0.5693_{\pm 0.019}$ & $0.5281_{\pm 0.014}$ & $0.5346_{\pm 0.02}$ & $0.5325_{\pm 0.006}$\\
MixedShapesRegularTrain & $0.9306_{\pm 0.003}$ & 0.9344 & 0.9299 & $0.9166_{\pm 0.001}$ & $0.9472_{\pm 0.001}$ & $0.9427_{\pm 0.001}$ & $0.9513_{\pm 0.001}$ & \textbf{0.9564}$_{\pm 0.003}$ & $0.9381_{\pm 0.002}$ & $0.9461_{\pm 0.003}$ & $0.9467_{\pm 0.0}$\\
MixedShapesSmallTrain & $0.8827_{\pm 0.002}$ & 0.8293 & 0.8767 & $0.8506_{\pm 0.008}$ & $0.909_{\pm 0.002}$ & $0.9019_{\pm 0.003}$ & \textbf{0.919}$_{\pm 0.005}$ & $0.9182_{\pm 0.002}$ & $0.908_{\pm 0.003}$ & $0.9146_{\pm 0.002}$ & $0.9157_{\pm 0.001}$\\
MoteStrain & $0.8818_{\pm 0.015}$ & 0.889 & 0.8794 & $0.8895_{\pm 0.008}$ & $0.9193_{\pm 0.003}$ & $0.9332_{\pm 0.004}$ & $0.8586_{\pm 0.005}$ & $0.9055_{\pm 0.002}$ & \textbf{0.9481}$_{\pm 0.002}$ & $0.9121_{\pm 0.003}$ & $0.931_{\pm 0.004}$\\
NonInvasiveFetalECGThorax1 & $0.8877_{\pm 0.004}$ & \textbf{0.941} & 0.9272 & $0.8863_{\pm 0.002}$ & $0.8656_{\pm 0.0}$ & $0.8295_{\pm 0.001}$ & $0.8137_{\pm 0.005}$ & $0.8244_{\pm 0.006}$ & $0.78_{\pm 0.005}$ & $0.8575_{\pm 0.004}$ & $0.864_{\pm 0.002}$\\
NonInvasiveFetalECGThorax2 & $0.9091_{\pm 0.0}$ & \textbf{0.9476} & 0.9405 & $0.9104_{\pm 0.001}$ & $0.888_{\pm 0.004}$ & $0.8675_{\pm 0.001}$ & $0.8755_{\pm 0.002}$ & $0.8692_{\pm 0.004}$ & $0.8175_{\pm 0.006}$ & $0.8906_{\pm 0.002}$ & $0.8867_{\pm 0.0}$\\
OSULeaf & $0.6832_{\pm 0.009}$ & 0.5661 & 0.595 & $0.7521_{\pm 0.007}$ & $0.9174_{\pm 0.007}$ & $0.8953_{\pm 0.01}$ & $0.9463_{\pm 0.004}$ & \textbf{0.9793}$_{\pm 0.004}$ & $0.8003_{\pm 0.005}$ & $0.9298_{\pm 0.004}$ & $0.9353_{\pm 0.002}$\\
OliveOil & $0.8444_{\pm 0.019}$ & \textbf{0.9333} & 0.9 & $0.8889_{\pm 0.019}$ & $0.8778_{\pm 0.019}$ & $0.8556_{\pm 0.019}$ & $0.5778_{\pm 0.038}$ & $0.8556_{\pm 0.019}$ & $0.7_{\pm 0.0}$ & $0.8333_{\pm 0.0}$ & $0.8667_{\pm 0.033}$\\
PLAID & $0.8752_{\pm 0.009}$ & 0.7896 & 0.5661 & $0.7393_{\pm 0.005}$ & $0.8684_{\pm 0.003}$ & $0.8591_{\pm 0.008}$ & $0.8709_{\pm 0.007}$ & \textbf{0.892}$_{\pm 0.005}$ & $0.7765_{\pm 0.004}$ & $0.8324_{\pm 0.004}$ & $0.8187_{\pm 0.004}$\\
PhalangesOutlinesCorrect & $0.8252_{\pm 0.003}$ & 0.8403 & \textbf{0.8613} & $0.8225_{\pm 0.002}$ & $0.8197_{\pm 0.003}$ & $0.8349_{\pm 0.004}$ & $0.796_{\pm 0.002}$ & $0.7949_{\pm 0.003}$ & $0.7766_{\pm 0.003}$ & $0.8116_{\pm 0.003}$ & $0.824_{\pm 0.002}$\\
Phoneme & $0.3216_{\pm 0.005}$ & 0.1097 & 0.1361 & $0.2938_{\pm 0.008}$ & $0.3771_{\pm 0.004}$ & \textbf{0.3936}$_{\pm 0.002}$ & $0.355_{\pm 0.001}$ & $0.3745_{\pm 0.004}$ & $0.2913_{\pm 0.008}$ & $0.3486_{\pm 0.001}$ & $0.3641_{\pm 0.005}$\\
PickupGestureWiimoteZ & $0.6933_{\pm 0.012}$ & 0.76 & 0.74 & $0.5867_{\pm 0.031}$ & $0.7333_{\pm 0.023}$ & $0.68_{\pm 0.0}$ & \textbf{0.82}$_{\pm 0.02}$ & \textbf{0.82}$_{\pm 0.035}$ & $0.6667_{\pm 0.031}$ & $0.7667_{\pm 0.031}$ & $0.7867_{\pm 0.031}$\\
PigAirwayPressure & $0.2372_{\pm 0.003}$ & 0.0192 & 0.1538 & $0.1074_{\pm 0.014}$ & $0.3462_{\pm 0.029}$ & $0.3333_{\pm 0.012}$ & $0.4744_{\pm 0.029}$ & \textbf{0.5769}$_{\pm 0.005}$ & $0.3478_{\pm 0.012}$ & $0.4663_{\pm 0.013}$ & $0.4904_{\pm 0.017}$\\
PigArtPressure & $0.891_{\pm 0.007}$ & 0.0337 & 0.2548 & $0.5369_{\pm 0.015}$ & $0.8734_{\pm 0.02}$ & $0.8061_{\pm 0.018}$ & $0.8173_{\pm 0.027}$ & $0.9087_{\pm 0.01}$ & $0.9359_{\pm 0.012}$ & \textbf{0.9391}$_{\pm 0.007}$ & $0.9343_{\pm 0.003}$\\
PigCVP & $0.5128_{\pm 0.011}$ & 0.0192 & 0.1731 & $0.4407_{\pm 0.007}$ & $0.8349_{\pm 0.007}$ & $0.6731_{\pm 0.024}$ & $0.6795_{\pm 0.01}$ & $0.7131_{\pm 0.025}$ & $0.8285_{\pm 0.01}$ & $0.8686_{\pm 0.007}$ & \textbf{0.8974}$_{\pm 0.02}$\\
Plane & \textbf{1.0}$_{\pm 0.0}$ & 0.9905 & 0.9905 & $0.9968_{\pm 0.005}$ & \textbf{1.0}$_{\pm 0.0}$ & \textbf{1.0}$_{\pm 0.0}$ & \textbf{1.0}$_{\pm 0.0}$ & \textbf{1.0}$_{\pm 0.0}$ & \textbf{1.0}$_{\pm 0.0}$ & \textbf{1.0}$_{\pm 0.0}$ & \textbf{1.0}$_{\pm 0.0}$\\
PowerCons & $0.9944_{\pm 0.0}$ & \textbf{1.0} & \textbf{1.0} & $0.9407_{\pm 0.003}$ & $0.8981_{\pm 0.008}$ & $0.9407_{\pm 0.006}$ & $0.8907_{\pm 0.008}$ & $0.9074_{\pm 0.017}$ & $0.9333_{\pm 0.006}$ & $0.95_{\pm 0.01}$ & $0.9648_{\pm 0.003}$\\
ProximalPhalanxOutlineAgeGroup & $0.8472_{\pm 0.007}$ & \textbf{0.8585} & 0.839 & $0.8341_{\pm 0.005}$ & $0.852_{\pm 0.003}$ & $0.8569_{\pm 0.007}$ & $0.8537_{\pm 0.013}$ & $0.8537_{\pm 0.01}$ & $0.8504_{\pm 0.003}$ & $0.8309_{\pm 0.003}$ & $0.8293_{\pm 0.005}$\\
ProximalPhalanxOutlineCorrect & $0.8648_{\pm 0.012}$ & 0.9038 & \textbf{0.9244} & $0.8603_{\pm 0.01}$ & $0.8774_{\pm 0.009}$ & $0.8706_{\pm 0.004}$ & $0.8511_{\pm 0.005}$ & $0.8328_{\pm 0.008}$ & $0.8373_{\pm 0.004}$ & $0.8442_{\pm 0.005}$ & $0.8671_{\pm 0.005}$\\
ProximalPhalanxTW & $0.8033_{\pm 0.007}$ & 0.8098 & \textbf{0.8293} & $0.8114_{\pm 0.01}$ & $0.8065_{\pm 0.023}$ & $0.7984_{\pm 0.003}$ & $0.7772_{\pm 0.011}$ & $0.7919_{\pm 0.01}$ & $0.8146_{\pm 0.005}$ & $0.7854_{\pm 0.015}$ & $0.8211_{\pm 0.016}$\\
RefrigerationDevices & $0.5493_{\pm 0.023}$ & 0.504 & 0.4933 & $0.5493_{\pm 0.01}$ & $0.568_{\pm 0.0}$ & $0.5502_{\pm 0.009}$ & $0.5467_{\pm 0.003}$ & \textbf{0.5911}$_{\pm 0.016}$ & $0.5369_{\pm 0.009}$ & $0.5316_{\pm 0.009}$ & $0.5636_{\pm 0.006}$\\
Rock & $0.62_{\pm 0.02}$ & 0.76 & 0.64 & $0.8333_{\pm 0.012}$ & \textbf{0.9}$_{\pm 0.04}$ & $0.8733_{\pm 0.031}$ & $0.8933_{\pm 0.031}$ & \textbf{0.9}$_{\pm 0.02}$ & $0.6533_{\pm 0.061}$ & $0.6733_{\pm 0.046}$ & $0.82_{\pm 0.02}$\\
ScreenType & $0.5271_{\pm 0.007}$ & 0.4187 & 0.4107 & $0.4284_{\pm 0.017}$ & $0.5156_{\pm 0.015}$ & $0.4871_{\pm 0.008}$ & \textbf{0.5449}$_{\pm 0.01}$ & $0.5396_{\pm 0.026}$ & $0.5058_{\pm 0.012}$ & $0.4498_{\pm 0.008}$ & $0.4596_{\pm 0.014}$\\
SemgHandGenderCh2 & $0.9272_{\pm 0.003}$ & \textbf{0.9467} & 0.8867 & $0.7778_{\pm 0.003}$ & $0.8817_{\pm 0.0}$ & $0.8906_{\pm 0.008}$ & $0.8394_{\pm 0.003}$ & $0.8717_{\pm 0.002}$ & $0.8561_{\pm 0.003}$ & $0.9283_{\pm 0.005}$ & $0.9139_{\pm 0.006}$\\
SemgHandMovementCh2 & \textbf{0.8615}$_{\pm 0.007}$ & 0.7711 & 0.5689 & $0.4252_{\pm 0.005}$ & $0.6489_{\pm 0.016}$ & $0.6259_{\pm 0.016}$ & $0.537_{\pm 0.008}$ & $0.597_{\pm 0.013}$ & $0.6756_{\pm 0.006}$ & $0.7711_{\pm 0.01}$ & $0.7311_{\pm 0.008}$\\
SemgHandSubjectCh2 & $0.8837_{\pm 0.007}$ & \textbf{0.9356} & 0.8333 & $0.6504_{\pm 0.006}$ & $0.8244_{\pm 0.008}$ & $0.8259_{\pm 0.007}$ & $0.7822_{\pm 0.008}$ & $0.8237_{\pm 0.008}$ & $0.7622_{\pm 0.004}$ & $0.8385_{\pm 0.012}$ & $0.8341_{\pm 0.011}$\\
ShakeGestureWiimoteZ & $0.8467_{\pm 0.031}$ & 0.82 & 0.74 & $0.84_{\pm 0.0}$ & $0.88_{\pm 0.035}$ & $0.86_{\pm 0.02}$ & $0.8467_{\pm 0.042}$ & $0.8267_{\pm 0.012}$ & $0.9133_{\pm 0.012}$ & \textbf{0.9333}$_{\pm 0.012}$ & \textbf{0.9333}$_{\pm 0.012}$\\
ShapeletSim & $0.9704_{\pm 0.008}$ & 0.4778 & 0.5056 & $0.9593_{\pm 0.008}$ & $0.9426_{\pm 0.008}$ & \textbf{1.0}$_{\pm 0.0}$ & \textbf{1.0}$_{\pm 0.0}$ & \textbf{1.0}$_{\pm 0.0}$ & $0.9204_{\pm 0.012}$ & $0.9556_{\pm 0.01}$ & $0.9593_{\pm 0.003}$\\
ShapesAll & $0.8206_{\pm 0.002}$ & 0.8017 & 0.7917 & $0.8322_{\pm 0.006}$ & $0.8356_{\pm 0.012}$ & \textbf{0.8839}$_{\pm 0.003}$ & $0.8678_{\pm 0.003}$ & $0.8444_{\pm 0.005}$ & $0.8394_{\pm 0.004}$ & $0.8628_{\pm 0.003}$ & $0.87_{\pm 0.009}$\\
SmallKitchenAppliances & $0.8302_{\pm 0.008}$ & 0.7867 & 0.7627 & $0.7111_{\pm 0.007}$ & $0.8382_{\pm 0.002}$ & $0.8373_{\pm 0.009}$ & $0.8284_{\pm 0.006}$ & $0.8302_{\pm 0.011}$ & $0.8196_{\pm 0.004}$ & \textbf{0.8436}$_{\pm 0.004}$ & $0.8373_{\pm 0.0}$\\
SmoothSubspace & $0.9867_{\pm 0.007}$ & \textbf{1.0} & \textbf{1.0} & $0.9578_{\pm 0.004}$ & $0.9222_{\pm 0.01}$ & $0.9289_{\pm 0.023}$ & $0.9333_{\pm 0.007}$ & $0.9267_{\pm 0.007}$ & $0.8733_{\pm 0.007}$ & $0.9578_{\pm 0.008}$ & $0.9511_{\pm 0.004}$\\
SonyAIBORobotSurface1 & $0.8397_{\pm 0.002}$ & 0.772 & 0.6722 & $0.8541_{\pm 0.008}$ & \textbf{0.8625}$_{\pm 0.03}$ & $0.6628_{\pm 0.02}$ & $0.7876_{\pm 0.01}$ & $0.7643_{\pm 0.012}$ & $0.8092_{\pm 0.012}$ & $0.8453_{\pm 0.012}$ & $0.8231_{\pm 0.011}$\\
SonyAIBORobotSurface2 & $0.8835_{\pm 0.021}$ & 0.809 & 0.8279 & $0.8279_{\pm 0.002}$ & $0.8255_{\pm 0.003}$ & $0.8475_{\pm 0.002}$ & $0.9038_{\pm 0.002}$ & $0.9185_{\pm 0.004}$ & $0.8391_{\pm 0.01}$ & $0.9164_{\pm 0.005}$ & \textbf{0.922}$_{\pm 0.009}$\\
StarLightCurves & $0.9702_{\pm 0.001}$ & 0.9732 & 0.9718 & $0.9768_{\pm 0.0}$ & $0.9789_{\pm 0.0}$ & $0.98_{\pm 0.0}$ & $0.9788_{\pm 0.0}$ & $0.9803_{\pm 0.0}$ & $0.979_{\pm 0.0}$ & $0.98_{\pm 0.0}$ & \textbf{0.9806}$_{\pm 0.0}$\\
Strawberry & $0.9333_{\pm 0.003}$ & 0.9811 & \textbf{0.9838} & $0.9568_{\pm 0.005}$ & $0.9532_{\pm 0.004}$ & $0.9432_{\pm 0.003}$ & $0.927_{\pm 0.007}$ & $0.9414_{\pm 0.002}$ & $0.936_{\pm 0.006}$ & $0.9649_{\pm 0.0}$ & $0.9595_{\pm 0.007}$\\
SwedishLeaf & $0.9115_{\pm 0.002}$ & 0.9504 & 0.9456 & $0.9211_{\pm 0.002}$ & $0.9381_{\pm 0.005}$ & $0.9381_{\pm 0.004}$ & $0.9408_{\pm 0.002}$ & $0.9376_{\pm 0.006}$ & $0.9221_{\pm 0.002}$ & $0.9456_{\pm 0.003}$ & \textbf{0.9547}$_{\pm 0.004}$\\
Symbols & $0.9618_{\pm 0.005}$ & 0.8824 & 0.8945 & $0.9377_{\pm 0.002}$ & $0.937_{\pm 0.004}$ & $0.9745_{\pm 0.006}$ & $0.9779_{\pm 0.005}$ & $0.9759_{\pm 0.003}$ & $0.9387_{\pm 0.005}$ & \textbf{0.9806}$_{\pm 0.002}$ & $0.9698_{\pm 0.005}$\\
SyntheticControl & $0.9922_{\pm 0.002}$ & 0.99 & 0.9833 & $0.9578_{\pm 0.002}$ & $0.9867_{\pm 0.0}$ & $0.9933_{\pm 0.003}$ & $0.9956_{\pm 0.002}$ & \textbf{0.9978}$_{\pm 0.002}$ & $0.9722_{\pm 0.002}$ & $0.9822_{\pm 0.002}$ & $0.99_{\pm 0.0}$\\
ToeSegmentation1 & $0.8553_{\pm 0.016}$ & 0.5746 & 0.6667 & $0.9313_{\pm 0.005}$ & $0.9474_{\pm 0.008}$ & $0.924_{\pm 0.024}$ & $0.9415_{\pm 0.003}$ & $0.8611_{\pm 0.028}$ & $0.8436_{\pm 0.022}$ & $0.9313_{\pm 0.009}$ & \textbf{0.9547}$_{\pm 0.007}$\\
ToeSegmentation2 & $0.7846_{\pm 0.013}$ & 0.6538 & 0.8154 & $0.8462_{\pm 0.008}$ & \textbf{0.9026}$_{\pm 0.004}$ & $0.8769_{\pm 0.008}$ & $0.8564_{\pm 0.012}$ & $0.8513_{\pm 0.016}$ & $0.7385_{\pm 0.008}$ & $0.8513_{\pm 0.009}$ & $0.8821_{\pm 0.004}$\\
Trace & \textbf{1.0}$_{\pm 0.0}$ & 0.91 & 0.98 & $0.99_{\pm 0.0}$ & \textbf{1.0}$_{\pm 0.0}$ & \textbf{1.0}$_{\pm 0.0}$ & \textbf{1.0}$_{\pm 0.0}$ & \textbf{1.0}$_{\pm 0.0}$ & $0.99_{\pm 0.0}$ & \textbf{1.0}$_{\pm 0.0}$ & \textbf{1.0}$_{\pm 0.0}$\\
TwoLeadECG & $0.8584_{\pm 0.015}$ & 0.9508 & 0.9254 & $0.9485_{\pm 0.007}$ & $0.9511_{\pm 0.007}$ & $0.9309_{\pm 0.011}$ & $0.9936_{\pm 0.001}$ & $0.983_{\pm 0.002}$ & $0.9166_{\pm 0.025}$ & $0.9886_{\pm 0.005}$ & \textbf{0.9956}$_{\pm 0.001}$\\
TwoPatterns & $0.9935_{\pm 0.001}$ & \textbf{0.995} & 0.9032 & $0.9158_{\pm 0.004}$ & $0.9813_{\pm 0.002}$ & $0.908_{\pm 0.006}$ & $0.9474_{\pm 0.001}$ & $0.9693_{\pm 0.003}$ & $0.8502_{\pm 0.002}$ & $0.9622_{\pm 0.003}$ & $0.9819_{\pm 0.001}$\\
UMD & $0.9398_{\pm 0.033}$ & \textbf{1.0} & 0.9375 & $0.9838_{\pm 0.004}$ & $0.919_{\pm 0.004}$ & $0.9931_{\pm 0.0}$ & $0.9861_{\pm 0.0}$ & $0.9907_{\pm 0.004}$ & $0.9352_{\pm 0.014}$ & $0.9861_{\pm 0.0}$ & $0.9815_{\pm 0.004}$\\
UWaveGestureLibraryAll & $0.9454_{\pm 0.001}$ & \textbf{0.9665} & 0.9629 & $0.9162_{\pm 0.002}$ & $0.9073_{\pm 0.002}$ & $0.9229_{\pm 0.001}$ & $0.8637_{\pm 0.0}$ & $0.8702_{\pm 0.003}$ & $0.8882_{\pm 0.002}$ & $0.8733_{\pm 0.001}$ & $0.8848_{\pm 0.001}$\\
UWaveGestureLibraryX & $0.8062_{\pm 0.003}$ & 0.8079 & 0.7984 & $0.7938_{\pm 0.002}$ & $0.7824_{\pm 0.003}$ & $0.8036_{\pm 0.001}$ & $0.7948_{\pm 0.003}$ & $0.7992_{\pm 0.001}$ & $0.8125_{\pm 0.002}$ & $0.8068_{\pm 0.001}$ & \textbf{0.813}$_{\pm 0.001}$\\
UWaveGestureLibraryY & $0.7248_{\pm 0.002}$ & 0.715 & 0.7164 & $0.7139_{\pm 0.002}$ & $0.7236_{\pm 0.002}$ & $0.7344_{\pm 0.003}$ & $0.739_{\pm 0.002}$ & \textbf{0.7633}$_{\pm 0.001}$ & $0.7391_{\pm 0.003}$ & $0.7475_{\pm 0.004}$ & $0.7572_{\pm 0.001}$\\
UWaveGestureLibraryZ & $0.7519_{\pm 0.002}$ & 0.7513 & 0.7379 & $0.7339_{\pm 0.004}$ & $0.7361_{\pm 0.001}$ & $0.7464_{\pm 0.004}$ & $0.743_{\pm 0.006}$ & $0.7562_{\pm 0.005}$ & $0.7518_{\pm 0.005}$ & $0.7495_{\pm 0.002}$ & \textbf{0.7587}$_{\pm 0.0}$\\
Wafer & $0.9988_{\pm 0.0}$ & 0.9953 & 0.9959 & $0.9878_{\pm 0.001}$ & $0.9986_{\pm 0.0}$ & $0.985_{\pm 0.001}$ & $0.9953_{\pm 0.0}$ & \textbf{0.9993}$_{\pm 0.0}$ & $0.9935_{\pm 0.001}$ & $0.995_{\pm 0.001}$ & $0.9944_{\pm 0.001}$\\
Wine & $0.6358_{\pm 0.039}$ & 0.7778 & 0.7222 & $0.7531_{\pm 0.021}$ & $0.6296_{\pm 0.049}$ & \textbf{0.8704}$_{\pm 0.019}$ & $0.5123_{\pm 0.021}$ & $0.6605_{\pm 0.028}$ & $0.7099_{\pm 0.039}$ & $0.716_{\pm 0.011}$ & $0.8086_{\pm 0.06}$\\
WordSynonyms & $0.639_{\pm 0.008}$ & \textbf{0.6442} & 0.5972 & $0.6186_{\pm 0.01}$ & $0.5204_{\pm 0.005}$ & $0.5711_{\pm 0.001}$ & $0.5674_{\pm 0.007}$ & $0.5569_{\pm 0.008}$ & $0.5303_{\pm 0.014}$ & $0.5914_{\pm 0.007}$ & $0.6181_{\pm 0.006}$\\
Worms & $0.7229_{\pm 0.015}$ & 0.5714 & 0.5584 & $0.7359_{\pm 0.015}$ & $0.8009_{\pm 0.007}$ & $0.7662_{\pm 0.0}$ & \textbf{0.8312}$_{\pm 0.013}$ & $0.8009_{\pm 0.02}$ & $0.7229_{\pm 0.02}$ & $0.7749_{\pm 0.007}$ & $0.7576_{\pm 0.02}$\\
WormsTwoClass & $0.8182_{\pm 0.013}$ & 0.6104 & 0.5714 & $0.8009_{\pm 0.007}$ & \textbf{0.8485}$_{\pm 0.015}$ & $0.8139_{\pm 0.027}$ & $0.8355_{\pm 0.02}$ & $0.8225_{\pm 0.015}$ & $0.7835_{\pm 0.007}$ & $0.7922_{\pm 0.022}$ & $0.8182_{\pm 0.013}$\\
Yoga & $0.8289_{\pm 0.005}$ & 0.8603 & \textbf{0.8653} & $0.8369_{\pm 0.006}$ & $0.7677_{\pm 0.002}$ & $0.8162_{\pm 0.003}$ & $0.8091_{\pm 0.007}$ & $0.8144_{\pm 0.002}$ & $0.8207_{\pm 0.002}$ & $0.789_{\pm 0.006}$ & $0.848_{\pm 0.003}$\\
\midrule
\textit{\textbf{Average}} & 0.7969 & 0.7806 & 0.7707 &  0.7789 & 0.8013 & 0.8002 & 0.7943 & 0.8029 & 0.7732 & 0.8061 & \textbf{0.8195}\\
\textit{\textbf{Best Counts}} & 13 & 30 & 24 &  1 & 11 & 14 & 13 & 21 & 6 & 14 & 27 \\
\bottomrule
\end{tabular}
    }}
\end{table}

\vspace{0.4cm}

\begin{table}[H]
    \centering
    \caption{Zero-shot feature extraction results on UEA-27 (Random Forest).}
    \label{tab:uea-sota-rf}
    {\scalebox{0.55}{
    \begin{tabular}{l|lllllllllll}
        \toprule
                                       & Catch22+ & TabPFN & TabICL & MOMENT & TiRex & Chronos2 & TiViT-H & TiConvNext & NuTime & Mantis+ & MantisV2\\
        \midrule
        ArticularyWordRecognition & $0.96_{\pm 0.007}$ & 0.93 & 0.91 & $0.9844_{\pm 0.002}$ & $0.99_{\pm 0.0}$ & $0.9933_{\pm 0.003}$ & $0.9833_{\pm 0.003}$ & $0.98_{\pm 0.003}$ & $0.9911_{\pm 0.004}$ & \textbf{0.9956}$_{\pm 0.002}$ & $0.9922_{\pm 0.002}$\\
        BasicMotions & \textbf{1.0}$_{\pm 0.0}$ & \textbf{1.0} & 0.95 & $0.975_{\pm 0.0}$ & $0.9667_{\pm 0.014}$ & \textbf{1.0}$_{\pm 0.0}$ & $0.9833_{\pm 0.014}$ & \textbf{1.0}$_{\pm 0.0}$ & \textbf{1.0}$_{\pm 0.0}$ & \textbf{1.0}$_{\pm 0.0}$ & \textbf{1.0}$_{\pm 0.0}$\\
        CharacterTrajectories & \textbf{0.9814}$_{\pm 0.001}$ & 0.9673 & 0.9694 & $0.9698_{\pm 0.002}$ & $0.9603_{\pm 0.0}$ & $0.9631_{\pm 0.001}$ & $0.9373_{\pm 0.002}$ & $0.9517_{\pm 0.002}$ & $0.967_{\pm 0.002}$ & $0.9761_{\pm 0.001}$ & $0.9761_{\pm 0.0}$\\
        Cricket & $0.9444_{\pm 0.014}$ & 0.8194 & 0.8472 & $0.9722_{\pm 0.0}$ & $0.9954_{\pm 0.008}$ & $0.9861_{\pm 0.0}$ & \textbf{1.0}$_{\pm 0.0}$ & $0.9954_{\pm 0.008}$ & $0.9907_{\pm 0.008}$ & $0.9861_{\pm 0.0}$ & $0.9861_{\pm 0.0}$\\
        DuckDuckGeese & $0.5133_{\pm 0.023}$ & 0.32 & 0.28 & $0.4467_{\pm 0.042}$ & $0.42_{\pm 0.02}$ & $0.4133_{\pm 0.023}$ & $0.4333_{\pm 0.023}$ & $0.44_{\pm 0.04}$ & $0.4267_{\pm 0.023}$ & $0.5067_{\pm 0.031}$ & \textbf{0.54}$_{\pm 0.02}$\\
        ERing & $0.8827_{\pm 0.022}$ & 0.8889 & 0.8519 & $0.9679_{\pm 0.008}$ & $0.9765_{\pm 0.008}$ & \textbf{0.9877}$_{\pm 0.004}$ & $0.9815_{\pm 0.004}$ & $0.963_{\pm 0.007}$ & $0.9765_{\pm 0.004}$ & \textbf{0.9877}$_{\pm 0.006}$ & \textbf{0.9877}$_{\pm 0.002}$\\
        EigenWorms & $0.8753_{\pm 0.019}$ & 0.4198 & 0.5649 & $0.7608_{\pm 0.004}$ & $0.7863_{\pm 0.038}$ & $0.8015_{\pm 0.015}$ & $0.888_{\pm 0.027}$ & \textbf{0.9313}$_{\pm 0.008}$ & $0.7354_{\pm 0.022}$ & $0.8193_{\pm 0.004}$ & $0.8142_{\pm 0.016}$\\
        Epilepsy & $0.9855_{\pm 0.0}$ & 0.913 & 0.9565 & $0.9928_{\pm 0.0}$ & \textbf{1.0}$_{\pm 0.0}$ & \textbf{1.0}$_{\pm 0.0}$ & \textbf{1.0}$_{\pm 0.0}$ & \textbf{1.0}$_{\pm 0.0}$ & \textbf{1.0}$_{\pm 0.0}$ & \textbf{1.0}$_{\pm 0.0}$ & $0.9976_{\pm 0.004}$\\
        EthanolConcentration & $0.4271_{\pm 0.002}$ & \textbf{0.7833} & 0.6046 & $0.2953_{\pm 0.019}$ & $0.275_{\pm 0.006}$ & $0.3485_{\pm 0.01}$ & $0.3866_{\pm 0.008}$ & $0.3663_{\pm 0.012}$ & $0.4132_{\pm 0.01}$ & $0.4056_{\pm 0.014}$ & $0.4183_{\pm 0.007}$\\
        FaceDetection & $0.5273_{\pm 0.0}$ & 0.6325 & \textbf{0.6402} & $0.554_{\pm 0.005}$ & $0.6133_{\pm 0.006}$ & $0.5716_{\pm 0.011}$ & $0.55_{\pm 0.005}$ & $0.5321_{\pm 0.003}$ & $0.5691_{\pm 0.004}$ & $0.5566_{\pm 0.005}$ & $0.5492_{\pm 0.004}$\\
        FingerMovements & $0.4967_{\pm 0.064}$ & 0.5 & 0.49 & $0.53_{\pm 0.02}$ & $0.52_{\pm 0.04}$ & $0.5233_{\pm 0.021}$ & $0.5367_{\pm 0.045}$ & $0.5333_{\pm 0.059}$ & $0.5267_{\pm 0.015}$ & $0.51_{\pm 0.026}$ & \textbf{0.55}$_{\pm 0.01}$\\
        HandMovementDirection & $0.3378_{\pm 0.049}$ & \textbf{0.4324} & 0.3919 & $0.2928_{\pm 0.021}$ & $0.3153_{\pm 0.028}$ & $0.2748_{\pm 0.008}$ & $0.2883_{\pm 0.031}$ & $0.3288_{\pm 0.055}$ & $0.2883_{\pm 0.021}$ & $0.3243_{\pm 0.049}$ & $0.2793_{\pm 0.067}$\\
        Handwriting & $0.2812_{\pm 0.009}$ & 0.1388 & 0.2259 & $0.26_{\pm 0.002}$ & $0.2525_{\pm 0.002}$ & $0.2576_{\pm 0.007}$ & $0.2373_{\pm 0.009}$ & $0.2494_{\pm 0.011}$ & $0.2055_{\pm 0.007}$ & \textbf{0.3086}$_{\pm 0.002}$ & $0.2808_{\pm 0.008}$\\
        Heartbeat & $0.7496_{\pm 0.006}$ & 0.722 & 0.7268 & $0.7317_{\pm 0.005}$ & $0.7252_{\pm 0.006}$ & $0.7317_{\pm 0.008}$ & $0.7252_{\pm 0.003}$ & $0.735_{\pm 0.007}$ & $0.7756_{\pm 0.005}$ & \textbf{0.7984}$_{\pm 0.017}$ & $0.7919_{\pm 0.003}$\\
        InsectWingbeatSubset & $0.3617_{\pm 0.01}$ & 0.238 & \texttt{NaN} & $0.267_{\pm 0.017}$ & $0.287_{\pm 0.004}$ & $0.2803_{\pm 0.018}$ & $0.3203_{\pm 0.014}$ & $0.3243_{\pm 0.006}$ & $0.6143_{\pm 0.012}$ & \textbf{0.6277}$_{\pm 0.006}$ & $0.6073_{\pm 0.007}$\\
        JapaneseVowels & \textbf{0.955}$_{\pm 0.003}$ & 0.8135 & 0.8162 & $0.8847_{\pm 0.006}$ & $0.8928_{\pm 0.015}$ & $0.8162_{\pm 0.026}$ & $0.882_{\pm 0.006}$ & $0.8721_{\pm 0.004}$ & $0.9342_{\pm 0.006}$ & $0.9405_{\pm 0.0}$ & $0.9396_{\pm 0.006}$\\
        LSST & $0.6158_{\pm 0.002}$ & 0.5479 & 0.5016 & $0.6313_{\pm 0.003}$ & $0.5733_{\pm 0.003}$ & $0.5892_{\pm 0.004}$ & $0.5953_{\pm 0.006}$ & $0.6004_{\pm 0.004}$ & $0.5673_{\pm 0.005}$ & \textbf{0.6676}$_{\pm 0.004}$ & $0.6599_{\pm 0.004}$\\
        Libras & $0.8389_{\pm 0.01}$ & 0.6722 & 0.6389 & $0.8463_{\pm 0.008}$ & $0.8963_{\pm 0.008}$ & $0.8648_{\pm 0.008}$ & $0.9037_{\pm 0.006}$ & $0.9074_{\pm 0.008}$ & $0.8778_{\pm 0.01}$ & \textbf{0.9278}$_{\pm 0.0}$ & $0.9241_{\pm 0.003}$\\
        MotorImagery & $0.4333_{\pm 0.086}$ & \textbf{0.58} & 0.57 & $0.5233_{\pm 0.015}$ & $0.4933_{\pm 0.038}$ & $0.53_{\pm 0.026}$ & $0.51_{\pm 0.01}$ & $0.49_{\pm 0.036}$ & $0.48_{\pm 0.026}$ & $0.4967_{\pm 0.038}$ & $0.4667_{\pm 0.025}$\\
        NATOPS & $0.7148_{\pm 0.029}$ & 0.7778 & 0.7944 & $0.8352_{\pm 0.012}$ & $0.8296_{\pm 0.012}$ & $0.8389_{\pm 0.011}$ & $0.8685_{\pm 0.017}$ & $0.8574_{\pm 0.014}$ & $0.8333_{\pm 0.031}$ & \textbf{0.8926}$_{\pm 0.02}$ & $0.8889_{\pm 0.01}$\\
        PEMS-SF & $0.8401_{\pm 0.009}$ & 0.948 & 0.9422 & \textbf{0.9961}$_{\pm 0.007}$ & \textbf{0.9961}$_{\pm 0.007}$ & \textbf{0.9961}$_{\pm 0.007}$ & $0.973_{\pm 0.007}$ & $0.9769_{\pm 0.0}$ & $0.9923_{\pm 0.007}$ & \textbf{0.9961}$_{\pm 0.007}$ & \textbf{0.9961}$_{\pm 0.007}$\\
        PhonemeSpectra & $0.2499_{\pm 0.003}$ & 0.1485 & 0.1712 & $0.2112_{\pm 0.006}$ & $0.2686_{\pm 0.003}$ & $0.2709_{\pm 0.003}$ & $0.2713_{\pm 0.005}$ & $0.2709_{\pm 0.003}$ & $0.2664_{\pm 0.005}$ & $0.3107_{\pm 0.005}$ & \textbf{0.3213}$_{\pm 0.004}$\\
        RacketSports & $0.8004_{\pm 0.004}$ & 0.8158 & 0.8487 & $0.8465_{\pm 0.025}$ & $0.8355_{\pm 0.007}$ & $0.8246_{\pm 0.03}$ & $0.8531_{\pm 0.01}$ & $0.8509_{\pm 0.004}$ & $0.9123_{\pm 0.019}$ & \textbf{0.9232}$_{\pm 0.01}$ & $0.9101_{\pm 0.008}$\\
        SelfRegulationSCP1 & $0.7702_{\pm 0.007}$ & \textbf{0.8942} & 0.8874 & $0.7747_{\pm 0.006}$ & $0.7884_{\pm 0.012}$ & $0.785_{\pm 0.003}$ & $0.7986_{\pm 0.007}$ & $0.7929_{\pm 0.01}$ & $0.7952_{\pm 0.003}$ & $0.7736_{\pm 0.005}$ & $0.8134_{\pm 0.009}$\\
        SelfRegulationSCP2 & $0.4926_{\pm 0.031}$ & 0.4778 & 0.5056 & $0.4907_{\pm 0.023}$ & $0.4963_{\pm 0.049}$ & $0.5167_{\pm 0.006}$ & $0.4907_{\pm 0.033}$ & $0.5056_{\pm 0.011}$ & $0.5074_{\pm 0.018}$ & \textbf{0.5611}$_{\pm 0.02}$ & $0.5167_{\pm 0.006}$\\
        SpokenArabicDigits & $0.8898_{\pm 0.003}$ & 0.9591 & \textbf{0.9613} & $0.9447_{\pm 0.003}$ & $0.7547_{\pm 0.01}$ & $0.7619_{\pm 0.006}$ & $0.8931_{\pm 0.005}$ & $0.9139_{\pm 0.004}$ & $0.9016_{\pm 0.002}$ & $0.9285_{\pm 0.001}$ & $0.9374_{\pm 0.003}$\\
        UWaveGestureLibrary & $0.876_{\pm 0.015}$ & 0.8062 & 0.7625 & \textbf{0.8917}$_{\pm 0.007}$ & $0.8615_{\pm 0.002}$ & $0.8833_{\pm 0.004}$ & $0.8542_{\pm 0.004}$ & $0.8156_{\pm 0.005}$ & $0.8885_{\pm 0.007}$ & $0.8906_{\pm 0.008}$ & $0.8896_{\pm 0.011}$\\
        \midrule
        \textbf{\textit{Average}} & 0.6963 & 0.6721 & 0.685 & 0.6991 & 0.6952 & 0.6967 & 0.7091 & 0.7105 & 0.7199 & \textbf{0.7449} & 0.742\\
        \textit{\textbf{Best Counts}} & 3 & 5 & 2 & 2 & 2 & 4 & 2 & 3 & 2 & \textbf{12} & 6\\
        \bottomrule
    \end{tabular}
    }}
\end{table}

\subsection{Tables for Section \ref{sec:log-reg}}
\label{sec:log-reg-exp-appendix}

We would like also to provide the complete results for all benchmarks when we use the Logistic Regression as a classifier. Table \ref{tab:sota-ucr-res-logreg-a}  and Table \ref{tab:sota-ucr-res-logreg-b} correspond to the results on UCR, Table \ref{tab:uea-sota-logreg} to UEA, Table \ref{tab:har-sota-logreg} to HAR and Table \ref{tab:eeg-sota-logreg} to EEG.

\vspace{0.3cm}
\begin{table}[H]
    \centering
    \caption{Zero-shot feature extraction results on UCR (Logistic Regression). First Part.}
    \label{tab:sota-ucr-res-logreg-a}
    {\scalebox{0.58}{
\begin{tabular}{l|lllllllllll}
    \toprule
                                   & Catch22+ & TabPFN & TabICL & MOMENT & TiRex & Chronos2 & TiViT-H & TiConvNext & NuTime & Mantis+ & MantisV2\\
    \midrule
    ACSF1 & 0.7 & 0.8 & 0.81 & 0.75 & 0.75 & \textbf{0.86} & 0.83 & 0.83 & 0.7 & 0.79 & 0.76\\
    Adiac & 0.665 & 0.8031 & 0.8031 & 0.7775 & 0.7826 & 0.8286 & 0.7187 & 0.7059 & 0.8005 & \textbf{0.8414} & 0.8363\\
    AllGestureWiimoteX & 0.5486 & 0.6229 & 0.5043 & 0.6614 & 0.7014 & 0.6843 & 0.67 & 0.6943 & 0.6071 & 0.7171 & \textbf{0.73}\\
    AllGestureWiimoteY & 0.5229 & 0.6329 & 0.5114 & 0.7029 & 0.7314 & 0.7043 & 0.7371 & 0.7229 & 0.6529 & 0.7414 & \textbf{0.7471}\\
    AllGestureWiimoteZ & 0.4829 & 0.5329 & 0.4529 & 0.6057 & 0.6343 & 0.66 & 0.6771 & 0.6586 & 0.5414 & 0.6657 & \textbf{0.6814}\\
    ArrowHead & 0.7143 & 0.7543 & 0.7429 & 0.7771 & 0.7829 & \textbf{0.8514} & 0.8229 & 0.8229 & 0.7257 & 0.8457 & \textbf{0.8514}\\
    BME & 0.9933 & \textbf{1.0} & 0.98 & 0.9933 & 0.9933 & \textbf{1.0} & \textbf{1.0} & \textbf{1.0} & 0.9667 & \textbf{1.0} & \textbf{1.0}\\
\bottomrule
    \end{tabular}
    }}
\end{table}
\newpage

\begin{table}[H]
    \centering
    \caption{Zero-shot feature extraction results on UCR (Logistic Regression). Second Part.}
    \label{tab:sota-ucr-res-logreg-b}
    {\scalebox{0.58}{
    \begin{tabular}{l|lllllllllll}
    \toprule
                                   & Catch22+ & TabPFN & TabICL & MOMENT & TiRex & Chronos2 & TiViT-H & TiConvNext & NuTime & Mantis+ & MantisV2\\
    \midrule
    ItalyPowerDemand & 0.9388 & \textbf{0.9699} & 0.9631 & 0.9543 & 0.9602 & 0.964 & 0.9407 & 0.9378 & 0.9116 & 0.9281 & 0.9456\\
    LargeKitchenAppliances & 0.8373 & 0.6373 & 0.696 & 0.76 & 0.8293 & 0.8613 & 0.872 & \textbf{0.8933} & 0.7707 & 0.824 & 0.816\\
    Lightning2 & 0.7541 & 0.6721 & 0.7049 & 0.7869 & 0.7377 & 0.7869 & 0.8197 & \textbf{0.8361} & 0.7377 & 0.7705 & 0.7869\\
    Lightning7 & 0.6986 & 0.6986 & 0.7397 & 0.7397 & 0.8082 & 0.7945 & 0.8493 & 0.7671 & 0.7808 & 0.8356 & \textbf{0.863}\\
    Mallat & 0.8866 & 0.9689 & 0.9484 & 0.9186 & 0.9548 & 0.9467 & \textbf{0.9765} & 0.9642 & 0.8486 & 0.9049 & 0.9198\\
    Meat & 0.9333 & 0.9833 & 0.9333 & \textbf{1.0} & 0.9 & 0.9333 & 0.8167 & 0.85 & 0.9333 & 0.9833 & 0.9667\\
    MedicalImages & 0.6789 & 0.7947 & \textbf{0.8079} & 0.7461 & 0.725 & 0.7474 & 0.7316 & 0.7513 & 0.7092 & 0.7618 & 0.7711\\
\bottomrule
    \end{tabular}
    }}
\end{table}

\begin{table}[H]
    \centering
    \caption{Zero-shot feature extraction results on UEA-27 (Logistic Regression).}
    \label{tab:uea-sota-logreg}
    {\scalebox{0.6}{
    \begin{tabular}{l|lllllllllll}
    \toprule
                                   & Catch22+ & TabPFN & TabICL & MOMENT & TiRex & Chronos2 & TiViT-H & TiConvNext & NuTime & Mantis+ & MantisV2\\
    \midrule
    ArticularyWordRecognition & 0.9333 & 0.93 & 0.91 & 0.9867 & 0.99 & \textbf{0.9933} & 0.9833 & 0.9767 & \textbf{0.9933} & \textbf{0.9933} & \textbf{0.9933}\\
    BasicMotions & \textbf{1.0} & \textbf{1.0} & 0.95 & \textbf{1.0} & \textbf{1.0} & \textbf{1.0} & \textbf{1.0} & \textbf{1.0} & \textbf{1.0} & \textbf{1.0} & \textbf{1.0}\\
    CharacterTrajectories & 0.984 & 0.9673 & 0.9694 & 0.984 & 0.9694 & 0.9742 & 0.9659 & 0.9714 & 0.9847 & 0.9812 & \textbf{0.9868}\\
    Cricket & 0.9028 & 0.8194 & 0.8472 & 0.9861 & \textbf{1.0} & \textbf{1.0} & \textbf{1.0} & \textbf{1.0} & \textbf{1.0} & 0.9861 & 0.9861\\
    DuckDuckGeese & 0.42 & 0.32 & 0.28 & 0.48 & 0.38 & 0.34 & 0.48 & \textbf{0.56} & 0.5 & 0.54 & 0.5\\
    ERing & 0.9074 & 0.8889 & 0.8519 & 0.9704 & 0.9778 & 0.9778 & 0.9852 & 0.9815 & 0.9815 & \textbf{0.9926} & 0.9889\\
    EigenWorms & 0.7634 & 0.4198 & 0.5649 & 0.8321 & 0.8397 & 0.8397 & 0.9313 & \textbf{0.9618} & 0.8702 & 0.916 & 0.8855\\
\bottomrule
    \end{tabular}
    }}
\end{table}

\vspace{1cm}
\begin{table}[H]
    \centering
    \caption{Zero-shot feature extraction results on HAR (Logistic Regression).}
    \label{tab:har-sota-logreg}
    {\scalebox{0.60}{
    \begin{tabular}{l|lllllllllll}
    \toprule
                                   & Catch22+ & TabPFN & TabICL & MOMENT & TiRex & Chronos2 & TiViT-H & TiConvNext & NuTime & Mantis+ & MantisV2\\
    \midrule
    Ego4D & $0.3719_{\pm 0.0}$ & \texttt{NaN} & \texttt{NaN} & $0.4324_{\pm 0.0}$ & $0.5267_{\pm 0.0}$ & $0.5216_{\pm 0.0}$ & 0.1394 & 0.1288 & $0.5248_{\pm 0.0}$ & $0.5519_{\pm 0.0}$ & \textbf{0.5551}$_{\pm 0.0}$\\
    EMOPain & $0.7915_{\pm 0.0}$ & 0.7831 & 0.7831 & $0.8225_{\pm 0.0}$ & $0.8254_{\pm 0.0}$ & $0.8338_{\pm 0.0}$ & \textbf{0.8507}$_{\pm 0.0}$ & $0.8028_{\pm 0.0}$ & $0.7831_{\pm 0.0}$ & $0.8451_{\pm 0.0}$ & $0.8169_{\pm 0.0}$\\
    HHAR-ID & $0.9111_{\pm 0.0}$ & 0.8938 & 0.9073 & $0.9491_{\pm 0.0}$ & $0.9708_{\pm 0.0}$ & $0.9699_{\pm 0.0}$ & $0.9661_{\pm 0.0}$ & $0.9693_{\pm 0.0}$ & $0.9751_{\pm 0.0}$ & $0.9871_{\pm 0.0}$ & \textbf{0.9883}$_{\pm 0.0}$\\
    HHAR-OOD & $0.4085_{\pm 0.0}$ & 0.5311 & 0.5441 & $0.3329_{\pm 0.0}$ & $0.3338_{\pm 0.0}$ & $0.3691_{\pm 0.0}$ & $0.318_{\pm 0.0}$ & $0.3222_{\pm 0.0}$ & $0.5084_{\pm 0.0}$ & $0.4907_{\pm 0.0}$ & \textbf{0.5604}$_{\pm 0.0}$\\
    MP8 & $0.642_{\pm 0.0}$ & 0.6235 & 0.6185 & $0.595_{\pm 0.0}$ & $0.6134_{\pm 0.0}$ & $0.6387_{\pm 0.0}$ & $0.6403_{\pm 0.0}$ & $0.642_{\pm 0.0}$ & $0.6084_{\pm 0.0}$ & \textbf{0.6588}$_{\pm 0.0}$ & $0.6521_{\pm 0.0}$\\
    MP50 & $0.4471_{\pm 0.0}$ & 0.5664 & 0.5042 & $0.7294_{\pm 0.0}$ & \textbf{0.7563}$_{\pm 0.0}$ & $0.7345_{\pm 0.0}$ & $0.7193_{\pm 0.0}$ & $0.6857_{\pm 0.0}$ & $0.6992_{\pm 0.0}$ & $0.6924_{\pm 0.0}$ & $0.716_{\pm 0.0}$\\
    UCI-HAR & $0.8483_{\pm 0.0}$ & 0.809 & 0.8157 & $0.8619_{\pm 0.0}$ & $0.8887_{\pm 0.0}$ & $0.8809_{\pm 0.0}$ & \textbf{0.9175}$_{\pm 0.0}$ & $0.9118_{\pm 0.0}$ & $0.8806_{\pm 0.0}$ & $0.8962_{\pm 0.0}$ & $0.9145_{\pm 0.0}$\\
\bottomrule
    \end{tabular}
    }}
\end{table}

\vspace{1cm}
\begin{table}[H]
    \centering
    \caption{Zero-shot feature extraction results on EEG (Logistic Regression).}
    \label{tab:eeg-sota-logreg}
    {\scalebox{0.60}{
    \begin{tabular}{l|lllllllllll}
    \toprule
                                   & Catch22+ & TabPFN & TabICL & MOMENT & TiRex & Chronos2 & TiViT-H & TiConvNext & NuTime & Mantis+ & MantisV2\\
    \midrule
    Blink & 0.96 & 0.9178 & 0.8978 & 0.9867 & 0.9889 & \textbf{1.0} & 0.9889 & 0.9822 & 0.5911 & 0.9489 & 0.9733\\
    CAP & 0.6415 & \texttt{NaN} & \texttt{NaN} & 0.7374 & 0.801 & \textbf{0.8168} & 0.8139 & 0.815 & 0.7767 & 0.8078 & 0.7945\\
    CAP-OOD & 0.6704 & \texttt{NaN} & \texttt{NaN} & 0.6946 & 0.7224 & \textbf{0.7535} & 0.7223 & 0.729 & 0.633 & 0.6415 & 0.6391\\
    Epilepsy-EEG & 0.947 & 0.9496 & 0.9447 & 0.9366 & \textbf{0.9563} & 0.9414 & 0.9322 & 0.9186 & 0.907 & 0.9424 & 0.9312\\
    FingerMovements & 0.52 & 0.5 & 0.49 & 0.57 & 0.62 & 0.6 & 0.57 & 0.53 & 0.61 & 0.57 & \textbf{0.64}\\
    PCL & 0.5402 & \texttt{NaN} & \texttt{NaN} & 0.5616 & 0.5667 & 0.5596 & 0.5685 & 0.5731 & 0.5754 & 0.575 & \textbf{0.5931}\\
    PCL-OOD & 0.5279 & \texttt{NaN} & \texttt{NaN} & 0.5215 & 0.5305 & 0.5291 & 0.5259 & 0.5237 & 0.5381 & 0.5329 & \textbf{0.5459}\\
    SEDFx & 0.7113 & \texttt{NaN} & \texttt{NaN} & 0.7714 & 0.8116 & \textbf{0.8218} & 0.8019 & 0.8097 & 0.7878 & 0.8126 & 0.8104\\
    SEDFx-OOD & 0.6934 & \texttt{NaN} & \texttt{NaN} & 0.7423 & 0.7763 & 0.7826 & 0.7692 & 0.7674 & 0.767 & 0.7777 & \textbf{0.7841}\\
    SelfRegulationSCP1 & 0.744 & \textbf{0.8942} & 0.8874 & 0.8259 & 0.8498 & 0.8498 & 0.8498 & 0.8498 & 0.8567 & 0.8737 & 0.8669\\
    SelfRegulationSCP2 & 0.4778 & 0.4778 & 0.5056 & 0.5111 & 0.4944 & 0.4833 & 0.5278 & 0.4944 & \textbf{0.55} & 0.5389 & 0.5111\\
    \midrule
    Avg & 0.6758 & \texttt{NaN} & \texttt{NaN} & 0.7145 & 0.738 & \textbf{0.7398} & 0.7337 & 0.7266 & 0.6902 & 0.7292 & 0.7354\\
    \bottomrule
    \end{tabular}
    }}
\end{table}

\newpage

\subsection{Tables for Section \ref{sec:final-comparison}}
Finally, we provide the complete tables that correspond to Figure \ref{fig:teaser-plot} (Table \ref{tab:final-sota-ucr-res-a} and Table \ref{tab:final-sota-ucr-res-b}) and Figure \ref{fig:more-baselines-comparison} (Table \ref{tab:more-baselines-comparison}).
\vspace{0.4cm}
\begin{table}[H]
    \centering
    \caption{The final comparison on UCR. First Part.}
    \label{tab:final-sota-ucr-res-a}
    {\scalebox{0.45}{
    \begin{tabular}{l|lllllllllll|ll|ll|l}
    \toprule
     & Catch22+ & TabPFN & TabICL & MOMENT & TiRex & Chronos2 & TiViT-H & TiConvNext & NuTime & Mantis+ & MantisV2 & SE-Mantis+ & SE-MantisV2 & MantisV2 \&  & MantisV2 \& & MantisV2-FT\\
     & & & & & & & & & & & & & &  TiViT-H & TiConvNext & \\
    \midrule
    ACSF1 & $0.8233_{\pm 0.021}$ & 0.8 & 0.81 & 0.75 & 0.75 & \textbf{0.86} & 0.83 & 0.83 & 0.7 & 0.79 & 0.76 & 0.8 & 0.77 & 0.81 & 0.82 & $0.77_{\pm 0.01}$\\
    Adiac & $0.734_{\pm 0.007}$ & 0.8031 & 0.8031 & 0.7775 & 0.7826 & 0.8286 & 0.7187 & 0.7059 & 0.8005 & 0.8414 & 0.8363 & 0.8312 & \textbf{0.8465} & 0.7519 & 0.7749 & $0.8321_{\pm 0.007}$\\
    AllGestureWiimoteX & $0.5957_{\pm 0.01}$ & 0.6229 & 0.5043 & 0.6614 & 0.7014 & 0.6843 & 0.67 & 0.6943 & 0.6071 & 0.7171 & 0.73 & 0.69 & 0.6871 & 0.7229 & 0.7357 & \textbf{0.779}$_{\pm 0.006}$\\
    AllGestureWiimoteY & $0.6319_{\pm 0.016}$ & 0.6329 & 0.5114 & 0.7029 & 0.7314 & 0.7043 & 0.7371 & 0.7229 & 0.6529 & 0.7414 & 0.7471 & 0.73 & 0.7271 & 0.7557 & 0.7486 & \textbf{0.7905}$_{\pm 0.004}$\\
    AllGestureWiimoteZ & $0.5462_{\pm 0.005}$ & 0.5329 & 0.4529 & 0.6057 & 0.6343 & 0.66 & 0.6771 & 0.6586 & 0.5414 & 0.6657 & 0.6814 & 0.6329 & 0.6357 & 0.6886 & 0.7043 & \textbf{0.7505}$_{\pm 0.017}$\\
    ArrowHead & $0.741_{\pm 0.012}$ & 0.7543 & 0.7429 & 0.7771 & 0.7829 & 0.8514 & 0.8229 & 0.8229 & 0.7257 & 0.8457 & 0.8514 & 0.84 & 0.8514 & \textbf{0.88} & 0.8457 & $0.8457_{\pm 0.006}$\\
\bottomrule
    \end{tabular}
    }}
\end{table}

\newpage

\begin{table}[H]
    \centering
    \caption{The final comparison on UCR. Second Part.}
    \label{tab:final-sota-ucr-res-b}
    {\scalebox{0.45}{
    \begin{tabular}{l|lllllllllll|ll|ll|l}
    \toprule
     & Catch22+ & TabPFN & TabICL & MOMENT & TiRex & Chronos2 & TiViT-H & TiConvNext & NuTime & Mantis+ & MantisV2 & SE-Mantis+ & SE-MantisV2 & MantisV2 \&  & MantisV2 \& & MantisV2-FT\\
     & & & & & & & & & & & & & &  TiViT-H & TiConvNext & \\
    \midrule
    ItalyPowerDemand & $0.9537_{\pm 0.002}$ & \textbf{0.9699} & 0.9631 & 0.9543 & 0.9602 & 0.964 & 0.9407 & 0.9378 & 0.9116 & 0.9281 & 0.9456 & 0.9436 & 0.9475 & 0.9427 & 0.9446 & $0.9517_{\pm 0.001}$\\
    LargeKitchenAppliances & $0.8169_{\pm 0.009}$ & 0.6373 & 0.696 & 0.76 & 0.8293 & 0.8613 & 0.872 & \textbf{0.8933} & 0.7707 & 0.824 & 0.816 & 0.832 & 0.816 & 0.8507 & 0.888 & $0.8596_{\pm 0.008}$\\
    Lightning2 & $0.7322_{\pm 0.009}$ & 0.6721 & 0.7049 & 0.7869 & 0.7377 & 0.7869 & 0.8197 & 0.8361 & 0.7377 & 0.7705 & 0.7869 & 0.8361 & 0.7705 & \textbf{0.8525} & 0.8033 & $0.847_{\pm 0.025}$\\
    Lightning7 & $0.7671_{\pm 0.014}$ & 0.6986 & 0.7397 & 0.7397 & 0.8082 & 0.7945 & 0.8493 & 0.7671 & 0.7808 & 0.8356 & 0.863 & 0.863 & 0.8219 & \textbf{0.8767} & 0.7945 & $0.7717_{\pm 0.032}$\\
    Mallat & $0.9555_{\pm 0.009}$ & 0.9689 & 0.9484 & 0.9186 & 0.9548 & 0.9467 & \textbf{0.9765} & 0.9642 & 0.8486 & 0.9049 & 0.9198 & 0.9339 & 0.9663 & 0.974 & 0.9697 & $0.9512_{\pm 0.021}$\\
    Meat & $0.9222_{\pm 0.01}$ & 0.9833 & 0.9333 & \textbf{1.0} & 0.9 & 0.9333 & 0.8167 & 0.85 & 0.9333 & 0.9833 & 0.9667 & 0.9 & 0.8833 & 0.8833 & 0.8833 & $0.8833_{\pm 0.033}$\\
\bottomrule
    \end{tabular}
    }}
\end{table}

\begin{table}[H]
\addtolength{\tabcolsep}{-0.3em}
\caption{Comparison with more baselines (taken from \citealp{goswami2024moment}) on 91 UCR datasets.}
\label{tab:more-baselines-comparison}
{\scalebox{0.36}{
    \begin{tabular}{l|lllllllllllllllllllllllllll}
    \toprule
    & TS2Vec & T-Loss & TNC & TS-TCC & CNN & Encoder & FCN & MLP & ResNet & TWIESN & DTW & Official & Improved & Catch22+ & TabPFN & TabICL & TiRex & Chronos2 & TiViT-H & TiConvNext & NuTime & Mantis+ & MantisV2 & SE-Mantis+ & SE-MantisV2 & MantisV2 \& & MantisV2 \& \\
    & &  &  & &  &  & &  & &  & & MOMENT & MOMENT &  &  &  & &  & &  &  &  &  &  &  & TiViT-H & TiConvNext\\
    \midrule
    Adiac & 0.762 & 0.675 & 0.767 & 0.767 & 0.393 & 0.318 & 0.841 & 0.391 & 0.833 & 0.428 & 0.604 & 0.688 & 0.7775 & 0.734 & 0.8031 & 0.8031 & 0.7826 & 0.8286 & 0.7187 & 0.7059 & 0.8005 & 0.8414 & 0.8363 & 0.8312 & \textbf{0.8465} & 0.7519 & 0.7749\\
    AllGestureWiimoteX & \textbf{0.777} & 0.763 & 0.697 & 0.697 & 0.411 & 0.475 & 0.713 & 0.477 & 0.741 & 0.522 & 0.716 & 0.607 & 0.6614 & 0.5957 & 0.6229 & 0.5043 & 0.7014 & 0.6843 & 0.67 & 0.6943 & 0.6071 & 0.7171 & 0.73 & 0.69 & 0.6871 & 0.7229 & 0.7357\\
    AllGestureWiimoteY & 0.793 & 0.726 & 0.741 & 0.741 & 0.479 & 0.509 & 0.784 & 0.571 & \textbf{0.794} & 0.6 & 0.729 & 0.666 & 0.7029 & 0.6319 & 0.6329 & 0.5114 & 0.7314 & 0.7043 & 0.7371 & 0.7229 & 0.6529 & 0.7414 & 0.7471 & 0.73 & 0.7271 & 0.7557 & 0.7486\\
    AllGestureWiimoteZ & \textbf{0.746} & 0.723 & 0.689 & 0.689 & 0.375 & 0.396 & 0.692 & 0.439 & 0.726 & 0.516 & 0.643 & 0.537 & 0.6057 & 0.5462 & 0.5329 & 0.4529 & 0.6343 & 0.66 & 0.6771 & 0.6586 & 0.5414 & 0.6657 & 0.6814 & 0.6329 & 0.6357 & 0.6886 & 0.7043\\
    ArrowHead & 0.857 & 0.766 & 0.737 & 0.737 & 0.717 & 0.63 & 0.843 & 0.784 & 0.838 & 0.689 & 0.703 & 0.743 & 0.7771 & 0.741 & 0.7543 & 0.7429 & 0.7829 & 0.8514 & 0.8229 & 0.8229 & 0.7257 & 0.8457 & 0.8514 & 0.84 & 0.8514 & \textbf{0.88} & 0.8457\\
\bottomrule
    \end{tabular}
}}
\end{table}

\end{document}